\let\TeXyear\year
\PassOptionsToPackage{dvipsnames}{xcolor}
\documentclass{IEEEaccess}
\let\setyear\year
\let\year\TeXyear

\usepackage{microtype}
\usepackage{etoolbox}

\patchcmd{\appendices}
{\thesectiondis{APPENDIX \Alph{section}}.}
{\thesectiondis{APPENDIX \Alph{section}.}}
{}{}

\setyear{2022}
\vol{V}
\makeatletter
\patchcmd{\ps@titlepage}{2016}{\theyear}{}{}
\patchcmd{\ps@headings}{2016}{\theyear}{}{}
\patchcmd{\ps@headings}{2016}{\theyear}{}{}
\ps@headings
\makeatother

\usepackage{cite}

\usepackage{xr}
\usepackage{graphicx}
\graphicspath{{./img/}}
\DeclareGraphicsExtensions{.pdf,.jpeg,.jpg,.png,.tex,.eps}

\usepackage[labelformat=simple]{subcaption}
\usepackage{etoolbox}
\usepackage{algorithm}
\usepackage{algpseudocode}
\algrenewcommand{\algorithmiccomment}[1]{\hfill\textcolor{black!50}{$\triangleright$ #1}}

\usepackage{tikz}
\NewSpotColorSpace{PANTONE}
\AddSpotColor{PANTONE} {PANTONE3015C} {PANTONE\SpotSpace 3015\SpotSpace C} {1 0.3 0 0.2}
\SetPageColorSpace{PANTONE}%

\usetikzlibrary{%
  calc,
  shapes,
  positioning,
  fit,
  scopes,
  intersections,
  3d,
  plotmarks,
  external,
  arrows.meta,
}

\usepackage{pgfplots}
\usepgfplotslibrary{%
  colorbrewer,
  external,
  fillbetween,
  groupplots,
}

\tikzset{%
}

\pgfplotsset{
  width=\linewidth, height=0.85\linewidth, %
  every axis plot post/.append style={thick}, %
  ymajorgrids,
  grid style={dashed}, %
  minimal plot grid/.style={
    y axis line style={opacity=0},
    axis x line*=bottom,
    x axis line style={black},
  },
  minimal plot grid,
  /pgfplots/legend pos/south center/.style={/pgfplots/legend style={at={(0.5, 0.03)}, anchor=south}},
  /pgfplots/legend pos/north center/.style={/pgfplots/legend style={at={(0.5, 0.97)}, anchor=north}},
  /pgfplots/legend pos/outer north center/.style={/pgfplots/legend style={at={(0.5, 1.02)}, anchor=south}},
  y tick label style={
    /pgf/number format/.cd,
    fixed,
    fixed zerofill,
    precision=1,
    /tikz/.cd,
    font=\footnotesize,
  },
  x tick label style={
    /pgf/number format/.cd,
    fixed,
    fixed zerofill,
    precision=1,
    /tikz/.cd,
    font=\footnotesize,
  },
}

\usepackage{amsmath}
\usepackage{mathtools}
\usepackage{commath}
\usepackage{amsfonts}           %
\usepackage{nicefrac}           %
\usepackage{amssymb}

\usepackage{array}
\usepackage{multirow}
\usepackage{booktabs}
\usepackage{siunitx}
\DeclareMathOperator{\Cat}{\Vert}

\DeclarePairedDelimiterX{\infdivx}[2]{(}{)}{%
  #1\;\delimsize\|\;#2%
}
\newcommand{\kl}{\operatorname{KL}\infdivx}

\let\given\givenbase

\makeatletter
\DeclareRobustCommand\bigop[1]{%
  \mathop{\vphantom{\sum}\mathpalette\bigop@{#1}}\slimits@
}
\newcommand{\bigop@}[2]{%
  \vcenter{%
    \sbox\z@{$#1\sum$}%
    \hbox{\resizebox{\ifx#1\displaystyle.9\fi\dimexpr\ht\z@+\dp\z@}{!}{$\m@th#2$}}%
  }%
}

\newcommand{\class}[2]{{\tikz{\draw[#1, fill=#1] (0, 0) circle (#2mm)}}}

\usepackage{tkz-kiviat}
\makeatletter
\def\tkz@KiviatGrad[#1](#2){%
  \begingroup
  \pgfkeys{/kiviatgrad/.cd,
    graduation distance= 0 pt,
    prefix ={},
    suffix={},
    unity=1,
    label precision/.store in=\gradlabel@precision,
    label precision=1,
    zero point/.store in=\tkz@grad@zero,
    zero point=0
  }
  \pgfqkeys{/kiviatgrad}{#1}%
  \let\tikz@label@distance@tmp\tikz@label@distance
  \global\let\tikz@label@distance\tkz@kiv@grad
  \foreach \nv in {0,...,\tkz@kiv@lattice}{
    \pgfmathsetmacro{\result}{\tkz@kiv@unity*\nv-\tkz@grad@zero} %
    \protected@edef\tkz@kiv@gd{%
      \tkz@kiv@prefix%
      \pgfmathprintnumber[precision=\gradlabel@precision,fixed]{\result}%
      \tkz@kiv@suffix}
    \path[/kiviatgrad/.cd,#1] (0:0)--(360/\tkz@kiv@radial*#2:\nv*\tkz@kiv@gap)
    node[label=(360/\tkz@kiv@radial*#2):\scriptsize\tkz@kiv@gd] {}; %
  }
  \let\tikz@label@distance\tikz@label@distance@tmp
  \endgroup
}%

\def\tkz@KiviatLine[#1](#2,#3){%
  \begingroup
  \pgfkeys{/kiviatline/.cd,
    fill= {},
    opacity=.5,
    zero point/.store in=\tkz@line@zero,
    zero point=0
  }
  \pgfqkeys{/kiviatline}{#1}%
  \ifx\tkzutil@empty\tkz@kivl@fill \else
  \begin{scope}[on background layer]
    \path[fill=\tkz@kivl@fill,opacity=\tkz@kivl@opacity] (360/\tkz@kiv@radial*0:{(#2+\tkz@line@zero)*\tkz@kiv@gap*\tkz@kiv@step})
    \foreach \v [count=\rang from 1] in {#3}{%
      -- (360/\tkz@kiv@radial*\rang:{(\v+\tkz@line@zero)*\tkz@kiv@gap*\tkz@kiv@step}) } -- (360/\tkz@kiv@radial*0:{(#2+\tkz@line@zero)*\tkz@kiv@gap*\tkz@kiv@step});
  \end{scope}
  \fi
  \draw[#1,opacity=1,overlay] (0:{(#2+\tkz@line@zero)*\tkz@kiv@gap}) plot coordinates {(360/\tkz@kiv@radial*0:{(#2+\tkz@line@zero)*\tkz@kiv@gap*\tkz@kiv@step})}
  \foreach \v [count=\rang from 1] in {#3}{%
    -- (360/\tkz@kiv@radial*\rang:{(\v+\tkz@line@zero)*\tkz@kiv@gap*\tkz@kiv@step}) plot coordinates {(360/\tkz@kiv@radial*\rang:{(\v+\tkz@line@zero)*\tkz@kiv@gap*\tkz@kiv@step})}} -- (360/\tkz@kiv@radial*0:{(#2+\tkz@line@zero)*\tkz@kiv@gap*\tkz@kiv@step});
  \endgroup
}%

\makeatother

\let\oldtimes\times
\def\times{{\mkern1mu\oldtimes\mkern1mu}}

\usepackage{xspace}
\makeatletter
\DeclareRobustCommand\onedot{\futurelet\@let@token\@onedot}
\def\@onedot{\ifx\@let@token.\else.\null\fi\xspace}

\def\eg{{e.g}\onedot} 
\def\ie{{i.e}\onedot} 
\def\cf{{cf}\onedot} 
 \def\vs{{vs}\onedot}
\def\wrt{w.r.t\onedot} 
 
\def\etal{{et al}\onedot}

\usepackage[]{hyperref}  %
\hypersetup{
  colorlinks,
  linkcolor = BrickRed,
  citecolor = NavyBlue,
  urlcolor  = WildStrawberry,
}

\begin{document}
\history{Pre-print to appear in IEEE Access}
\doi{10.1109/ACCESS.2022.3192605}

\title{Global and Local Features through Gaussian Mixture Models on Image Semantic Segmentation}
\author{\uppercase{Darwin Saire,\authorrefmark{1} and Ad\'in Ram\'irez~Rivera,\authorrefmark{1,2}}
\IEEEmembership{Senior Member, IEEE}}
\address[1]{Institute of Computing, University of Campinas, Brazil (e-mail: \texttt{darwin.pilco@ic.unicamp.br})}
\address[2]{Department of Informatics, University of Oslo, Norway (e-mail: \texttt{adinr@uio.no})}
\tfootnote{This work was financed in part by the S{\~a}o Paulo Research Foundation (FAPESP) under grants No.~2017/16597-7, 2019/07257-3, and~2019/18678-0, and in part by the Brazilian National Council for Scientific and Technological Development (CNPq) under grant No.~307425/2017-7.  The Laboratoire Lorrain de Recherche en Informatique et ses Applications (LORIA) institute provided, in part, infrastructure for this work.
\newline
Code available at \url{https://gitlab.com/mipl/phgmm}.
}

\markboth{Saire and Ram\'irez~Rivera: Global and Local Features through Gaussian Mixture Models on Image Semantic Segmentation}%
{Saire and Ram\'irez~Rivera: Global and Local Features through Gaussian Mixture Models on Image Semantic Segmentation}

\corresp{Corresponding author: Ad\'in Ram\'irez~Rivera (e-mail: \texttt{adinr@uio.no}).}


\titlepgskip=-15pt

\begin{abstract}
The semantic segmentation task aims at dense classification at the pixel-wise level.
Deep models exhibited progress in tackling this task.
However, one remaining problem with these approaches is the loss of spatial precision, often produced at the segmented objects' boundaries.
Our proposed model addresses this problem by providing an internal structure for the feature representations while extracting a global representation that supports the former.
To fit the internal structure, during training, we predict a Gaussian Mixture Model from the data, which, merged with the skip connections and the decoding stage, helps avoid wrong inductive biases.
Furthermore, our results show that we can improve semantic segmentation by providing both learning representations (global and local) with a clustering behavior and combining them.
Finally, we present results demonstrating our advances in Cityscapes and Synthia datasets.
\end{abstract}

\maketitle

\section{Introduction}
\label{sec:introduction}
Humans naturally (innate way) receive and understand a large amount of information (multiple samples)~\cite{Tenenbaum1983}.
Trying to replicate this process through computers with visual perception, \ie, endowing the machines the ability to see, is called computer vision~\cite{Gonzalez2006}, which is difficult due largely to variance.
A vision system is required to infer objects across huge variations in pose, appearance, viewpoint, illumination, and occlusion throughout the image.

Scene understanding is one of the most challenging tasks in computer vision.
It plays an essential role in many novels and future applications, \eg,  autonomous driving or object detection.
While humans can easily capture a scene at a glance (\ie, perceive general and specific details), it is difficult for a machine.
Thus, semantic scene segmentation is one crucial step toward scene understanding, and it is one of the key challenges in computer vision.
In a broader view, semantic segmentation (SS)~\cite{Arnab2016} is a high-level task that makes complete scene understanding possible.
The goal of scene understanding is to make machines see like humans, \ie, to have a complete understanding of visual scenes.
SS aims to annotate each pixel of an image with a class label describing what this pixel represents.
SS task is also called dense prediction since each pixel in the image is classified.
How to extract and interpret the different levels of information details (\ie, global and local context information) is still an open problem and even more for a dense pixel-wise classification.
Note that in solving the SS task, this one directly influences different applications, for example, self-drive vehicles~\cite{Chen2016c, Zhou2019}, segmentation on X-ray~\cite{Bullock2019}, detect crown on dental X-ray~\cite{Wang2016}, brain tumor segmentation~\cite{Havaei2017, Pereira2019}, remote sensing~\cite{Sherrah2016, Volpi2017, Bokhovkin2019}, among others.

\begin{figure}[tb]
\centering
\newlength{\colfig}
\setlength{\colfig}{0.5pt}
{\renewcommand{\arraystretch}{0}
  \resizebox{\linewidth}{!}{%
  \begin{tabular}{%
      @{}%
      c@{\hspace{\colfig}}
      c@{\hspace{\colfig}}
      c@{}
    }
    \includegraphics[width=0.15\textwidth, height=.55in]{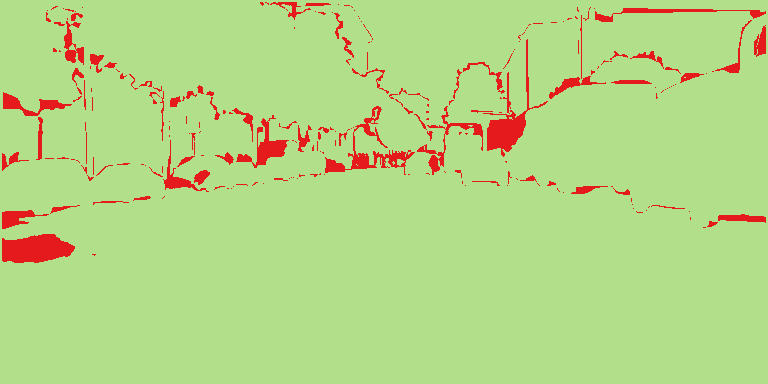}&
    \includegraphics[width=0.15\textwidth, height=.55in]{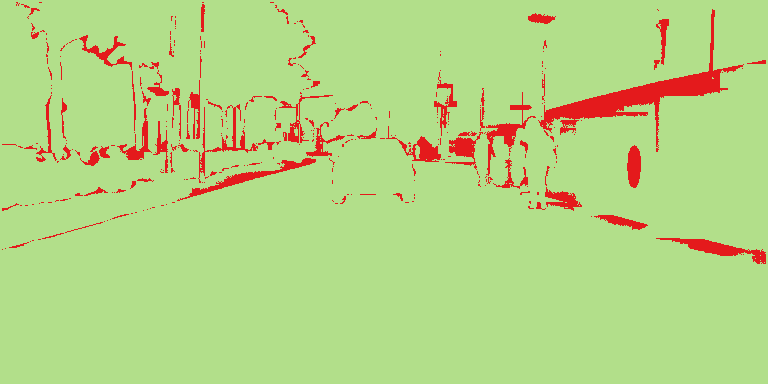}&
    \includegraphics[width=0.15\textwidth, height=.55in]{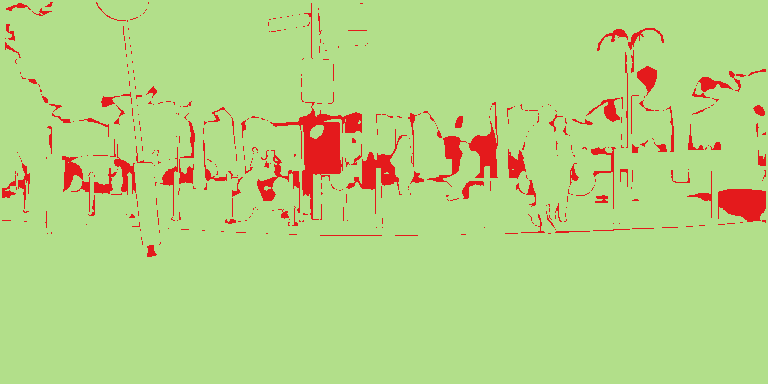}\\[0.1cm]
    \footnotesize (a) FastNet & \footnotesize (b) DUNet & \footnotesize (c) HRNet\\[0.2cm]
    \includegraphics[width=0.15\textwidth, height=.55in]{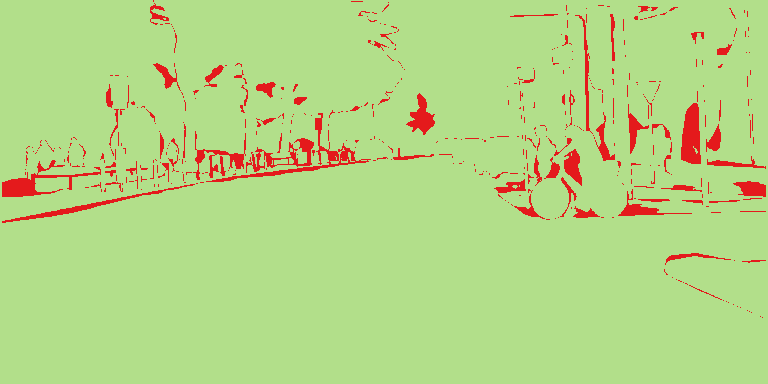}&
    \includegraphics[width=0.15\textwidth, height=.55in]{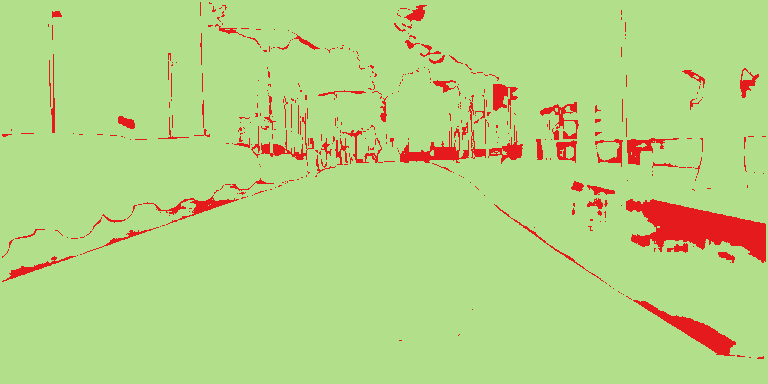}&
    \includegraphics[width=0.15\textwidth, height=.55in]{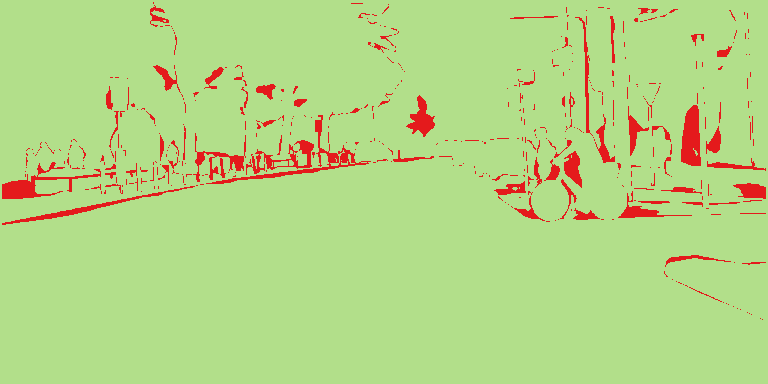}\\[0.1cm]
    \footnotesize (d) OCNet & \footnotesize (e) CCNet & \footnotesize (f) HRNet+OCR
  \end{tabular}
  }
}
\caption{The loss of spatial precision, often produced at the objects' boundary (red color), remains a problem in semantic segmentation.}
\label{fig:SS_problems}
\vspace{-6pt}
\end{figure}

Over the last decade, computer vision has benefited from significant developments in deep learning, particularly by convolutional neural networks (CNNs) that perform remarkably on complex vision tasks such as object classification and SS\@.
With the help of ever-growing amounts of labeled data, the deep networks can better generalize and model their context information than the traditional methods \cite{Thoma2016}.
Long \etal~\cite{Long2016} showed that CNNs could be adapted for the segmentation task by adding upsampling layers to perform pixel-wise predictions.
However, with the deep learning approach, new problems have been observed (Fig.~\ref{fig:SS_problems}) that derive from using CNNs for the specific task of semantic segmentation.
These problems are~\cite{Chen2017a, Lin2017}:
(i)~low-resolution obtained in the output of the CNNs, and
(ii)~loss of spatial precision of objects within the image.

These problems are not produced by a specific operation, \eg, downsampling, but rather by a set of operations.
For instance, the absence of reconstruction and refinement methods, the excessive down-sampling, the gradient vanishing, or the lack of a better extractor of feature maps~\cite{Saire2021}.
Nowadays, different models~\cite{Garcia-Garcia2018, Lateef2019, Hao2020, Minaee2021} have addressed these problems, especially in the low resolution on the output maps.
However, the loss of spatial precision problem persists and is commonly visualized at the edge of segmented objects, \cf Fig.~\ref{fig:SS_problems}.
In this work, we found (see Tables~\ref{tab:result_cityscape} and~\ref{tab:result_synthia}) that to preserve spatial precision, we need mechanisms that provide general and specific details to segment, \ie, methods for better extraction, modeling, and treatment of global and local context information.

Thus, some models~\cite{Zheng2015, Chen2017a, Jayasumana2020} use heatmap refinement with post-processing steps---\eg, conditional random field~\cite{Krahenbuhl2011}---as well as models adjusting the maps from bounding boxes~\cite{He2017a, Liu2018, Chen2018}.
Although post-processing helps to have a broader-view field of objects (\ie, global features), the models lack effective global feature extractors embedded in the architecture.
The encoder-decoder models~\cite{Ronneberger2015, Noh2015, Badrinarayanan2017} carry out more detailed work to adjust the heat maps by adding operations in the reconstruction, \ie, upsampling and deconvolution.
These models perform local and global feature extraction at some level.
However, they require a more robust combination of features than skip connections and concatenations.
Some models~\cite{Paszke2016, Lin2017a, Zhao2018, Tao2020} work with samples at different scales (multi-scale models) to obtain a full context of the images, \ie, global and local context information from upper and lower scales, respectively.
However, multiple-size inputs make the merging process more complicated than it needs to be.
In contrast, instead of resizing the inputs, models~\cite{Chen2017, Zhao2017, Chen2018a, Lian2021} increase the field of vision of the kernels (\ie, receptive field) through multiple dilated convolutions~\cite{Chen2017a}.
With the information from different scopes, they have the ability to tackle objects of varying sizes.
However, the sampling ranges distribution (\eg, global features and context priors) cannot ensure that the information can be contained in particular ranges.
That is, they have drawbacks with objects of bigger-size than the convolution kernel pyramid.
Finally, models~\cite{Tao2020, Jin2021a} are emerging, focusing on smart feature extraction through attention.
However, performing attention in limited regions of the image shows similar behavior to the previous models, \ie, larger-size objects escape the focus of attention.

We address the loss of spatial precision problem for SS tasks by including specific structures to extract local and global features (\ie, holistic and specific features extraction blocks).
Thus, we propose a Probabilistic Hourglass Gaussian Mixture Model (PHGMM), which combines Gaussian Mixture Model (GMM)~\cite{Reynolds2009, Bishop2006} by assuming that the data points are generated from a Mixture-of-Gaussians and a Variational AutoEncoder (VAE)~\cite{Kingma2014}.
Unlike the previous models, PHGMM learns two distributions by providing structures to the internal representations (\ie, latent spaces) of the SS task given an input.
The first distribution extracts the global features (\ie, global context) useful for coarse SS\@.
In contrast, the second distribution focuses on obtaining the specific features (\ie, local context) for a fine SS\@.
Furthermore, in contrast to the previous models, we use a single sample for the multi-scale features extraction; our distributions (\ie, our learning representation) extract information independently of the objects-size to segment, combining the information in one stage robust reconstruction.
Thus, we address the loss of spatial precision problem by combining the local and global features of both latent spaces and the spatial features of the encoder through skip connections and upsampling in the decoder.
Our main contributions are

\begin{itemize}
  \item an end-to-end trainable deep model, PHGMM, that combines VAE and GMM for the features' internal representations;
  \item two latent spaces modeled as a GMM to extract global and local context information that jointly improve the detection of the semantic classes (\ie, holistic and specific features extraction);
  \item pipeline for image reconstruction by merging the different scales of context information, recovering and improving the geometric information through the decoding stage; and
  \item a demonstration of the advantages of using coarse and fine modeling for the image information.
\end{itemize}

\makeatletter
\begin{figure*}[tb]
\centering
\resizebox{\linewidth}{!}{%
  \begin{tikzpicture}[
  node distance=1.cm,
  pic shift/.store in=\shiftcoord,
  pic shift={(0,0)},
  shift to anchor/.code={
    \tikz@scan@one@point\pgfutil@firstofone(.\tikz@anchor)\relax
    \pgfkeysalso{shift={(-\pgf@x,-\pgf@y)}}
  },
  pics/named scope code/.style 2 args={code={\tikz@fig@mustbenamed%
    \begin{scope}[pic shift, local bounding box/.expanded=\tikz@fig@name, #1]#2\end{scope}%
  }},
  layer/.style={draw, text width=4pt, minimum width=4pt, rounded corners, minimum height=#1},
  pics/encoder/.style = {named scope code={node distance=4pt}{%
    \node[layer=.8cm, fill=cs-encoder, fill opacity=0.4] (-l2) {};
    \node[layer=1.2cm, left=of -l2, fill=cs-encoder, fill opacity=0.4] (-l1) {};
    \node[layer=.4cm, right=of -l2, fill=cs-encoder, fill opacity=0.4] (-l3) {};
    \node[layer=1.6cm, left=of -l1, fill=cs-encoder, fill opacity=0.4] (-l0) {};
  }},
  pics/encoderpost/.style = {named scope code={node distance=4pt}{%
      \node[layer=.8cm] (-l2) {};
      \node[layer=1.2cm, left=of -l2] (-l1) {};
      \node[layer=.4cm, right=of -l2] (-l3) {};
      \node[layer=1.6cm, left=of -l1] (-l0) {};
    }},
  pics/decoder/.style = {named scope code={node distance=4pt}{%
    \node[layer=.8cm, fill=cs-decoder, fill opacity=0.4] (-l1) {};
    \node[layer=.4cm, left=of -l1, fill=cs-decoder, fill opacity=0.4] (-l0) {};
    \node[layer=1.2cm, right=of -l1, fill=cs-decoder, fill opacity=0.4] (-l2) {};
    \node[layer=1.6cm, right=of -l2, fill=cs-decoder, fill opacity=0.4] (-l3) {};
    \node[layer=1.6cm, right=12pt of -l3] (-l4) {};
    \draw [rounded corners] (-l4.north west) -- (-l4.south west) [sharp corners]-- (-l4.south) -- (-l4.north) [fill=black!50, rounded corners]-- cycle;
  }},
  pics/gmm/.style={named scope code={}{
    \draw[yscale=2.0,yshift=-3,xscale=0.1,domain=0:10] plot[id=gauss1]
    function{exp(-(x-5)*(x-5)/2/2/2.0)/sqrt(pi)/2};
    \draw[yscale=2.0,yshift=-3,xscale=0.1,domain=5:15] plot[id=gauss2]
    function{exp(-(x-10)*(x-10)/1/1/2.0)/sqrt(pi)/2};
    \draw[yscale=2.0,yshift=-3,xscale=0.1,domain=10:20] plot[id=gauss3]
    function{exp(-(x-15)*(x-15)/2/2/2.0)/sqrt(pi)/2};
  }},
  pics/gmm posterior/.style={named scope code={}{
    \draw[yscale=2.0,yshift=-3,xscale=0.1,domain=0:8] plot[id=gauss4]
    function{exp(-(x-4)*(x-4)/1.25/1.25/2.0)/sqrt(pi)/2};
    \draw[yscale=2.0,yshift=-3,xscale=0.1,domain=5:11] plot[id=gauss5]
    function{exp(-(x-8)*(x-8)/.5/.5/2.0)/sqrt(pi)/2};
    \draw[yscale=2.0,yshift=-3,xscale=0.1,domain=7:20] plot[id=gauss6]
    function{exp(-(x-13)*(x-13)/3/3/2.0)/sqrt(pi)/2};
  }},
  pics/gauss/.style={named scope code={}{
    \draw[yscale=2.0,yshift=-3,xscale=0.1,domain=0:10] plot[id=gauss]
    function{exp(-(x-5)*(x-5)/2/2/2.0)/sqrt(pi)/2};
  }},
  label/.style={
    node distance=5pt,
    font=\footnotesize,
  },
  edge/.style={
    >=Latex,
    ->,
    rounded corners,
    shorten >= 5pt,
    shorten <= 5pt,
  },
  thick/.style={
    >=Latex,
    -,
    rounded corners,
    shorten >= 5pt,
    shorten <= 5pt,
  },
  loss/.style={
    dashed,
    draw,
    black!75,
    font=\scriptsize,
  },
  ]

  \definecolor{cs-encoder}{RGB}{255, 127, 0}
  \definecolor{cs-decoder}{RGB}{202,178,214}
  \definecolor{cs-localz}{RGB}{166,206,227}
  \definecolor{cs-globalg}{RGB}{178,223,138}

  \node (img) {\includegraphics[width=2.5cm]{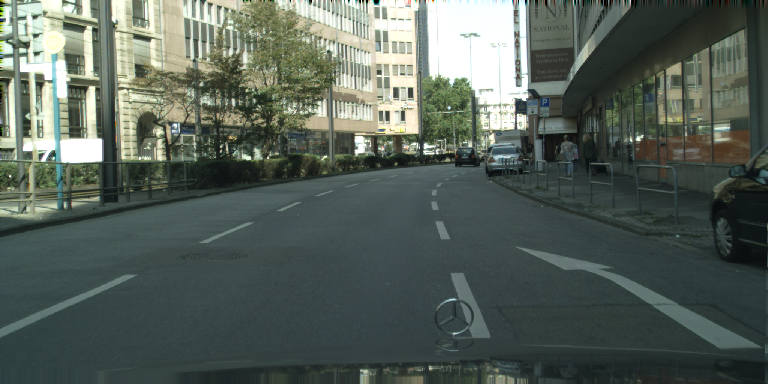}};

  \draw pic[right=2.5cm of img] (enc) {encoder};
  \draw pic[below right=1.25cm and -25pt of enc] (enc2) {encoderpost};

  \newsavebox{\tbox}
  \node[right=of enc] (gmm0) {
    \usebox{\tbox{\includegraphics[width=2.0cm]{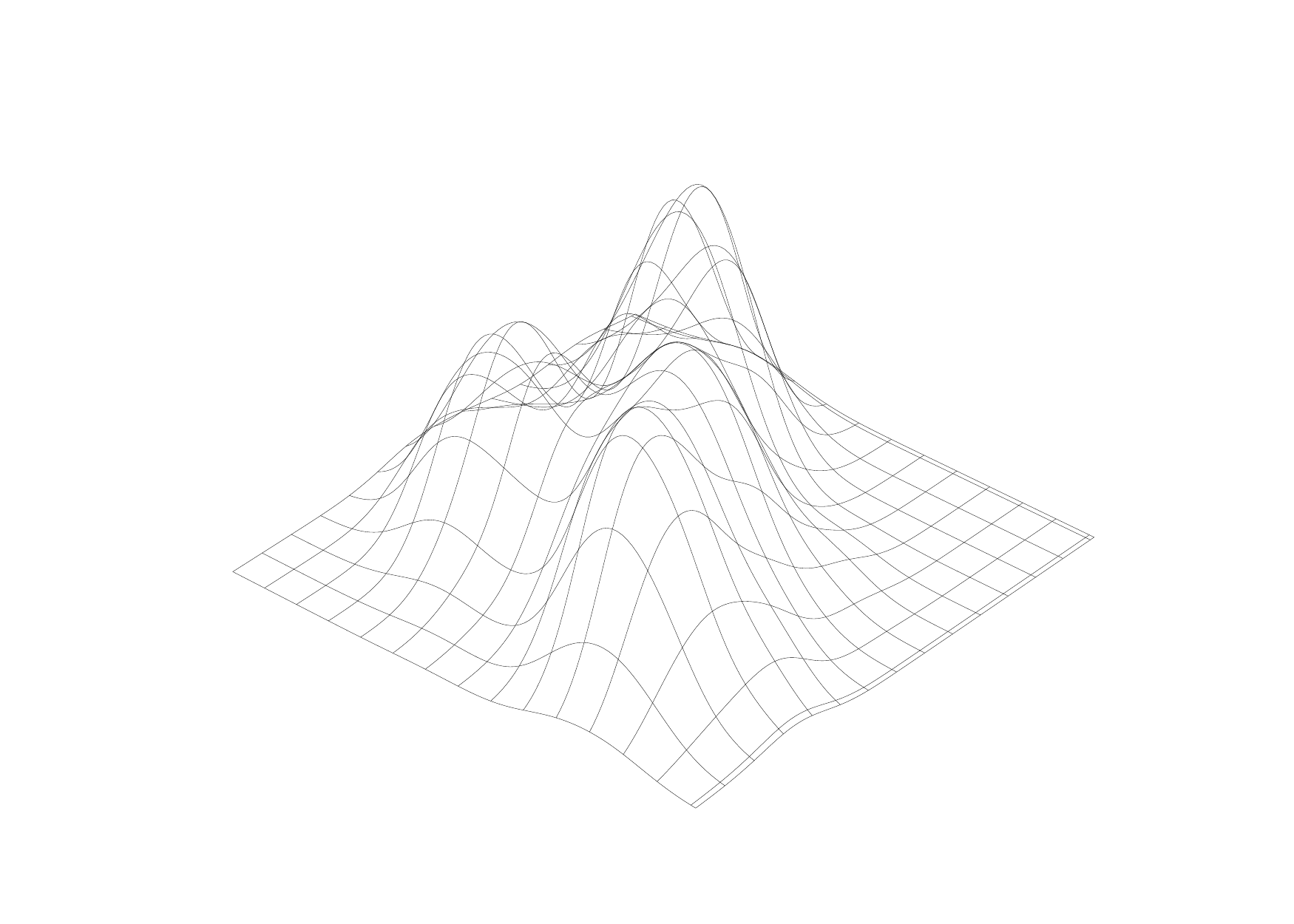}}}
  };
  \node[fit=(gmm0), inner sep=-2mm, draw, rounded corners, fill=cs-localz, fill opacity=0.2] (gmm) {};

  \node[right=of enc2] (gmm3) {
    \usebox{\tbox{\includegraphics[width=2.0cm]{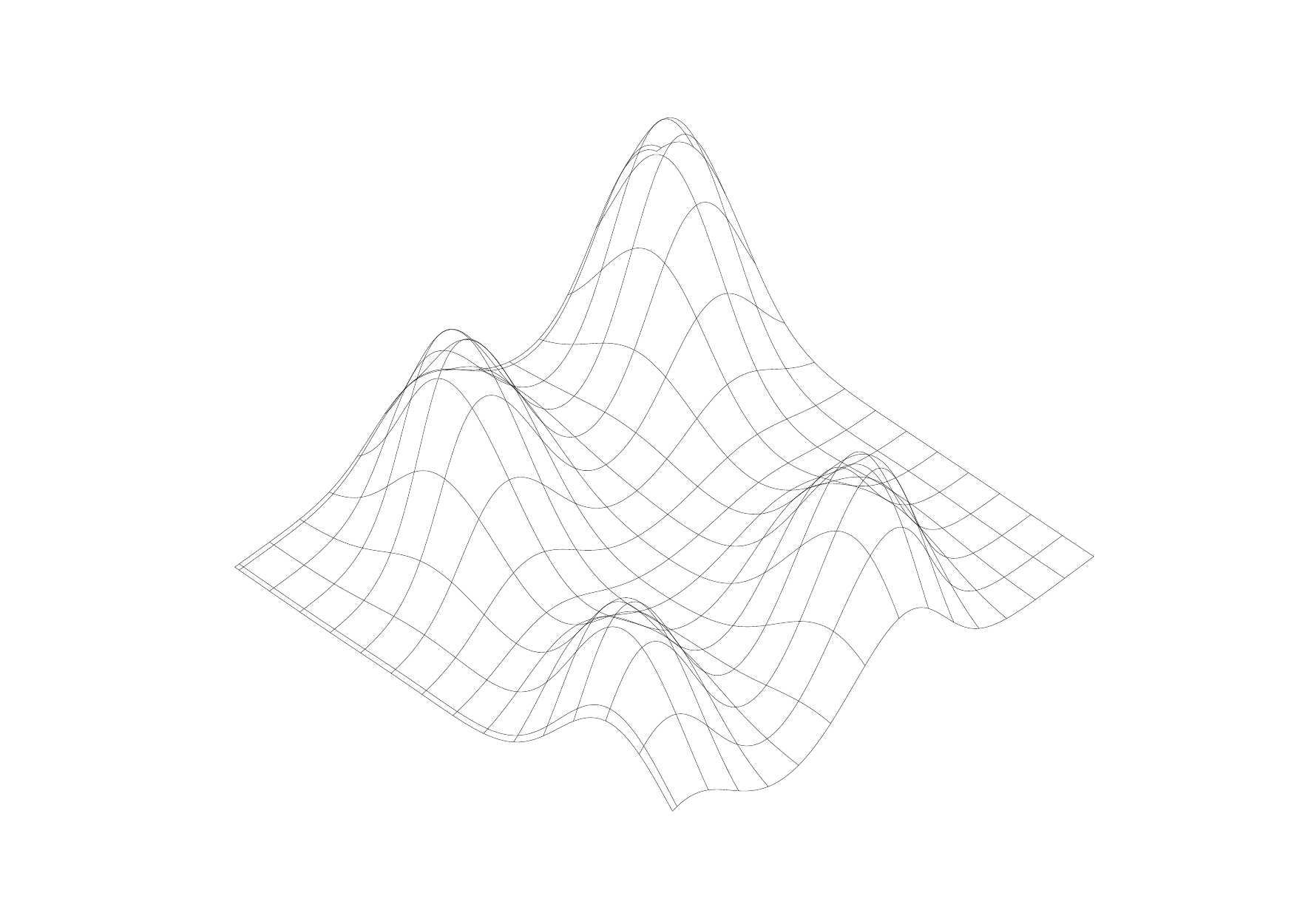}}}
  };
  \node[fit=(gmm3), inner sep=-2mm, draw, rounded corners] (gmm2) {};

  \node[above=of gmm] (gauss0) {
    \usebox{\tbox{\includegraphics[width=2.0cm]{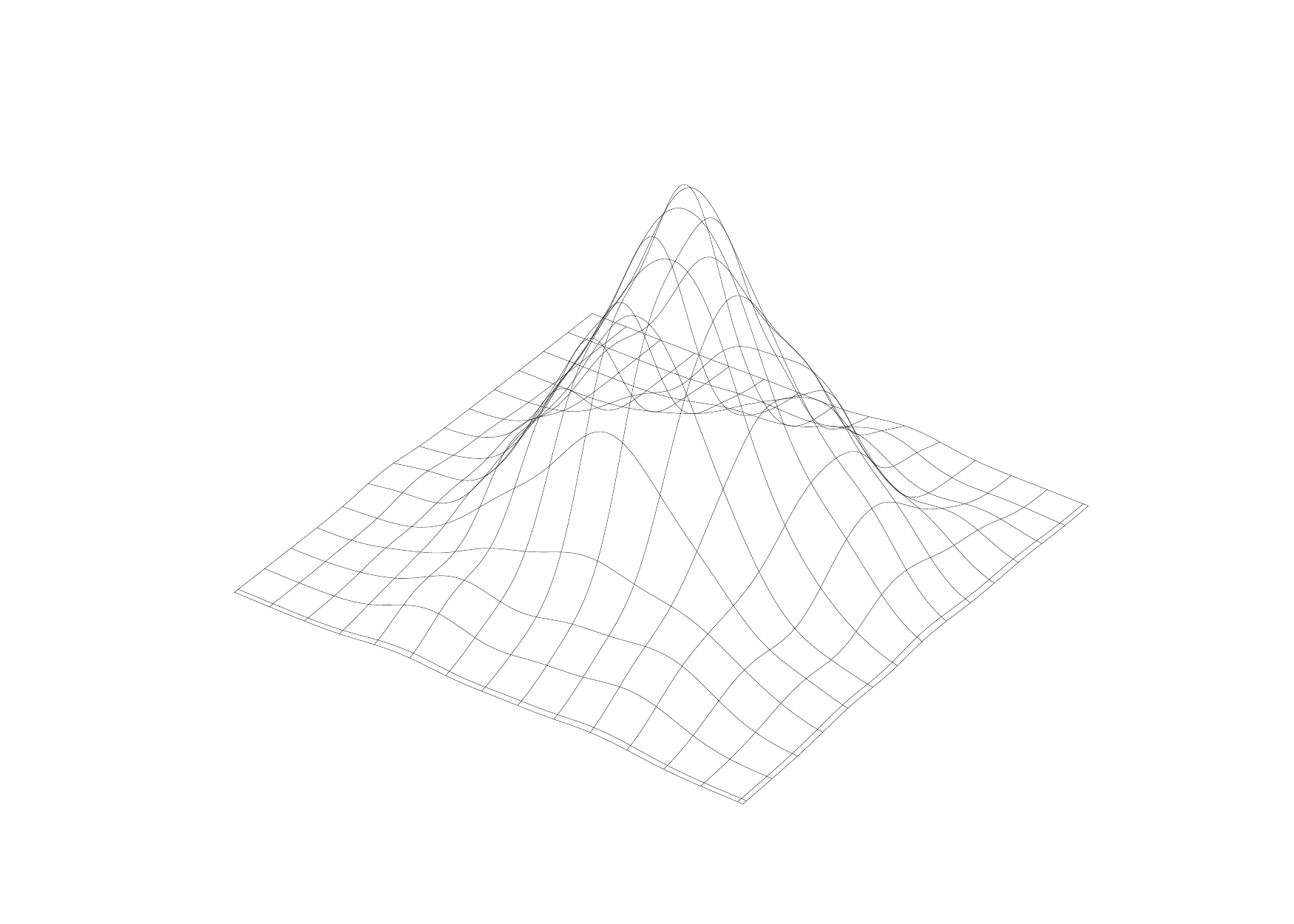}}}
  };
  \node[fit=(gauss0), inner sep=-2mm, draw, rounded corners, fill=cs-globalg, fill opacity=0.2] (gauss) {};

  \pic[right=1.5cm of gmm] (dec) {decoder};

  \node[right=of dec] (pred) {\includegraphics[width=2.5cm]{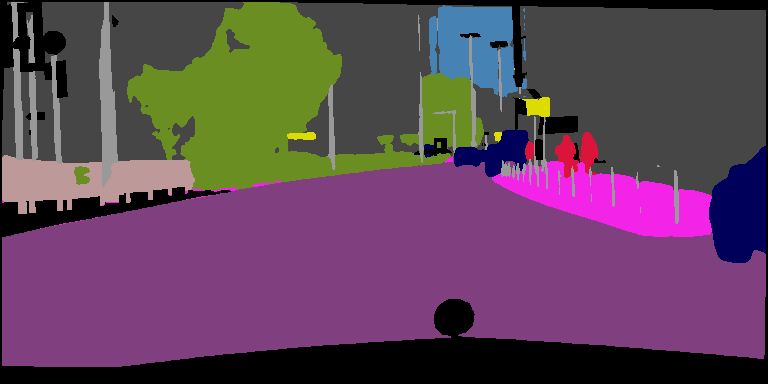}};
  \node[right=.25cm of pred] (gt) {\includegraphics[width=2.5cm]{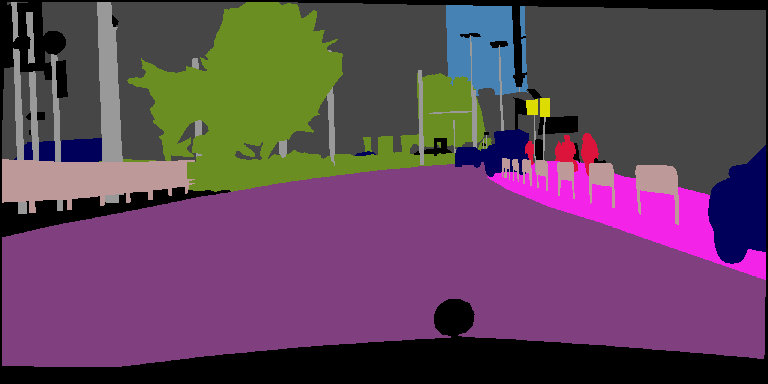}};

  \node[label, above=of enc] (enc-lbl) {Encoder};
  \node[label, below=of img] (img-lbl) {Input};

  \node[label, anchor=west] at (enc-lbl -| dec-l0) {Decoder ($f$)};
  \node[label, below=of pred] (pred-lbl) {Prediction};
  \node[label, below=of gt] (gt-lbl) {Ground Truth};

  \node[label, below=of enc2] (enc2-lbl) {PostNet};
  \node[label] at (enc2-lbl -| gmm2) {Posterior};

  \node[label, below=0.1cm of gauss] (gauss-lbl) {Global space ($g$)};

  \draw[edge] (enc-l0.north) -| ($(enc-l3.east)+(5pt,-3pt)$);
  \draw[edge] (enc-l1.north) -| ($(enc-l3.east)+(5pt,-3pt)$);
  \draw[edge] (enc-l2.north) -| ($(enc-l3.east)+(5pt,-3pt)$);

  \draw[edge] (enc2-l0.north) -| ($(enc2-l3.east)+(5pt,-3pt)$);
  \draw[edge] (enc2-l1.north) -| ($(enc2-l3.east)+(5pt,-3pt)$);
  \draw[edge] (enc2-l2.north) -| ($(enc2-l3.east)+(5pt,-3pt)$);

  \draw[edge] (img) -- (enc);
  \draw[edge] (img-lbl) |- ($(enc2.west)+(0,5pt)$);

  \draw[edge] (enc) -- (gmm);
  \draw[edge] (enc2) -- (gmm2);

  \draw[edge] (gmm) -- (dec);
  \draw[edge] (dec) -- (pred);

  \draw[edge] (enc.east) -- ++(.5cm, .0cm) |- (gauss);
  \draw[edge] (gauss) -| ($(dec-l4.north)$);

  \coordinate (p-enc2) at ($(enc2-lbl.south)+(0,-5pt)$);
  \draw[edge] (gt-lbl) |- (p-enc2) -- (p-enc2 -| img) |- ($(enc2.west)+(0,-5pt)$);

  \draw[thick, loss] ($(dec-l0.south)+(5pt,4pt)$) |- ($(dec-l3.south)+(15pt,-20pt)$);
  \draw[thick, loss] ($(dec-l1.south)+(5pt,4pt)$) |- ($(dec-l3.south)+(-15pt,-20pt)$);
  \draw[thick, loss] ($(dec-l2.south)+(5pt,4pt)$) |- ($(dec-l3.south)+(15pt,-20pt)$);
  \draw[edge, loss] ($(dec-l3.south)+(5pt,-20pt)$) -| ($(dec-l3.east)+(13pt,-20pt)$);

  \draw[thick] ($(enc-l2.south)+(0,3pt)$) |- ($(gmm.south)+(5pt,-5pt)$);
  \draw[edge] ($(gmm.south)+(-5pt,-5pt)$) -| ($(dec-l1.south)+(-5pt,3pt)$);

  \draw[thick] ($(enc-l1.south)+(0,3pt)$) |- ($(gmm.south)+(5pt,-10pt)$);
  \draw[edge] ($(gmm.south)+(-5pt,-10pt)$) -| ($(dec-l2.south)+(-5pt,3pt)$);

  \draw[thick] ($(enc-l0.south)+(0,3pt)$) |- ($(gmm.south)+(5pt,-15pt)$);
  \draw[edge] ($(gmm.south)+(-5pt,-15pt)$) -| ($(dec-l3.south)+(-5pt,3pt)$);

  \node[label, fill=white, fill opacity=.75, text opacity=1, rounded corners] (gmm-lbl) at (img-lbl -| gmm) {Local space ($z$)};
  \node[label, fill=white, fill opacity=.75, text opacity=1, rounded corners] at (enc-lbl -| dec-l4.north) {Merger $(f_m)$};

  \node[circle, loss, inner sep=1pt] (ls) at ($(pred.north)!.5!(gt.north)+(0pt,15pt)$) {$\ell_s$};
  \node[circle, loss, inner sep=1pt] (lz) at ($(gmm.east)!.5!(gmm2.east)+(15pt,-10pt)$) {$\ell_z$};

  \node[circle, draw] (prior) at (enc |-gauss) {$\mathcal{N}$};
  \node[label, left=of prior] (prior-lbl) {Prior};

  \node[circle, loss, inner sep=1pt] (lg) at ($(gauss.north)!.5!(prior.north)+(0,20pt)$) {$\ell_g$};

  \node[circle, loss, inner sep=1pt] (lz) at ($(gmm.east)!.5!(gmm2.east)+(15pt,-10pt)$) {$\ell_z$};

  \draw[edge, loss] (ls) -| (pred);
  \draw[edge, loss] (ls) -| (gt);

  \draw[edge, loss] (lz) |- ($(gmm.south east)+(0,5pt)$);
  \draw[edge, loss] (lz) |- ($(gmm2.north east)+(0,-5pt)$);

  \draw[edge, loss] (lg) -| (prior);
  \draw[edge, loss] (lg) -| (gauss);

  \pgfresetboundingbox
  \path[use as bounding box] (img.west |- lg.north) rectangle (gt.east |-p-enc2);
  \end{tikzpicture}}
\caption{Given an input image, we extract local, $z$, and global, $g$, features (orange box) to improve the prediction of the semantic classes.
  We model the local features as GMM that fit a predicted posterior from the data (light blue box).
  For the global features, we use a single Gaussian to hold the holistic information of the image (green box).
  Note that we managed to recover the local features at the decoder stage by combining the GMMs of the latent space $z$ with the encoder features through skip connections and deconvolution operations (purple box).
  Finally, the whole process is regularized by making the predictions for matching the ground truth, $\ell_s$, and making the local and global spaces similar to their respective priors, $\ell_z$, and $\ell_g$, respectively.}
\label{fig:PHGMMPos_train}
\end{figure*}
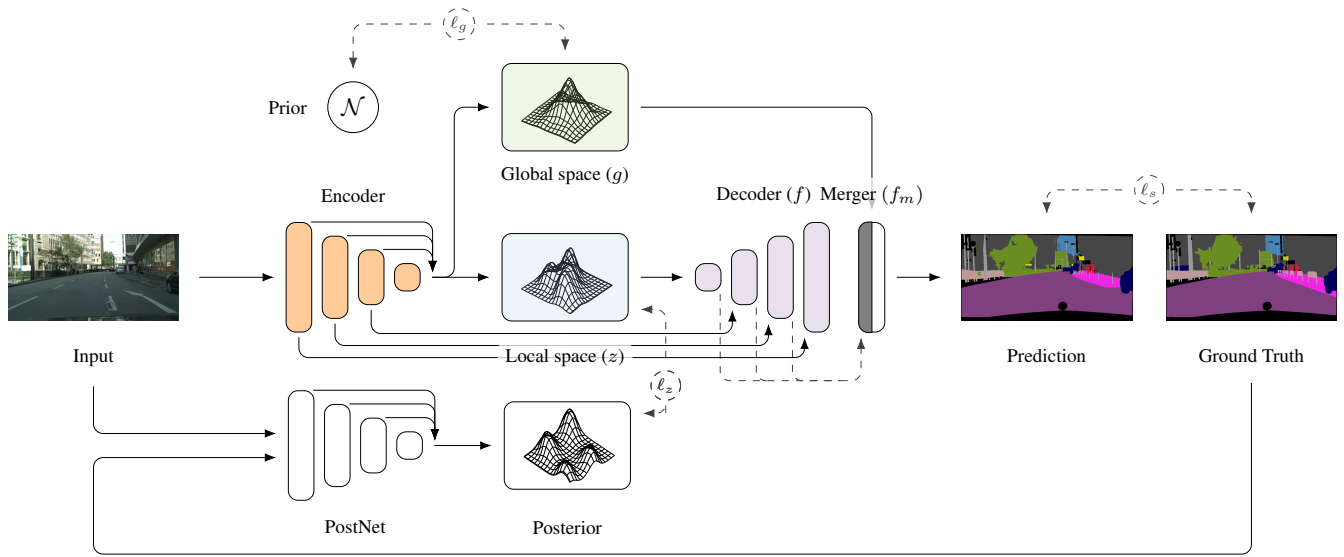
\makeatother

\section{Related work}
\label{sec:related_work}
Semantic segmentation aims at dense pixel-wise classification, \ie, assign labels to each pixel in fine-grained fashion.
It is repeatedly used as an intermediate step in computer vision applications by facilitating high-level image processing and analysis.
Currently, the deep segmentation approaches based on Fully Convolutional Network (FCN) have made remarkable progress.
However, this architecture is inherently limited to local receptive fields and short-range contextual information.
Due to this shortage of contextual information, the FCN model suffers from two main drawbacks: low-resolution output maps and a loss in spatial precision.

Existing models~\cite{Zheng2015, Chen2017a, Vemulapalli2016} present the first effort to address these problems by refining the output maps using Conditional Random Field (CRF)~\cite{Krahenbuhl2011}.
However, most of the models that perform post-processing steps lack local and global feature extractors that are efficient enough to discriminate the object boundaries.
In contrast, instead of performing holistic post-processing, other models~\cite{He2017a, Liu2018, Chen2018} focus on the refinement from bounding boxes, \ie, they obtain a segmentation heat map in each region of interest.

However, the transfer of information (context) remains limited.
Recent works~\cite{Chen2017, Zhao2017, Huang2020, Saire2021} verify the importance of context information in SS\@.
How to extract, merge, and employ this context information is the next step in SS\@.
Thus, encoder-decoder models, such as U-Net~\cite{Ronneberger2015}, DeconvNet~\cite{Noh2015}, SegNet~\cite{Badrinarayanan2017}, or DUNet~\cite{Jin2019}, add a decoder stage (\ie, a set of deconvolution and unpooling operations) with skip connection, managing to merge, at some level, local context from low-level features and global context from high-level ones.
For instance, ParseNet~\cite{Liu2015} adds global information to layers through the global average pooling operation, and FC-DenseNet~\cite{Jegou2017} combines knowledge by concatenating all the features of the previous output layers.
Although the encoder-decoder models show to focus on details, they still require additional detailed connections to improve their information reconstruction stage.

\makeatletter
\begin{figure*}[tb]
\centering
\resizebox{\linewidth}{!}{%
\begin{tikzpicture}[
scale=1,
every node/.style={scale=0.8},
node distance=1.25cm,
pic shift/.store in=\shiftcoord,
pic shift={(0, 0)},
shift to anchor/.code={
  \tikz@scan@one@point\pgfutil@firstofone(.\tikz@anchor)\relax
  \pgfkeysalso{shift={(-\pgf@x, -\pgf@y)}}
},
pics/named scope code/.style 2 args={code={\tikz@fig@mustbenamed%
  \begin{scope}[pic shift, local bounding box/.expanded=\tikz@fig@name, #1]#2\end{scope}%
}},
layer/.style={draw, text width=4pt, minimum width=4pt, rounded corners, minimum height=#1},
pics/encoder2/.style={named scope code={node distance=4pt}{%
  \node[layer=.8cm, fill=cs-encoder, fill opacity=0.4] (-l2) {};
  \node[layer=1.2cm, above left=of -l2, fill=cs-encoder, fill opacity=0.4] (-l1) {};
  \node[layer=.4cm, below right=of -l2, fill=cs-encoder, fill opacity=0.4] (-l3) {};
  \node[layer=1.6cm, above left=of -l1, fill=cs-encoder, fill opacity=0.4] (-l0) {};
}},
label/.style={
  node distance=5pt,
  font=\tiny,
},
edge/.style={
  >=Latex,
  ->,
  rounded corners,
  shorten >= 5pt,
  shorten <= 5pt,
},
linea/.style={
  >=Latex,
  rounded corners,
  shorten >= 5pt,
  shorten <= 5pt,
},
]

\newcommand{\gaussellipse}[5]{{
  \foreach\i in {0, 0.1, ..., 0.5} {
    \pgfmathsetmacro{\radx}{#3-\i}
    \pgfmathsetmacro{\rady}{#4-\i}
    \fill[#5, opacity=\i, rotate around={#2:#1}] #1 ellipse ({\radx} and {\rady});
  }
}}

\definecolor{cs-sky}{RGB}{70, 130, 180}
\definecolor{cs-build}{RGB}{70, 70, 70}
\definecolor{cs-road}{RGB}{128, 64, 128}
\definecolor{cs-sidewalk}{RGB}{244, 35, 232}
\definecolor{cs-fence}{RGB}{190, 153, 153}
\definecolor{cs-vege}{RGB}{107, 142, 35}
\definecolor{cs-pole}{RGB}{153, 153, 153}
\definecolor{cs-car}{RGB}{0, 0, 142}
\definecolor{cs-sign}{RGB}{220, 220, 0}
\definecolor{cs-person}{RGB}{220, 20, 60}
\definecolor{cs-cyclist}{RGB}{119, 11, 32}

\definecolor{cs-encoder}{RGB}{255, 127, 0}
\definecolor{cs-localz}{RGB}{166,206,227}
\definecolor{cs-globalg}{RGB}{178,223,138}

\draw pic[] (enc3) {encoder2};

\node[fit={($(enc3-l0.north)+(0pt, 0pt)$) ($(enc3-l3.south)+(385pt, -48pt)$)}, right=0.5cm of enc3, inner sep=+2mm, draw, rounded corners, fill=cs-localz, fill opacity=0.2] (localspace) {};

\node[layer=1.6cm, right=2.2cm of enc3-l0.east, fill=white] (convsig0) {};
\node[layer=1.2cm, below=0.1cm of convsig0, fill=white] (convsig1) {};
\node[layer=0.8cm, below=0.1cm of convsig1, fill=white] (convsig2) {};
\node[layer=0.4cm, below=0.1cm of convsig2, fill=white] (convsig3) {};

\node[layer=1.6cm, right=0.8cm of convsig1, fill=white] (gap) {};
\draw[linea] ($(gap.west)+(-5pt, 11pt)$) -- ($(gap.east)+(5pt, 11pt)$);
\draw[linea] ($(gap.west)+(-5pt, 0)$) -- ($(gap.east)+(5pt, 0)$);
\draw[linea] ($(gap.west)+(-5pt, -11pt)$) -- ($(gap.east)+(5pt, -11pt)$);;

\gaussellipse{(5, 2)}{60}{0.5}{0.8}{cs-sky} (gaussgmm0);
\gaussellipse{(5.3, 1)}{20}{0.8}{0.6}{cs-vege} (gaussgmm1)
\gaussellipse{(4, 1)}{40}{0.5}{0.9}{cs-cyclist} (gaussgmm2)
\gaussellipse{(4, 0)}{0}{0.8}{0.5}{cs-sign} (gaussgmm3)
\gaussellipse{(4.2, 2)}{0}{0.5}{0.5}{cs-person} (gaussgmm4)
\gaussellipse{(5.3, 0.1)}{10}{0.6}{0.7}{cs-car} (gaussgmm5)

\draw[edge] ($(gap.east)+(82pt, 0pt)$) -- ($(gap.east)+(103pt, 0pt)$);
\draw[linea] ($(gap.east)+(99pt, -40pt)$) -- ($(gap.east)+(99pt, 40pt)$);
\draw[linea] ($(gap.east)+(94pt, 35pt)$) -- ($(gap.east)+(107pt, 35pt)$);
\draw[linea] ($(gap.east)+(113pt, 35pt)$) -- ($(gap.east)+(126pt, 35pt)$);
\node (mupi0) at ($(gap.east)+(110pt, 30pt)$) [cs-road, label] {$\mu_z^0 \times \pi_z^0$};
\node (mupi1) at ($(gap.east)+(110pt, 15pt)$) [cs-vege, label] {$\mu_z^1 \times \pi_z^1$};
\node (mupi2) at ($(gap.east)+(110pt, 0pt)$) [cs-person, label, opacity=0.6] {$\mu_z^2 \times \pi_z^2$};
\node (mupidots) at ($(gap.east)+(110pt, -15pt)$) [label] {$\vdots$};
\node (mupik) at ($(gap.east)+(110pt, -30pt)$) [cs-car, label, opacity=0.8] {$\mu_z^k \times \pi_z^k$};
\draw[linea] ($(gap.east)+(121pt, -40pt)$) -- ($(gap.east)+(121pt, 40pt)$);
\draw[linea] ($(gap.east)+(94pt, -35pt)$) -- ($(gap.east)+(107pt, -35pt)$);
\draw[linea] ($(gap.east)+(113pt, -35pt)$) -- ($(gap.east)+(126pt, -35pt)$);

\node[cs-road, layer=0.8cm, rotate=90, right=1.6cm of mupi0.south east, fill=white] (proc-gmm0) {};
\node[cs-vege, layer=0.8cm, rotate=90, right=1.6cm of mupi1.south east, fill=white] (proc-gmm1) {};
\node[cs-person, opacity=0.6, layer=0.8cm, rotate=90, right=1.6cm of mupi2.south east, fill=white] (proc-gmm2) {};
\node[cs-car, opacity=0.8, layer=0.8cm, rotate=90, right=1.6cm of mupik.south east, fill=white] (proc-gmmk) {};

\node[cs-road, layer=0.8cm, rotate=90, right=3cm of mupi0.south east, fill=white] (proc-gmm-up0) {};
\node[cs-vege, layer=0.8cm, rotate=90, right=3cm of mupi1.south east, fill=white] (proc-gmm-up1) {};
\node[cs-person, opacity=0.6, layer=0.8cm, rotate=90, right=3cm of mupi2.south east, fill=white] (proc-gmm-up2) {};
\node[cs-car, opacity=0.8, layer=0.8cm, rotate=90, right=3cm of mupik.south east, fill=white] (proc-gmm-upk) {};

\node[layer=1.6cm, right=4cm of mupi2.east, fill=white] (concat) {};
\draw[linea] ($(concat.west)+(-5pt, 14pt)$) -- ($(concat.east)+(5pt, 14pt)$);
\draw[linea] ($(concat.west)+(-5pt, 5pt)$) -- ($(concat.east)+(5pt, 5pt)$);
\draw[linea] ($(concat.west)+(-5pt, -5pt)$) -- ($(concat.east)+(5pt, -5pt)$);
\node (dotsconcat) at ($(concat.west)+(5pt, -7.2pt)$) [label] {$\vdots$};
\draw[linea] ($(concat.west)+(-5pt, -14pt)$) -- ($(concat.east)+(5pt, -14pt)$);

\node[layer=1.6cm, right=0.7cm of concat, fill=white] (proc-concat) {};

\draw[edge] (enc3-l0) -- node[label, below]{conv $+$ sig} ($(convsig0.west)+(3pt, 0)$);
\draw[edge] (enc3-l1) -- node[label, below]{conv $+$ sig} ($(convsig1.west)+(3pt, 0)$);
\draw[edge] (enc3-l2) -- node[label, below]{conv $+$ sig} ($(convsig2.west)+(3pt, 0)$);
\draw[edge] (enc3-l3) -- ($(convsig3.west)+(3pt, 0)$);

\draw[edge] (convsig0) -| node[label, above]{GAP} (gap.north);
\draw[edge] ($(convsig1.east)+(0, 5pt)$) -- node[label, above]{GAP} ($(gap.west)+(0, 5pt)$);
\draw[edge] (convsig2.east) -- node[label, below right]{GAP} ++(10pt,0) |-  ($(gap.west)+(0, -5pt)$);
\draw[edge] (convsig3) -| node[label, below]{GAP} ($(gap.south)+(0, -5pt)$); %
\draw[edge] ($(gap.east)+(-2pt, 0)$) -- node[label, above]{FC} ($(gap.east)+(18pt, 0)$);

\draw[edge] ($(mupi0.east)+(0, -2pt)$) -- node[label, below]{FC $+$ relu} (proc-gmm0);
\draw[edge] ($(mupi1.east)+(0, -2pt)$) -- node[label, below]{FC $+$ relu} (proc-gmm1);
\draw[edge] ($(mupi2.east)+(0, -2pt)$) -- node[label, below]{FC $+$ relu} (proc-gmm2);
\draw[edge] ($(mupik.east)+(0, -2pt)$) -- node[label, below]{FC $+$ relu} (proc-gmmk);

\draw[edge] (proc-gmm0) -- node[label, below]{up} (proc-gmm-up0);
\draw[edge] (proc-gmm1) -- node[label, below]{up} (proc-gmm-up1);
\draw[edge] (proc-gmm2) -- node[label, below]{up} (proc-gmm-up2);
\draw[edge] (proc-gmmk) -- node[label, below]{up} (proc-gmm-upk);

\draw[edge] (proc-gmm-up0.south) -| node[label, above]{} ($(concat.north)+(0, -5pt)$);
\draw[edge] ($(proc-gmm-up1.south)+(0, -1pt)$) -- node[label, above]{} ($(concat)+(-2pt, 12pt)$);
\draw[edge] ($(proc-gmm-up2.south)+(0, -1pt)$) -- node[label, above]{} ($(concat)+(-2pt, -2pt)$);
\draw[edge] (proc-gmm-upk.south) -| node[label, below]{} ($(concat.south)+(0, 5pt)$);

\draw[edge] (concat) -- node[label, below]{conv} (proc-concat);
\draw[edge] ($(proc-concat.east)+(-3pt, 0)$) -- ($(proc-concat.east)+(30pt, 0)$);

\node[label] at (5, 2.1) {$\mu_z^0, \sigma_z^0, \pi_z^0$};
\node[label] at (5.3, 1) {$\mu_z^1, \sigma_z^1, \pi_z^1$};
\node[label] at (4, 1) {$\mu_z^2, \sigma_z^2, \pi_z^2$};
\node[label] at (4.2, 1.8) {$\mu_z^2, \sigma_z^2, \pi_z^2$};
\node[label] at (4, -0.1) {$\mu_z^{k-1}, \sigma_z^{k-1}, \pi_z^{k-1}$};
\node[label] at (5.3, 0.2) {$\mu_z^k, \sigma_z^k, \pi_z^k$};
\node[label] at ($(proc-concat.north)+(-20pt, 40pt)$) {\scriptsize Local Space ($z$)};

\node[label] at ($(enc3-l0.west)+(-17pt, 0pt)$) {$\frac{h}{4} \times \frac{w}{4} \times 256$};
\node[label] at ($(enc3-l1.west)+(-17pt, 0pt)$) {$\frac{h}{8} \times \frac{w}{8} \times 512$};
\node[label] at ($(enc3-l2.west)+(-20pt, 0pt)$) {$\frac{h}{16} \times \frac{w}{16} \times 1024$};
\node[label] at ($(enc3-l3.west)+(-20pt, 0pt)$) {$\frac{h}{32} \times \frac{w}{32} \times 2048$};

\node[label] at ($(convsig0.north)+(0pt, 5pt)$) {$\frac{h}{4} \times \frac{w}{4} \times C$};
\node[label] at ($(convsig3.south)+(0pt, -5pt)$) {$\frac{h}{32} \times \frac{w}{32} \times C$};
\node[label] at ($(gap.south)+(0pt, -5pt)$) {$1 \times 256$};

\node[label] at ($(gap.south east)+(50pt, -30pt)$) {Latent Space z};
\node[label] at ($(gap.south east)+(50pt, -37pt)$) {\scriptsize $q_\theta(z | x)$};

\node[label] at ($(5.3, 0.1)+(45pt, -17pt)$) {$K \times 1 \times 256$};
\node[label] at ($(proc-gmm2.center)+(0pt, -10pt)$) {$1 \times 1 \times D$};
\node[label] at ($(proc-gmmk.center)+(0pt, -10pt)$) {$1 \times 1 \times D$};
\node[label] at ($(proc-gmm-up2.center)+(0pt, -10pt)$) {$\frac{h}{32} \times \frac{w}{32} \times D$};
\node[label] at ($(proc-gmm-upk.center)+(0pt, -10pt)$) {$\frac{h}{32} \times \frac{w}{32} \times D$};
\node[label] at ($(concat.north)+(0pt, 15pt)$) {$\frac{h}{32} \times \frac{w}{32} \times [K \times D]$};
\node[label] at ($(proc-concat.south)+(0pt, -7pt)$) {$\frac{h}{32} \times \frac{w}{32} \times 512$};

\end{tikzpicture}
}
\caption{
Local Space diagram.
Note, the blocks and arrows represent the feature vectors and operations, respectively.
We extract features at different scales through convolution operations with a sigmoid activation function.
Next, we use the Global Average Pooling operation (GAP) to summarize the information and gather it into a concatenated feature vector.
To model the latent space $z$ through a GMM, we use the Fully Connected layer (FC) to extract $K$ different Gaussian mixtures.
Next, the mean of each Gaussian is used and stored in a list, and apply an FC with ReLU is to do further feature processing.
Finally, we return to our initial vector size (\ie, $\frac{h}{32} \times \frac{w}{32}$).
We concatenate the $K$ Gaussians processed and perform the last convolution operation to fuse all the information from the different Gaussians.
Although we compress the information using GAP to model the Gaussians distribution, the information contained in the latent space $z$ plays an important role in adequately recovering the geometric information (background and contour of the segmented objects) through the decoder stage.
}
\label{fig:PHGMM_localspace}
\end{figure*}
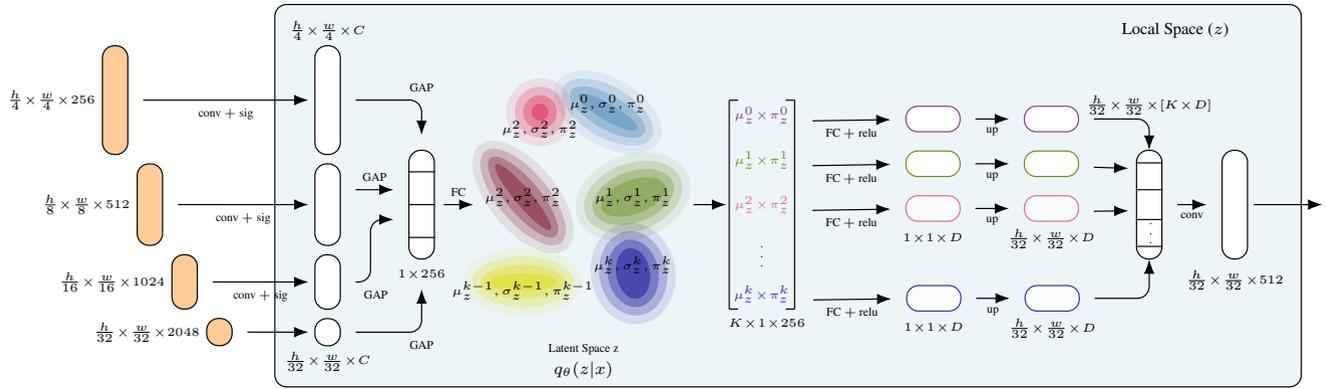
\makeatother

Although theoretically, in the layers with lower resolution (\eg, size of $H / 32 \times W / 32$), there is a greater receptive field; empirical experiments showed that these fields are significantly smaller and are not enough to capture global context information (\ie, global features)~\cite{Liu2015, Zhao2017}.
One way to address these drawbacks is through multi-scale models~\cite{Lin2017a, Li2019, Tao2020} (\eg, ICNet~\cite{Zhao2018} is fed with different input sizes).
DeepLabv2~\cite{Chen2017a}, DeepLabv3~\cite{Chen2017}, DeepLabv3++~\cite{Chen2018a}, DecoupleSegNets~\cite{Li2020a} and Gated-SCNN~\cite{Takikawa2019a} preserve the spatial size of the features maps by proposing Atrous Spatial Pyramid Pooling (ASSP).
ASPP employs a set of atrous convolutions operations~\cite{Yu2016} and average pooling to capture several context information (from local to global features).
However, dilated convolutions can cause grinding problems~\cite{Wang2018,Wang2021}.
It can induce a loss of local information in the models and capture irrelevant information on a large scale
attention models:
A different approach is presented by PSPNet~\cite{Zhao2017} and ESPNet~\cite{Mehta2019}, operating not on the convolution kernels but on  sub-regions of the feature maps generated at different levels.
They use the Pyramid Pooling Module (PMM) to get global and local context information by reducing the spatial size of features.
Also, HRNetv2~\cite{Wang2020} and HRNet+OCR~\cite{Yuan2020} perform multi-scale feature extraction by sharing feature maps across different branches (scales), \ie, broadcasting context information at various resolutions.
Nevertheless, how and where to combine multi-scale information still represents a challenge.

Although the previous context fusion models help capture different scales' features, The relationship between objects in a global view (essential to SS) is still limited.
Thus, attention mechanisms~\cite{Vaswani2017, Wang2018b, Fu2019a, Li2019a} can use to extract long-range contextual information.
Furthermore, CCNet~\cite{Huang2020} efficiently makes a feature from any position perceive the other features, \ie, it adds contextual information in horizontal and vertical directions with sparse attention.
In contrast, DANet~\cite{Fu2019} proposes two attention modules (dual attentions), focused on spatial features and channels.
Finally, OCNet~\cite{Yuan2018} extracts global and local context by grouping (by permutation), splitting, and operating each sub-region with a self-attention module (\ie, fully matrix).

Unlike previous models, PHGMM extracts local and global context information from two specific latent spaces, which we provide an adequate structure and behavior to obtain this context information precisely. In addition to providing a decoding stage capable of recovering the geometric information of the objects (background and boundary) by combining both context information (local y global) with the decoder information through skip connections, deconvolutions, and upsampling.

\section{Global and Local Feature Modeling}
\label{sec:model}

\makeatletter
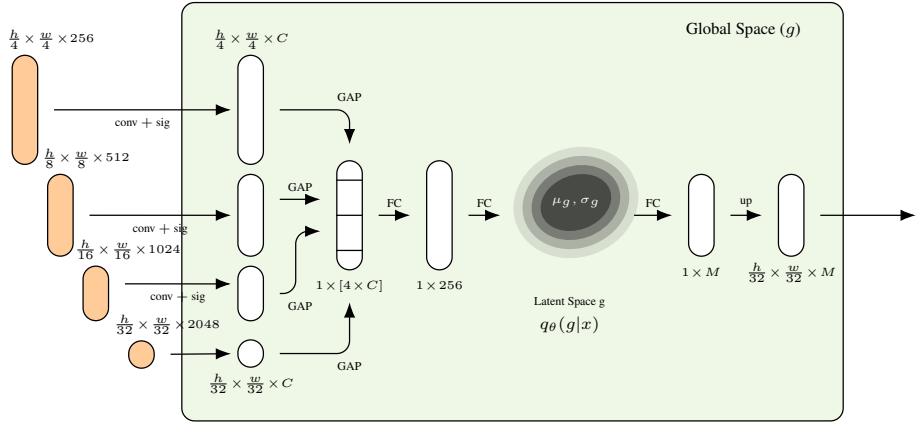
\begin{figure*}[tb]
\centering
\resizebox{.7\linewidth}{!}{%
  \begin{tikzpicture}[
  scale=1,
  every node/.style={scale=0.75},
  node distance=1.25cm,
  pic shift/.store in=\shiftcoord,
  pic shift={(0,0)},
  shift to anchor/.code={
    \tikz@scan@one@point\pgfutil@firstofone(.\tikz@anchor)\relax
    \pgfkeysalso{shift={(-\pgf@x,-\pgf@y)}}
  },
  pics/named scope code/.style 2 args={code={\tikz@fig@mustbenamed%
    \begin{scope}[pic shift, local bounding box/.expanded=\tikz@fig@name, #1]#2\end{scope}%
  }},
  layer/.style={draw, text width=4pt, minimum width=4pt, rounded corners, minimum height=#1},
  pics/encoder2/.style = {named scope code={node distance=4pt}{%
    \node[layer=.8cm, fill=cs-encoder, fill opacity=0.4] (-l2) {};
    \node[layer=1.2cm, above left=of -l2, fill=cs-encoder, fill opacity=0.4] (-l1) {};
    \node[layer=.4cm, below right=0.3cm of -l2.south east, fill=cs-encoder, fill opacity=0.4] (-l3) {};
    \node[layer=1.6cm, above left=of -l1, fill=cs-encoder, fill opacity=0.4] (-l0) {};
  }},
  pics/decoder/.style = {named scope code={node distance=4pt}{%
    \node[layer=.4cm] (-l0) {};
    \node[layer=.8cm, right=of -l0] (-l1) {};
    \node[layer=1.2cm, right=of -l1] (-l2) {};
    \node[layer=1.6cm, right=of -l2] (-l3) {};
  }},
  label/.style={
    node distance=5pt,
    font=\tiny,
  },
  edge/.style={
    >=Latex,
    ->,
    rounded corners,
    shorten >= 5pt,
    shorten <= 5pt,
  },
  linea/.style={
    >=Latex,
    rounded corners,
    shorten >= 5pt,
    shorten <= 5pt,
  },
  linea/.style={
    >=Latex,
    rounded corners,
    shorten >= 5pt,
    shorten <= 5pt,
  },
  ]

  \newcommand{\gaussellipse}[5]{{
    \foreach\i in {0,0.1,...,0.5} {
      \pgfmathsetmacro{\radx}{#3-\i}
      \pgfmathsetmacro{\rady}{#4-\i}
      \fill[#5,opacity=\i,rotate around={#2:#1}] #1 ellipse ({\radx} and {\rady});
    }
  }}

  \definecolor{cs-sky}{RGB}{70,130,180}
  \definecolor{cs-build}{RGB}{70,70,70}
  \definecolor{cs-road}{RGB}{128,64,128}
  \definecolor{cs-sidewalk}{RGB}{244,35,232}
  \definecolor{cs-fence}{RGB}{190,153,153}
  \definecolor{cs-vege}{RGB}{107,142, 35}
  \definecolor{cs-pole}{RGB}{153,153,153}
  \definecolor{cs-car}{RGB}{0,0,142}
  \definecolor{cs-sign}{RGB}{220,220,0}
  \definecolor{cs-person}{RGB}{220,20,60}
  \definecolor{cs-cyclist}{RGB}{119,11,32}
  \definecolor{cs-white}{RGB}{255,255,255}

  \definecolor{cs-encoder}{RGB}{255, 127, 0}
  \definecolor{cs-localz}{RGB}{166,206,227}
  \definecolor{cs-globalg}{RGB}{178,223,138}

  \draw pic[] (enc4) {encoder2};

  \node[fit={($(enc4-l0.north)+(0pt, 0pt)$) ($(enc4-l3.south)+(229pt, -65pt)$)}, right=0.3cm of enc4, inner sep=+2mm, draw, rounded corners, fill=cs-globalg, fill opacity=0.2] (globalspaceg) {};

  \node[layer=1.6cm, right=2.2cm of enc4-l0.east, fill=white] (convsig0) {};
  \node[layer=1.2cm, below=0.1cm of convsig0, fill=white] (convsig1) {};
  \node[layer=0.8cm, below=0.1cm of convsig1, fill=white] (convsig2) {};
  \node[layer=0.4cm, below=0.2cm of convsig2.south, fill=white] (convsig3) {};

  \node[layer=1.6cm, right=0.8cm of convsig1, fill=white] (gap) {};
  \draw[linea] ($(gap.west)+(-5pt,11pt)$) -- ($(gap.east)+(5pt,11pt)$);
  \draw[linea] ($(gap.west)+(-5pt,0)$) -- ($(gap.east)+(5pt,0)$);
  \draw[linea] ($(gap.west)+(-5pt,-11pt)$) -- ($(gap.east)+(5pt,-11pt)$);

  \node[layer=1.6cm, right=0.7cm of gap, fill=white] (lsg) {};

  \gaussellipse{(5.3,1)}{20}{0.8}{0.7}{} (gaussg0)

  \node[layer=1.2cm, right=2.6cm of lsg, fill=white] (procg0) {};

  \node[layer=1.2cm, right=0.7cm of procg0, fill=white] (procg1) {};

  \draw[edge] (enc4-l0) -- node[label,below]{conv $+$ sig} ($(convsig0.west)+(3pt, 0)$);
  \draw[edge] (enc4-l1) -- node[label,below]{conv $+$ sig} ($(convsig1.west)+(3pt, 0)$);
  \draw[edge] ($(enc4-l2.east)+(0pt, 3pt)$) -- node[label,below]{conv $+$ sig} ($(convsig2.west)+(3pt, 3pt)$);
  \draw[edge] (enc4-l3) -- ($(convsig3.west)+(3pt, 0)$);

  \draw[edge] (convsig0) -| node[label,above]{GAP} (gap.north);
  \draw[edge] ($(convsig1.east)+(0, 5pt)$) -- node[label,above]{GAP} ($(gap.west)+(0, 5pt)$);
  \draw[edge] (convsig2.east) -- node[label,below right]{GAP} ++(10pt,0) |- ($(gap.west)+(0, -5pt)$);
  \draw[edge] (convsig3) -| node[label,below]{GAP} ($(gap.south)+(0, -5pt)$); %
  \draw[edge] (gap) -- node[label,above]{FC} (lsg);
  \draw[edge] (lsg) -- node[label,above]{FC} ($(lsg.east)+(20pt, 0pt)$);
  \draw[edge] ($(lsg.east)+(52pt, 0pt)$) -- node[label,above]{FC} (procg0);
  \draw[edge] (procg0) -- node[label,above]{up} (procg1);

  \draw[edge] (procg1) -- ($(procg1.east)+(40pt, 0pt)$);

  \node[label] at ($(procg1.north)+(-15pt, 45pt)$) {\scriptsize Global Space ($g$)};

  \node[label,cs-white] at (5.3,1.0) {$\mu_g, \sigma_g$};

  \node[label] at ($(enc4-l0.north)+(8pt, 5pt)$) {$\frac{h}{4} \times \frac{w}{4} \times 256$};
  \node[label] at ($(enc4-l1.north)+(8pt, 5pt)$) {$\frac{h}{8} \times \frac{w}{8} \times 512$};
  \node[label] at ($(enc4-l2.north)+(10pt, 5pt)$) {$\frac{h}{16} \times \frac{w}{16} \times 1024$};
  \node[label] at ($(enc4-l3.north)+(8pt, 5pt)$) {$\frac{h}{32} \times \frac{w}{32} \times 2048$};

  \node[label] at ($(convsig0.north)+(0pt, 5pt)$) {$\frac{h}{4} \times \frac{w}{4} \times C$};
  \node[label] at ($(convsig3.south)+(0pt, -5pt)$) {$\frac{h}{32} \times \frac{w}{32} \times C$};
  \node[label] at ($(gap.south)+(0pt, -5pt)$) {$1 \times [4 \times C]$};

  \node[label] at ($(lsg.south)+(0pt, -5pt)$) {$1 \times 256$};

  \node[label] at ($(gap.south east)+(65pt, -10pt)$) {Latent Space g};
  \node[label] at ($(gap.south east)+(65pt, -17pt)$) {\scriptsize $q_\theta(g|x)$};

  \node[label] at ($(procg0.south)+(0pt, -5pt)$) {$1 \times M$};

  \node[label] at ($(procg1.south)+(0pt, -5pt)$) {$\frac{h}{32}\times \frac{w}{32} \times M$};

  \end{tikzpicture}
}
\caption{
  Global Space diagram.
  The blocks and arrows represent the feature vectors and operations, respectively.
  First, we extract features at different scales through convolution operations with a sigmoid activation function.
  Next, we use the Global Average Pooling operation (GAP) to summarize the information and gather it into a concatenated feature vector.
  Subsequently, we carry out two Fully Connected layers (FC) for a better features extraction, obtaining our Normal Gaussian.
  Then another FC transforms the mean of the Gaussian into a vector feature that is scaled to $\frac{h}{32} \times \frac{w}{32}$.
}
\label{fig:PHGMM_globalspace}
\end{figure*}
\makeatother

In segmentation, global features can generalize entire objects or a set in a compact representation, while local features, on the other hand, compute representations for particular (more focused) parts of the image.
However, due to the sensitivity to occlusion in global features or noise in local ones, it is not straightforward to combine both features reliably~\cite{Gonzalez2006}.
Consequently, we propose a model capable of merging these local and global features in a suitable manner.
Our contribution comes from combining the features in two stages.
First, we decode the local features representation, and then add the global ones once the image structure has been recovered in the decoder.
We recover structural information, \ie, background and contour of the objects, by merging local and global features (latent spaces) with decoder features through skip connections and upsampling in the decoding stage.
Thus, we obtain diverse types of contextual information, reducing overfitting and addressing the loss of spatial precision problem.

Current semantic segmentation models work by extracting features and pooling them to reduce the representation dimension.
Then, while decoding the encoded features, these models pass part of the encoded information to aid the decoder in producing better features.
In a way, these models use local features and share them across the encoding and decoding tasks to improve the final prediction~\cite{Noh2015}.
Differently, we propose to use local and global features to improve the classes' representation by adding information to the shared knowledge between these two stages.
Our proposal is to create a global model that holds the information of macro objects of the scene, while using the traditional local features, as well as the helpful combination of these and encoder features; through the decoder.
We propose to use Gaussian Mixture Models to represent both spaces (\ie, global and local), since we need to hold multi-modal representations due to the different classes present in a scene (\cf Fig.~\ref{fig:PHGMMPos_train}).

\begin{figure}[tb]
\centering
\resizebox{\linewidth}{!}{%
\begin{tikzpicture}[
scale=1,
every node/.style={scale=0.75},
node distance=1.25cm,
pic shift/.store in=\shiftcoord,
pic shift={(0,0)},
shift to anchor/.code={
  \tikz@scan@one@point\pgfutil@firstofone(.\tikz@anchor)\relax
  \pgfkeysalso{shift={(-\pgf@x,-\pgf@y)}}
},
pics/named scope code/.style 2 args={code={\tikz@fig@mustbenamed%
  \begin{scope}[pic shift, local bounding box/.expanded=\tikz@fig@name, #1]#2\end{scope}%
}},
layer/.style={draw, text width=4pt, minimum width=4pt, rounded corners, minimum height=#1},
pics/encoder2/.style = {named scope code={node distance=10pt}{%
  \node[layer=1.6cm, fill=cs-encoder, fill opacity=0.4] (-l0) {};
  \node[layer=1.2cm, below right=of -l0, fill=cs-encoder, fill opacity=0.4] (-l1) {};
  \node[layer=.8cm, below right=of -l1, fill=cs-encoder, fill opacity=0.4] (-l2) {};
  \node[layer=.4cm, below right=of -l2, fill=cs-encoder, fill opacity=0.4] (-l3) {};
}},
pics/decoder/.style = {named scope code={node distance=10pt}{%
  \node[layer=.4cm, fill=cs-decoder, fill opacity=0.4] (-l0) {};
  \node[layer=.8cm, above right=of -l0, fill=cs-decoder, fill opacity=0.4] (-l1) {};
  \node[layer=1.2cm, above right=of -l1.north, fill=cs-decoder, fill opacity=0.4] (-l2) {};
  \node[layer=1.6cm, above right=of -l2.north, fill=cs-decoder, fill opacity=0.4] (-l3) {};
}},
label/.style={
  node distance=5pt,
  font=\tiny,
},
edge/.style={
  >=Latex,
  ->,
  rounded corners,
  shorten >= 5pt,
  shorten <= 5pt,
},
linea/.style={
  >=Latex,
  rounded corners,
  shorten >= 5pt,
  shorten <= 5pt,
},
nodecircle/.style={
  circle,
  draw=black,
  fill=white,
  opacity=1,
  font=\tiny,
},
]

\newcommand{\gaussellipse}[5]{{
  \foreach\i in {0,0.1,...,0.5} {
    \pgfmathsetmacro{\radx}{#3-\i}
    \pgfmathsetmacro{\rady}{#4-\i}
    \fill[#5,opacity=\i,rotate around={#2:#1}] #1 ellipse ({\radx} and {\rady});
  }
}}

\definecolor{cs-sky}{RGB}{70,130,180}
\definecolor{cs-build}{RGB}{70,70,70}
\definecolor{cs-road}{RGB}{128,64,128}
\definecolor{cs-sidewalk}{RGB}{244,35,232}
\definecolor{cs-fence}{RGB}{190,153,153}
\definecolor{cs-vege}{RGB}{107,142, 35}
\definecolor{cs-pole}{RGB}{153,153,153}
\definecolor{cs-car}{RGB}{0,0,142}
\definecolor{cs-sign}{RGB}{220,220,0}
\definecolor{cs-person}{RGB}{220,20,60}
\definecolor{cs-cyclist}{RGB}{119,11,32}
\definecolor{cs-white}{RGB}{255,255,255}

\definecolor{cs-encoder}{RGB}{255, 127, 0}
\definecolor{cs-decoder}{RGB}{202,178,214}
\definecolor{cs-localz}{RGB}{166,206,227}
\definecolor{cs-globalg}{RGB}{178,223,138}
\definecolor{cs-fenc}{RGB}{129,15,124}
\definecolor{cs-fm}{RGB}{37,52,148}

\draw pic[] (enc3) {encoder2};

\node[layer=1.7cm, rotate=90, below right=0.8cm of enc3-l3, fill=cs-localz, fill opacity=0.2] (lszlocal) {};
\node[layer=1.7cm, rotate=90, below=0.5cm of lszlocal.north, fill=cs-globalg, fill opacity=0.2] (lsgglobal) {};

\pic[right=2.4cm of enc3-l3] (dec) {decoder};

\node[draw, right=15.7pt of dec-l3, minimum width=18pt, rounded corners, minimum height=1.8cm, fill=cs-fm, fill opacity=0.2] (deconcat) {};
\node[draw, right=18.8pt of dec-l3, minimum width=2pt, minimum height=1.8cm] (deconcat1) {};
\node[draw, right=21.2pt of dec-l3, minimum width=2pt, minimum height=1.8cm] (deconcat2) {};

\draw[edge, draw=cs-fenc] (enc3-l3) -- (dec-l0);
\draw[edge, draw=cs-fenc] (enc3-l2) -- (dec-l1);
\draw[edge, draw=cs-fenc] (enc3-l1) -- (dec-l2);
\draw[edge, draw=cs-fenc] (enc3-l0) -- (dec-l3);

\draw[edge, draw=black!50] ($(enc3-l0.south)+(2pt, 4pt)$) |- ($(enc3-l1.north)+(0pt, -4pt)$);
\draw[edge, draw=black!50] ($(enc3-l1.south)+(2pt, 4pt)$) |- ($(enc3-l2.north)+(0pt, -4pt)$);
\draw[edge, draw=black!50] ($(enc3-l2.south)+(2pt, 4pt)$) |- ($(enc3-l3.north)+(0pt, -4pt)$);

\draw[edge, draw=black!50] ($(dec-l2.north)+(0pt, -3pt)$) -| ($(dec-l3.south)+(0pt, 4pt)$);
\draw[edge, draw=black!50] ($(dec-l1.north)+(0pt, -3pt)$) -| ($(dec-l2.south)+(0pt, 4pt)$);
\draw[edge, draw=black!50] ($(dec-l0.north)+(0pt, -2pt)$) -| ($(dec-l1.south)+(0pt, 4pt)$);

\draw[linea] ($(enc3-l0.south)+(0pt, 4pt)$) |- ($(lszlocal.west)+(-25pt, -3pt)$);
\draw[linea] ($(enc3-l1.south)+(0pt, 4pt)$) |- ($(lszlocal.west)+(-25pt, -3pt)$);
\draw[linea] ($(enc3-l2.south)+(0pt, 4pt)$) |- ($(lszlocal.west)+(-25pt, -3pt)$);
\draw[linea] ($(enc3-l3.south)+(0pt, 4pt)$) |- ($(lszlocal.west)+(-25pt, -3pt)$);
\draw[edge] ($(lszlocal.west)+(-30pt, -8pt)$) |- ($(lszlocal)+(-14pt, 0pt)$);
\draw[edge] ($(lszlocal.west)+(-30pt, -8pt)$) |- ($(lsgglobal)+(-14pt, 0pt)$);

\draw[edge] ($(lszlocal.center)+(14pt, 0pt)$) -| ($(dec-l0)+(0pt, 0pt)$);
\draw[edge, draw=cs-fm] ($(dec-l3)+(0pt, 0pt)$) -- ($(deconcat)+(-3pt, 0pt)$);
\draw[edge, draw=cs-fm] ($(dec-l2)+(0pt, 0pt)$) -| ($(deconcat.south)+(-3pt, 5pt)$);
\draw[edge, draw=cs-fm] ($(dec-l1)+(0pt, 0pt)$) -| ($(deconcat.south)+(0pt, 5pt)$);
\draw[edge, draw=cs-fm] ($(dec-l0)+(0pt, 0pt)$) -| ($(deconcat.south)+(3pt, 5pt)$);
\draw[edge, draw=cs-fm] ($(lsgglobal)+(14pt, 0pt)$) -| ($(deconcat.south)+(6pt, 5pt)$);

\node[label] at ($(lszlocal.center)+(0pt, 0pt)$) {\scriptsize $z$};
\node[label] at ($(lsgglobal.center)+(0pt, 0pt)$) {\scriptsize $g$};

\node[label] at ($(enc3-l0.north)+(8pt, 5pt)$) {$\frac{h}{4} \times \frac{w}{4} \times 256$};
\node[label] at ($(enc3-l1.north)+(8pt, 5pt)$) {$\frac{h}{8} \times \frac{w}{8} \times 512$};
\node[label] at ($(enc3-l2.north)+(10pt, 5pt)$) {$\frac{h}{16} \times \frac{w}{16} \times 1024$};
\node[label] at ($(enc3-l3.north)+(8pt, 5pt)$) {$\frac{h}{32} \times \frac{w}{32} \times 2048$};

\node[label] at ($(dec-l0.north)+(-11pt, 5pt)$) {$\frac{h}{32} \times \frac{w}{32} \times [3 \times C]$};
\node[label] at ($(dec-l1.north)+(-16pt, 5pt)$) {$\frac{h}{16} \times \frac{w}{16} \times [3 \times C]$};
\node[label] at ($(dec-l2.north)+(-14pt, 5pt)$) {$\frac{h}{8} \times \frac{w}{8} \times [3 \times C]$};
\node[label] at ($(dec-l3.north)+(-13pt, 4pt)$) {$\frac{h}{4} \times \frac{w}{4} \times [3 \times C]$};

\node[label] at ($(deconcat.north)+(-3pt, 9pt)$) {$\frac{h}{4} \times \frac{w}{4} \times [3 \times C \times M]$};

\node[label] at ($(lszlocal.west)+(-55pt, -8pt)$){\scriptsize conv $+$ sig};
\node[label] at ($(lsgglobal)+(60pt, 5pt)$){\scriptsize conv $+$ up};
\node[label] at ($(dec-l2)+(18pt, 5pt)$){\scriptsize conv $+$ relu};
\node[label] at ($(dec-l2)+(15pt, -5pt)$){\scriptsize $+$ up};
\node[label] at ($(dec-l1)+(26pt, 5pt)$){\scriptsize conv $+$ relu};
\node[label] at ($(dec-l1)+(26pt, -5pt)$){\scriptsize $+$ up};
\node[label] at ($(dec-l0)+(40pt, 5pt)$){\scriptsize conv $+$ relu $+$ up};

\node[label, above=of dec-l3 -| lszlocal] (c3-lbl) {\scriptsize conv};
\node[label, above=of dec-l2 -| lszlocal] (c2-lbl) {\scriptsize conv};
\node[label, above=of dec-l1 -| lszlocal] (c1-lbl) {\scriptsize conv};
\node[label, above=of dec-l0 -| lszlocal] (c0-lbl) {\scriptsize conv};

\node[nodecircle, inner sep=0.5pt] (zop) at ($(lszlocal)+(30pt, 0pt)$) {$||$};
\node[nodecircle, inner sep=0.5pt] (gop) at ($(lsgglobal)+(94pt, 0pt)$) {$||$};
\node[nodecircle, inner sep=0.5pt] (decl0op) at ($(dec-l0)+(56pt, 0pt)$) {$||$};
\node[nodecircle, inner sep=0.5pt] (decl1op) at ($(dec-l1)+(41pt, 0pt)$) {$||$};
\node[nodecircle, inner sep=0.5pt] (decl2op) at ($(dec-l2)+(30pt, 0pt)$) {$||$};

\node[nodecircle, inner sep=0.5pt] (encl0op) at ($(dec-l0)+(-20pt, 0pt)$) {$+$};
\node[nodecircle, inner sep=0.5pt] (encl1op) at ($(dec-l1)+(-20pt, 0pt)$) {$+$};
\node[nodecircle, inner sep=0.5pt] (encl2op) at ($(dec-l2)+(-20pt, 0pt)$) {$+$};
\node[nodecircle, inner sep=0.5pt] (encl3op) at ($(dec-l3)+(-20pt, 0pt)$) {$+$};

\node[layer=0.4cm, rotate=90, above right=1.5cm of enc3-l0.north, fill=cs-encoder, fill opacity=0.4] (legendenc) {};
\node[label] at ($(legendenc.center)+(15pt, 0pt)$){\scriptsize Encoder};
\node[layer=0.4cm, rotate=90, below=0.4cm of legendenc.north, fill=cs-decoder, fill opacity=0.4] (legenddec) {};
\node[label] at ($(legenddec.center)+(19pt, 0pt)$){\scriptsize Decoder ($f$)};
\node[layer=0.4cm, rotate=90, right=1.6cm of legendenc.south west, fill=cs-localz, fill opacity=0.2] (legendz) {};
\node[label] at ($(legendz.center)+(24pt, 0pt)$){\scriptsize Local space $(z)$};
\node[layer=0.4cm, rotate=90, below=0.4cm of legendz.north, fill=cs-globalg, fill opacity=0.2] (legendg) {};
\node[label] at ($(legendg.center)+(25pt, 0pt)$){\scriptsize Global space $(g)$};
\draw[edge, draw=cs-fenc] ($(legendz)+(50pt, 0pt)$) -- ($(legendz)+(70pt, 0pt)$);
\node[label] at ($(legendz)+(79pt, 0pt)$){\scriptsize Skip $(f_{\text{enc}})$};
\node[layer=0.4cm, rotate=90, right=2.3cm of legendg.south west, fill=cs-fm, fill opacity=0.2] (legendfm) {};
\draw[edge, draw=cs-fm] ($(legendg)+(50pt, 0pt)$) -- ($(legendg)+(70pt, 0pt)$);
\node[label] at ($(legendg)+(91pt, 0pt)$){\scriptsize Merger $(f_m)$};

\end{tikzpicture}
}
\caption{
The decoding diagram is fed by the previously processed feature vector  (\ie, local space~$z$).
Next, the decoder takes the skip connections, deconvolution, and upsampling from the encoding.
Finally, in its last layer, it concatenates the information of the different levels (\ie, multi-scale information) with the global information (\ie, global space~$g$).
Note, we call the latent space $g$ a global representation because this space summarizes features of objects in a holistic way (\ie, a global view of the object).
Also, our decoding stage is in charge of merging the different context information extracted at the different levels of our network.
This smart fusion recovers the geometric information of the segmented objects.
}
\label{fig:PHGMM_f_merge}
\end{figure}
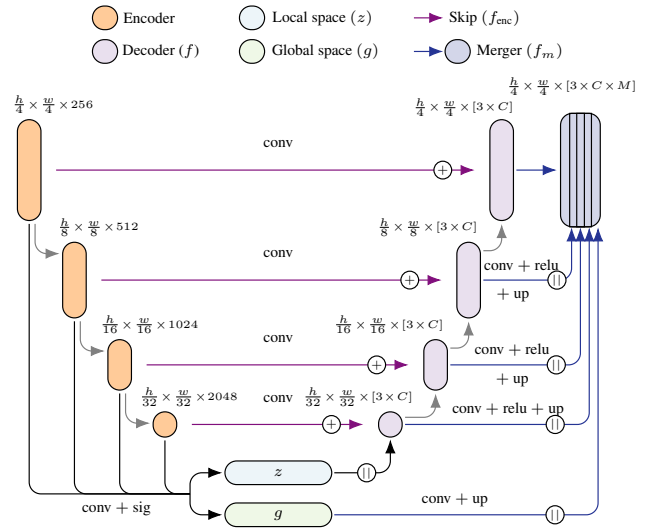

Our objective is to model two latent spaces to perform a better and more robust extraction of local (Fig.~\ref{fig:PHGMM_localspace}) and global (Fig.~\ref{fig:PHGMM_globalspace}) features and an effective way to combine them through the decoder (Fig.~\ref{fig:PHGMM_f_merge}) to recover the geometric information of the objects.
Thus, we propose using an end-to-end model that fits and employs two latent spaces suitable for the segmentation task.
In order to provide an internal structure to the learned representation (\ie, latent space), we merge VAE with ResNet-101~\cite{He2016} (for an encoder-decoder stage for SS).
A VAE allows us to represent the latent space through a structure, \ie, a predefined distribution.
Furthermore, the ResNet-101 backbone provides robust processing (\eg, residual block) to extract the basic features that will turn into the distributions.
As a result, our model obtains a local context for the discrimination of segmented objects (\ie, class-level information) as a mixture of Gaussians.
The model mixes the structure of the local-global features by employing concatenation operation and upsampling throughout the deconvolution stage.
In addition, the decoder stage is responsible for recovering the geometric information of the objects.
Combining the context information (local-global structure features) previously extracted with the features coming from the encoder through skip connections in the different scales.
Integrating these two latent spaces provides the neural network with a powerful representation (\cf Fig.~\ref{fig:ablation_study_z_t}).
The latent spaces provide complementary information that improves the semantic segmentation (\cf Tables~\ref{tab:result_cityscape} and~\ref{tab:result_synthia}).

We present an overview of our PHGMM model in Fig.~\ref{fig:PHGMMPos_train}, and detail the inference and generation processes in Sections~\ref{sec:inference} and~\ref{sec:generative}, respectively.
Also, in Section~\ref{sec:architecture}, we split the description of the operations used for local and global context extraction.
The local and global extraction processes are shown in Figs.~\ref{fig:PHGMM_localspace} and~\ref{fig:PHGMM_globalspace}, respectively.
Fig.~\ref{fig:PHGMM_f_merge} shows that our decoder is in charge of intelligently merging the previously extracted features so that PHGMM recovers the geometric information of the segmented objects addressing the problem of loss of spatial precision (see Table.~\ref{fig:cityscape_trimap}).
Finally, in Section~\ref{sec:training}, we describe how our model is trained.

\makeatletter
\makeatother

\subsection{Inference Process}
\label{sec:inference}

We jointly model the global information of the scene~$g$ and the local features~$z$ of the image~$x$ and assume that they are independent.
We set the distribution that models this relation as $q_\theta(z, g \given x) = q_\theta(z \given x) q_\theta(g \given x)$.
The local distribution
\begin{equation}
\label{eq:latent_space_z}
q_\theta(z \given x) = \sum_{k=1}^K  \pi^k(x) \mathcal{N} \left( z; \mu_z^k(x), \sigma_z^k(x)^2 \mathbf{I} \right)
\end{equation}
is a mixture of Gaussians that models $K$~clusters of related features that are helpful in obtaining different classes for the segmentation.
The functions $\mu_z^k(\cdot)$, $\sigma_z^k(\cdot)$, and $\pi^k(\cdot)$ are the parameters of the $k$-th component of the mixture and depend on the given image~$x$ (see Fig.~\ref{fig:PHGMMPos_train}).
All are approximated through a ResNet-101 backbone with separate projection heads that output the corresponding parameter.
On the other hand, the global distribution
\begin{equation}
\label{eq:latent_space_g}
q_\theta(g \given x) = \mathcal{N} \left( g; \mu_g(x), \sigma_g(x)^ 2 \mathbf{I} \right)
\end{equation}
is a single Gaussian that models a global context that holds all the image~$x$ information.
The parameter functions $\mu_g(\cdot)$ and $\sigma_g(\cdot)$ are computed in a similar way as the previous parameters.

The local features~$z$ hold information about the features, while the global ones~$g$ serve as a latent variable to model holistic information from the whole scene.
Additionally, due to the way we engineer the decoding stage (see Fig.~\ref{fig:PHGMMPos_train}), the global features also serve to spread the gradient back to the encoder.

\subsection{Generative Process}
\label{sec:generative}
To obtain the classes~$y$ for the segmentation, we model the joint probability $p(x, y, z, g)$ of the data, the images~$x$ and the labels~$y$, and the latent variables, the local~$z$ and global~$g$.
We factorize the joint as $p_{\theta,\phi}(x, y, z, g) = p_{\phi}(y \given z, g) p_\theta(z \given x) p_\theta(g \given x) p(x)$.

To implement the generative process from $p(y \given z, g)$, we use the decoder $f(\bar{\mu}_z)$, where $\bar{\mu}_z = \Cat_{k=1}^K \mu_z^k$ is the concatenation of all the means from the mixture.
Later, a merge function $f_m(f(\bar{\mu}_z), \mu_g, f_{\text{enc}})$ combines the partially decoded segmentations with the parameters from the global context and the encoder features, $f_{\text{enc}}$, from the skip connections.
Hence, the predicted labels are $y' = f_m(f(\bar{\mu}_z), \mu_g, f_{\text{enc}})$.
For a better interpretation, we present the associated diagram in Fig.~\ref{fig:PHGMM_f_merge}, in later sections.

\section{Implementation}
\label{sec:implement}
In this section, we explain in detail the PHGMM architecture, \ie, local space $z$ and global space $g$, as well as the operations involved in modeling latent spaces.
In addition, we explain how to merge the different local-global context information and the encoder features to recover the details of the segmented objects (\ie, the geometric information).
Finally, we perform the loss functions used to adjust the PHGMM parameters and show the hyperparameters used for our model.

\subsection{Architecture}
\label{sec:architecture}
We use a ResNet-101 as a backbone for the PHGMM model due to its good behavior, \ie, the trade-off between good feature extraction and the number of parameters used, that is, most of the residual blocks and depths of the layers.
The samples that PHGMM receives as input are passed through convolution with kernel $7 \times 7$ and depth $64$ to carry out the information compression, thus converting the samples in size $\frac{h}{4} \times \frac{w}{4} \times 64$.
Next, we use a defined decoder in four groups of blocks of depth $256$, $512$, $1024$, and $2048$ respectively.
Each group presents a set of $3$, $4$, $23$, and $3$ internal residual blocks.
Notice that our feature vector shows a reduction in the size of $(\frac{h}{4} \times \frac{w}{4})$, $(\frac{h}{8} \times \frac{w}{8})$, $(\frac{h}{16} \times \frac{w}{16})$, and $(\frac{h}{32} \times \frac{w}{32})$ as it passes through the decoder.

One of the issues we address in this paper is how to take advantage of this multi-scale information to improve the edges of the segmented objects and thus address the problem of the loss of spatial precision.
To address this, we create local space $z$ (Fig.~\ref{fig:PHGMM_localspace}) to model the latent space through a mixture of Gaussians, focusing on local features and global space $g$ (Fig.~\ref{fig:PHGMM_globalspace}) to extract global information through Gaussian modeling of a second latent space.
Additionally, we use skip connections to pass complemental multi-scale information from the encoder to the decoder.

Thus, in local space $z$ (see Fig.~\ref{fig:PHGMM_localspace}), we use $3 \times 3$ kernel convolutions together with a non-linear sigmoid function to compress the information of the different levels, reducing the depth from $256$, $512$, $1024$, and $2048$ for the number of classes ($C$).
We concatenate the feature vectors to merge the multi-scale information through Global Average Pooling (GAP).
To model the latent space $z$, we use a Fully Connected layer (FC) and produce $K$ mixtures of Gaussians (\ie, $\mu_z^k$, $\sigma_z^k$, $\pi_z^k$).
Note that the concatenation vector and the Gaussian parameters (\ie, $\mu$, $\sigma$) have a size of $1 \times 256$ (chosen empirically).
We combine the local information from the GMMs by stacking the $K$ vectors (\ie, $\mu_z^k \times \pi_z^k$).
For each item in the stack, we perform an FC with a non-linear function RELu, transforming the features of size $1 \times 256$ to $1 \times 1 \times D$.
Then we scale the vectors achieving $\frac{h}{32} \times \frac{w}{32} \times D$.
Finally, we perform a non-linear combination on the concatenation vector of size $\frac{h}{32} \times \frac{w}{32} \times [K \times D]$, reaching a depth of $512$.
These feature vector locales will be used as input for our encoder.
Note that even though we use GAPs in the process of converting the samples to Gaussian distributions, we call local latent space to the local space $z$ because, from this feature vector, the geometric reconstruction is performed to recover and refine the details of the edges of segmented objects.

On the other hand, the global space $g$ (see Fig.~\ref{fig:PHGMM_globalspace}) performs the treatment of multi-scale information to a certain extent similar to the local space $z$, through the concatenation (of size $1 \times [4 \times C]$) of feature vectors summarized by GAP operations.
Next, we use two FC to model the latent space $g$ and give it the structure of a Normal Gaussian with parameters $\mu_g$ and $\sigma_g$, both of size $1 \times 256$ (chosen empirically).
Finally, the information is combined by taking $\mu_g$ and passing it through an FC, obtaining a vector of global features of size $1 \times M$, followed by an upscale operation achieving a vector size of $\frac{h}{32} \times \frac{w}{32} \times M$.
Note that we use the latent space $g$ as a global representation because $g$ is in charge of extracting the holistic information of the objects and then concatenates it in the last reconstruction layer of the decoder.

We merge both latent spaces ($z$ and $g$), and decoder features through the decoding stage.
The decoder (see Fig.~\ref{fig:PHGMM_f_merge}) obtains as input the latent space $z$ and has four groups of residual units, where each group has $3$, $23$, $4$, and $3$ residual units, all with $3 \times C$ depth and a feature vector of size $\frac{h}{32} \times \frac{w}{32}$, $\frac{h}{16} \times \frac{w}{16}$, $\frac{h}{8} \times \frac{w}{8}$ and $\frac{h}{4} \times \frac{w}{4}$ respectively.
In each residual unit group of the decoder, our PHGMM model performs a geometric information retrieval by combining the local features coming from the GMM (latent space $z$) with the features extracted in the encoding through an addition operation.
We transform the encoder features through nonlinear combination, \ie, convolution operations and non-linearity functions, using a $3 \times 3$ kernel with depth $C$.
These features are passed from the different levels of the encoder to the decoder (\ie, multi-level information) with the use of skip connections.
In the last layer, we concatenate the decoder's multi-scale reconstruction information through convolution operations with kernel $3 \times 3$ and a non-linearity function RELu and upscale applied to each level (\ie, group) of the decoder.
Besides, a convolution operation of $3 \times 3$ with upscale to latent space $g$.
The concatenation of all this information gives a resulting feature vector of size $\frac{h}{4} \times \frac{w}{4} \times [3 \times C \times M]$.
Finally, for the output of PHGMM, a convolution operation with kernel $8 \times 8$ and a Softmax are applied to combine all the information and obtain our final heat map.

\begin{table*}[tb]
\centering
\sisetup{
  table-format = 2.2,
}
\newlength{\colsep}
\definecolor{cs-sky}{RGB}{70, 130, 180}
\definecolor{cs-build}{RGB}{70, 70, 70}
\definecolor{cs-road}{RGB}{128, 64, 128}
\definecolor{cs-sidewalk}{RGB}{244, 35, 232}
\definecolor{cs-fence}{RGB}{190, 153, 153}
\definecolor{cs-vege}{RGB}{107, 142, 35}
\definecolor{cs-pole}{RGB}{153, 153, 153}
\definecolor{cs-car}{RGB}{0, 0, 142}
\definecolor{cs-sign}{RGB}{220, 220, 0}
\definecolor{cs-person}{RGB}{220, 20, 60}
\definecolor{cs-cyclist}{RGB}{119, 11, 32}

\caption[Results on Cityscapes validation]{IoU results on Cityscapes validation set for semantic segmentation, using $11$ classes and with crop size of $384 \times 768$. The classes are: \class{cs-sky}{1} Sky, \class{cs-build}{1} Building, \class{cs-road}{1} Road, \class{cs-sidewalk}{1} Sidewalk, \class{cs-fence}{1} Fence, \class{cs-vege}{1} Vegetation, \class{cs-pole}{1} Pole, \class{cs-car}{1} Car, \class{cs-sign}{1} Sign, \class{cs-person}{1} Person, \class{cs-cyclist}{1} Cyclist.}
\label{tab:result_cityscape}

\newrobustcmd{\B}{\bfseries}
\setlength{\colsep}{10pt}
\scriptsize
\renewcommand{\arraystretch}{0.8}
\resizebox{0.85\linewidth}{!}{%
\begin{tabular}{@{\hspace{5pt}}
    l@{\hspace{\colsep}} %
    S@{\hspace{\colsep}} %
    S@{\hspace{\colsep}} %
    S@{\hspace{\colsep}} %
    S@{\hspace{\colsep}} %
    S@{\hspace{\colsep}} %
    S@{\hspace{\colsep}} %
    S@{\hspace{\colsep}} %
    S@{\hspace{\colsep}} %
    S@{\hspace{\colsep}} %
    S@{\hspace{\colsep}} %
    S@{\hspace{5pt}} %
    @{\hspace{1pt}}%
    S@{\hspace{\colsep}} %
  }
  \toprule
  \B{Model} & \class{cs-sky}{1.1} & \class{cs-build}{1.1} & \class{cs-road}{1.1} & \class{cs-sidewalk}{1.1} & \class{cs-fence}{1.1} & \class{cs-vege}{1.1} & \class{cs-pole}{1.1} & \class{cs-car}{1.1} & \class{cs-sign}{1.1} & \class{cs-person}{1.1} & \class{cs-cyclist}{1.1} & \B{mIoU (\%)}\\
  \midrule
  \text{SegNet}~\cite{Badrinarayanan2017} & 73.74 & 79.29 & 92.70 & 59.88 & 13.63 & 81.89 & 26.18 & 78.83 & 31.44 & 45.03 & 43.46 & 56.92 \\
  \text{DeconvNet}~\cite{Noh2015} & 89.38 &  83.08 &  95.26 &  68.07 &  27.58 & 85.80 &  34.20 & 85.01 &  27.62 & 45.11 &  41.11 &  62.02  \\
  \text{DeepLab v2}~\cite{Chen2017a} & 74.28 & 81.66 & 90.86 & 63.3 & 26.29 & 84.33 & 27.96 & 86.24 & 44.79 & 58.89 & 60.92 & 63.59 \\
  \text{FCN8~\cite{Long2016}} & 76.51 &  83.97 &  93.82 &  67.67 &  24.91 &  86.38 &  31.71 &  84.80 &  50.92 &  59.89 & 59.11 &  65.43  \\
  \text{FastNet~\cite{Oliveira2016}} & 77.69 & 86.25 & 94.97 & 72.99 & 31.02 & 88.06 & 38.34 & 88.42 & 52.34 & 61.76 & 61.83 & 68.52 \\
  \text{ParseNet~\cite{Liu2015}} & 90.76340732 & 85.20362826 & 92.00606416 & 63.49442359 & 38.99991853 & 88.20904710 & 46.81669972 & 89.19935668 & 61.90421068 & 63.91352437 & 62.72517683 & 71.2032233889497 \\
  \text{ESPNet}~~\cite{Mehta2019} & 91.78638812 & 86.35574516 & 95.73185838 & 71.84438459 & 48.52102068 & 88.4447178 & 49.06453645 & 87.29345279 & 54.60143022 & 61.82588989 & 57.51110799 & 72.08913927909090 \\
  \text{FC-DenseNet67~\cite{Jegou2017}} & 92.19346337765376 & 86.76988562706273 & 96.59884889225275 & 75.39657905029527 & 41.551418122907954 & 88.07405997289057 & 52.923282245565694 & 87.09217726828671 & 63.88987642687341 & 60.48381609817361 & 52.92274378685763 & 72.5360137153472 \\
  \text{BiSeNet}~\cite{Yu2018c} & 91.63755455 & 87.42330637 & 96.48050315 & 75.41466065 & 44.04604326 & 89.07274113 & 39.50127053 & 89.33916815 & 58.62617238 & 66.80642806 & 63.07673701 & 72.85678047715217 \\
  \text{ENet}~\cite{Paszke2016} & 91.62527061 & 87.48086007 & 96.43995471 & 75.34161092 & 48.44360288 & 89.23168657 & 43.13950425 & 89.24173128 & 56.03890594 & 65.12868147 & 63.7010055 & 73.25571038043044 \\
  \text{DeepLab v3}~\cite{Chen2017} & 92.82 & 89.02 & 96.74 & 78.13 & 41.00 & 90.81 & 49.74 & 91.02 & 64.48 & 66.52 & 66.98 & 75.21 \\
  \text{PSPNet}~\cite{Zhao2017} & 91.94289991 & 89.93398895 & 96.93718646 & 78.37073789 & 53.6440468 & 90.18757238 & 43.47155108 & 92.11920061 & 64.40262511 & 70.70955096 & 70.94303548 & 76.60567233102417 \\
  \text{DANet}~\cite{Fu2019} & 92.24915714 & 90.26078544 & 97.24916248 & 79.95381396 & 51.33334745 & 90.59671054 & 45.20256646 & 92.50120637 & 66.37661578 & 71.4706024 & 71.24568649 & 77.1308776837044 \\
  \text{AdapNet++~\cite{Valada2019}} & 93.06606033717834 & 89.46445452619719 & 97.05622901868213 & 80.02680655465022 & 49.46480016553393 & 90.57898007902187 & 52.099869078274374 & 92.22088910324935 & 66.2616099406663 & 72.88264400519273 & 70.6234837933447 & 77.61325696381738 \\
  \text{CCNet}~\cite{Huang2020} & 90.96516 & 89.00806 & 96.58898 & 77.36207 & 42.48835 & 91.36251 & 58.24336 & 91.21552 & 71.17928 & 74.80497 & 70.57466 & 77.61753851250874 \\
  \text{OCNet}~\cite{Yuan2018} & 92.71713413 & 90.73445133 & 97.39139266 & 80.80080867 & 54.58097682 & 90.85808692 & 45.59792652 & 92.6447007 & 67.35220639 & 71.80571641 & 71.97828486 & 77.86015321793398 \\
  \text{CGBNet~\cite{Ding2020}} & 92.97664103982117 & 89.39958276438676 & 96.66423105831508 & 77.59557837050379 & 42.80119490412156 & 91.88485470080798 & 57.52385700042729 & 91.13894547133451 & 73.28850836068722 & 75.2576623028372 & 71.1815062265886 & 78.15568747271192 \\
  \text{DUNet}~\cite{Jin2019} & 93.3292409 & 91.04735692 & 97.27624635 & 80.17870018 & 55.14552055 & 91.3528578 &  53.6978367 & 92.88570205 & 68.3278322 & 73.33193179 & 72.29190738 & 78.9877393463894 \\
  \text{HRNet}~\cite{Wang2020} & 94.56263803392699 & 90.9782601331852 & 97.4834383731036 & 82.46280689648792 & 50.269730202459684 & 92.35495660868773 & 61.571180462866714 & 93.96316213743727 & 73.14349328079304 & 78.55382561397901 & 75.44369460350559 & 80.98065330422115 \\
  \text{HRNet + OCR}~\cite{Yuan2021} & 94.4358545529273 & 91.56636315886351 & 97.5902237064308 & 82.99885185368369 & \B 55.464824930830105 & 92.49025959872975 & 62.67544657375434 & 93.97781763087684 & \B 75.66912451675692 & 78.69563372557992 & 75.34932305850215 & 81.90124757335776 \\
  \text{PHGMM} & \B 94.63233601001029 &\B 91.60106905124312 & \B 97.69572777111502 & \B 83.64714307025132 & 55.26833420934679 & \B 92.51727151053036 & \B 62.71005972946803 & \B 94.2014523189135 & 75.34570228830016 & \B 79.21058371057573 & \B 76.45046505096894 & \B 82.11637679279302 \\
  \bottomrule
\end{tabular}}
\end{table*}

\begin{table*}[tb]
\centering
\sisetup{
  table-format = 2.2,
}
\definecolor{st-sky}{RGB}{128, 128, 128}
\definecolor{st-build}{RGB}{128, 0, 0}
\definecolor{st-road}{RGB}{128, 64, 128}
\definecolor{st-sidewalk}{RGB}{0, 0, 192}
\definecolor{st-fence}{RGB}{64, 64, 128}
\definecolor{st-vege}{RGB}{128, 128, 0}
\definecolor{st-pole}{RGB}{192, 192, 128}
\definecolor{st-car}{RGB}{64, 0, 128}
\definecolor{st-sign}{RGB}{192, 128, 128}
\definecolor{st-person}{RGB}{64, 64, 0}
\definecolor{st-cyclist}{RGB}{0, 128, 192}

\caption[Results on Synthia validation]{IoU results on Synthia validation set for semantic segmentation, using $11$ classes and with crop size of $384 \times 768$. The classes are: \class{st-sky}{1} Sky, \class{st-build}{1} Building, \class{st-road}{1} Road, \class{st-sidewalk}{1} Sidewalk, \class{st-fence}{1} Fence, \class{st-vege}{1} Vegetation, \class{st-pole}{1} Pole, \class{st-car}{1} Car, \class{st-sign}{1} Sign, \class{st-person}{1} Person, \class{st-cyclist}{1} Cyclist.}
\label{tab:result_synthia}

\newrobustcmd{\B}{\bfseries}
\setlength{\colsep}{10pt}
\scriptsize
\renewcommand{\arraystretch}{0.8}
\resizebox{0.85\linewidth}{!}{%
\begin{tabular}{@{\hspace{5pt}}
    l@{\hspace{\colsep}} %
    S@{\hspace{\colsep}} %
    S@{\hspace{\colsep}} %
    S@{\hspace{\colsep}} %
    S@{\hspace{\colsep}} %
    S@{\hspace{\colsep}} %
    S@{\hspace{\colsep}} %
    S@{\hspace{\colsep}} %
    S@{\hspace{\colsep}} %
    S@{\hspace{\colsep}} %
    S@{\hspace{\colsep}} %
    S@{\hspace{5pt}} %
    @{\hspace{1pt}}%
    S@{\hspace{\colsep}} %
  }
  \toprule
  \B{Model} & \class{st-sky}{1.1} & \class{st-build}{1.1} & \class{st-road}{1.1} & \class{st-sidewalk}{1.1} & \class{st-fence}{1.1} & \class{st-vege}{1.1} & \class{st-pole}{1.1} & \class{st-car}{1.1} & \class{st-sign}{1.1} & \class{st-person}{1.1} & \class{st-cyclist}{1.1} & \B{mIoU (\%)}\\
  \midrule
  \text{SegNet}~\cite{Badrinarayanan2017} & 91.90 & 87.19 & 83.72 & 80.94 & 50.02 & 71.63 & 26.12 & 71.31 & 1.01 & 52.34 & 32.64 & 58.98 \\
  \text{FCN8~\cite{Long2016}} & 92.36 &  91.92 & 88.94 & 86.46 & 48.22 & 77.41 & 36.02 & 82.63 & 30.37 & 57.10 & 46.84 & 67.11 \\
  \text{DeconvNet}~\cite{Noh2015} & 95.88 & 93.83 & 92.85 & 90.79 & 66.40 & 81.04 & 48.23 & 84.65 & 0.00 & 69.46 & 52.79 & 70.54 \\
  \text{ParseNet~\cite{Liu2015}} & 93.80 & 93.09 & 91.05 & 88.98 & 53.22 & 79.48 & 46.15 & 85.37 & 36.00 & 63.30 & 50.82 & 71.02 \\
  \text{DeepLab v2}~\cite{Chen2017a} & 94.07 & 93.34 & 88.07 & 88.93 & 55.57 & 80.22 & 45.97 & 85.87 & 38.73 & 64.40 & 52.54 & 71.61 \\
  \text{FC-DenseNet67~\cite{Jegou2017}} & 92.73814556976933 & 89.9395336203245 & 83.35090004988923 & 85.72130330179895 & 74.47123045060108 & 70.46750674345112 & 51.13256847382862 & 84.20938649729955 & 35.10596107373156 & 68.7789144013552 & 54.543095772903094 &  71.85986781408657 \\
  \text{FastNet~\cite{Oliveira2016}} & 92.21 & 92.41 & 91.85 & 89.89 & 56.64 & 78.59 & 51.17 & 84.75 & 32.03 & 69.87 & 55.65 & 72.28 \\
  \text{ESPNet}~\cite{Mehta2019} & 95.43823286 & 93.15770559 & 91.5216543 & 89.62311607 & 65.97587661 & 80.11211991 & 48.08913894 & 84.9277357 & 39.62219107 & 66.80256737 & 53.91453925 & 73.56226160672742 \\
  \text{BiSeNet}~\cite{Yu2018c} & 95.03744954 & 92.97008983 & 91.24806482 & 89.07589772 & 63.72464811 & 79.66567326 & 48.43960044 & 84.76681067 & 45.67717217 & 65.88504126 & 54.01226264 & 73.68206458813599 \\
  \text{ENet}~\cite{Paszke2016} & 94.80623702 & 93.01018077 & 91.51169991 & 89.67088157 & 65.56147227 & 79.03701313 & 50.78473701 & 85.2168258 & 42.08862661 & 67.67363271 & 54.13601256 & 73.9543017600822 \\
  \text{ICNet}~\cite{Zhao2018} & 95.7844326 & 94.31929631 & 92.61893427 & 90.76403926 & 64.91143153 & 83.17492561 & 54.37105364 & 87.52909058 & 51.15457222 & 68.02915624 & 56.4693035 & 76.28420325029799 \\
  \text{DeepLab v3}~\cite{Chen2017} & 95.30 & 92.75 & 93.58 & 91.56 & 73.37 & 80.71 & 55.83 & 88.09 & 44.17 & 75.65 & 60.15 & 77.38 \\
  \text{DANet}~\cite{Fu2019} & 96.93447143 & 96.00161362 & 95.21915878 & 93.82209322 & 70.73170426 & 87.36041471 & 64.00945124 & 92.6669109 & 61.39760396 & 76.18953394 & 65.70174948 & 81.82133686747711 \\
  \text{PSPNet}~\cite{Zhao2017} & 96.92212494 & 96.00194732 & 95.26771804 & 93.84681381 & 71.17356701 & 87.29600412 & 65.19687298 & 92.97791942 & 62.7326132 & 76.62439962 & 66.24129414 & 82.20738859888307 \\
  \text{CGBNet~\cite{Ding2020}} & 97.06787 & 96.20894 & 95.4536 & 94.06276 & 71.59252 & 87.90565 & 66.27628 &
  93.27113 & 64.45714 & 77.08918 & 66.96545 & 82.7591380704133 \\
  \text{AdapNet++~\cite{Valada2019}} & 97.09202 & 96.23805 & 95.47351 & 94.09333 & 71.78852 & 87.97434 & 66.23204 &
  93.32698 & 64.92306 & 77.15782 & 67.20667 & 82.86421346545318 \\
  \text{CCNet}~\cite{Huang2020} & 97.22151909 & 96.17349084 & 95.44657886 & 93.88143657 & 72.13568526 & 88.36453937 & 66.73581338 & 93.19174904 & 67.74466553 & 76.88199048 & 67.69651846 & 83.22490789873517 \\
  \text{OCNet}~\cite{Yuan2018} & 97.39121591 & \B 96.49751938 & 95.68929423 & 94.37198817 & 72.49477248 & \B 88.83951937 & 67.49981915 & 93.63770127 & 67.92060139 & 77.5858862  & 67.68750816 & 83.60143870072893 \\
  \text{HRNet}~\cite{Wang2020} & 96.72645 & 94.75738 & 94.828 & 93.94678 & 86.99789 & 83.62414 & 72.81181 & 91.90143 & 61.74201 & 85.30969 & 71.08358 & 84.8844687057615 \\
  \text{HRNet + OCR}~\cite{Yuan2021} & 96.7443 & 94.82684 & 94.29712 & 93.46798 & \B 87.79663 & 84.33265 & \B 75.36277 &
  92.54169 & 62.77966 & \B 87.13963 & \B 72.88337 &  85.65205732467531 \\
  \text{DUNet}~\cite{Jin2019} & \B 97.57455889 & 96.47748927 & 96.11347477 & 95.04700978 & 80.56647131 & 88.08831442 & 73.95292407 & 93.70595222 & 67.47618125 & 82.3821732 & 71.0970727 & 85.68014744482672 \\
  \text{PHGMM} & 97.00813502905118 & 95.89240129113708 & \B 96.2347068203646 & \B 95.17129574558614 & 85.0293086295568 & 84.9081549899404 & 71.77954488726537 & \B 94.2647961456622 & \B 68.26379118088354 & 85.49675649722813 & 72.41706457197729 & \B 86.04235961715027 \\
  \bottomrule
\end{tabular}}
\end{table*}

\subsection{Training}
\label{sec:training}
The target function used to train our PHGMM model $\mathcal{L}$, per data sample, is a linear combination of the specific loss functions $\ell_g$, $\ell_z$, and $\ell_s$ functions, defined by
\begin{equation}
\label{eq:loss_general}
\mathcal{L} = \lambda_g \ell_g + \lambda_z \ell_z + \lambda_s \ell_s,
\end{equation}
where $\ell_g$ and $\ell_z$ are bound to provide a useful structure (\ie, helpful embedding) for the latent spaces,
and $\ell_s$ on the loss function $\mathcal{L}$ focuses on dense pixel-level classification.
Also, $\lambda_{g, z, s}$ are hyperparameters of each term, respectively.
Note, to train with a batch of data, we aggregate the losses over it.

For learning the global representation $g$, we need to find a relevant embedding for the segmentation task (\ie, find useful low-level features).
Consequently, we use Kullback-Leibler divergence (KL)~\cite{Sherman1960} to provide a structure to~$g$ \wrt its prior, \ie,
\begin{equation}
\label{eq:KL_g}
\ell_g = \kl[\big]{q_\theta(g \given x)}{p(g)}.
\end{equation}
The KL works as a regularizer over $g$ and measures the divergence between the encoder distribution $q(g \given x)$ and the prior $p(g)$.
We specify $p(g)$ as a standard Normal distribution $p(g) = \mathcal{N}(g; 0, \text{I})$, and define $q(g \given x)$ as before~\eqref{eq:latent_space_g}.

Our second term provides the latent space $z$ with a clustering behavior. %
Instead of using a default prior, we infer a conditional $p(z \given x, y)$ from data with another net, called PostNet (see Fig.~\ref{fig:PHGMMPos_train}).
This PostNet uses the ground truth labels and image data to produce the posterior distribution while training to serve as regularizer.
The posterior is also a mixture of Gaussians defined by
\begin{equation}
p_\gamma(z \given x, y) = \sum_{k=1}^K  \pi^k_{z^*}(x, y) \mathcal{N} \left( z; \mu_{z^*}^k(x, y), \sigma_{z^*}^k(x, y)^2 \mathbf{I} \right).
\end{equation}
We fit the predicted distribution with the infer one through
\begin{equation}
\label{eq:KL_z}
\ell_z = \kl[\big]{q_\theta(z \given x)}{p_\gamma(z \given x, y)}.
\end{equation}

Recall that our PHGMM model learns to densely predict the semantic classes given an input image~$x$.
Thus, the reconstruction loss function is expressed as a cross-entropy loss between the prediction~$y'$ and its ground-truth~$y$.
Penalizing the pixel-wise prediction for each class is not enough.
Thus, we penalize the segmented objects' contours as well.
We employ the loss function soft intersection-over-union~\cite{Csurka2013}.
So our third term, $\ell_s$, is
\begin{equation}
\label{eq:loss_recons}
\ell_s  = -\frac{1}{C}\sum_{c=1}^C y_c \log(y'_c) + 1 - \frac{1}{C}\sum_{c=1}^C \left(\frac{y'_c \cap y_c }{y'_c \cup y_c}\right),
\end{equation}
where $y'_c$ is our prediction of the ground truth~$y_c$ for the class~$c$.

Finally, We provide an algorithm in Alg.~\ref{alg:training_PHGMM} to better understand the training framework, which summarizes the PHGMM steps process.

\begin{algorithm}[tb]
\small
\caption{Training the PHGMM model}
\label{alg:training_PHGMM}
\begin{algorithmic}
  \State $\theta_{\text{Enc}}, \phi_{\text{Dec}}, \gamma_{\text{PosNet}} \gets $ Initialize networks parameters
  \Repeat
  \State $X \gets$ mini-batch image from dataset
  \State $Y \gets$ mini-batch label from dataset
  \State $X'_1, X'_2, X'_3, X'_4 \gets$ Encoder$(X)$ \Comment{Multi-level features}
  \State $Z, q_\theta(z|x) \gets $ LocalSpace$(X'_1, X'_2, X'_3, X'_4)$
  \State $p_\gamma(z|x,y) \gets$ PostNet$(X,Y)$
  \State $G, q_\theta(g|x) \gets $ GlobalSpace$(X'_1, X'_2, X'_3, X'_4)$
  \State $\ell_z \gets \kl[\big]{q_\theta(z \given x)}{p_\gamma(z \given x, y)}$ \Comment{Eq.~\ref{eq:KL_z}}
  \State $\ell_g \gets \kl[\big]{q_\theta(g \given x)}{p(g)}$ \Comment{Eq.~\ref{eq:KL_g}}
  \State $X''_1, X''_2, X''_3, X''_4 \gets $ Decoder$(Z)$ \Comment{Multi-level features}
  \State $Y' \gets$ convolution of $[X''_1, X''_2, X''_3, X''_4, G]$
  \State $\ell_s \gets \text{CrossEntropy}(Y', Y) + \text{IoU}_{\text{soft}}(Y', Y)$ \Comment{Eq.~\ref{eq:loss_recons}}
  \State $\mathcal{L} = \lambda_g \ell_g + \lambda_z \ell_z + \lambda_s \ell_s$ \Comment{Eq.~\ref{eq:loss_general}}

  \Comment{Update parameters according to gradients}
  \State $\theta_{\text{Enc}} \stackrel{+}\leftarrow - \nabla_{\theta_{\text{Enc}}}(\ell_z + \ell_g + \ell_s)$
  \State $\phi_{\text{Dec}} \stackrel{+}\leftarrow - \nabla_{\phi_{\text{Dec}}}(\ell_g + \ell_s)$
  \State $\gamma_{\text{PostNet}} \stackrel{+}\leftarrow - \nabla_{\gamma_{\text{PostNet}}}(\ell_g + \ell_z)$
  \Until static stopping criterion (epoch = 100) \Comment{Number stactic of epochs}
\end{algorithmic}
\end{algorithm}

\subsection{Setup}
\label{sec:architecture_setup}
The ResNet inspires PHGMM's architecture because it presents an extensive feature extraction process in the encoding stage.
The encoder has four blocks of depths $256$, $512$, $1024$, and $2048$.
The convolution, atrous convolution operations, and ReLU activation functions are performed within each block.
Because of our limited computational resources, we reduce our input image by $\frac{1}{4}$.
We use pooling and convolution with a stride size of two to achieve it.
To produce the parameters of the latent spaces, we concatenate the output of each block.
Thus, we use convolution and a Sigmoid function to adjust the depth (\ie, number of classes).
Then, by mean reduction over the height and width, we obtain a feature vector of each block.
We concatenate all of them and pass these through a Fully Connected layer (FC) producing our parameters ($\mu_{*}$ or $\sigma_{*}$) in size $1\times1\times256$ (empirically chosen).
Note, we use different parameters (weights) in FC to infer each cluster $k$ (\ie, $\mu^k_z$, $\sigma^k_z$).
The same configuration is also applied for PostNet.

In the decoder, we concatenate the $\mu^k_z$ of every $K$ clusters.
Then, we use FC to bring our GMM vector (from $z$) from 1D to 2D and feed our decoder with it.
The decoding stage presents operations similar to the encoder adding a bilinear interpolation operation.
Moreover, it also has four blocks, all with depth equal to the number of classes.
Note that in each block, the PHGMM model retrieves the geometric information (background and borders of the objects) by merging the local-context features $z$ with the convolution operations, upscale, and the features coming from the encoder through skip connections.
Finally, we add a vector of specific features from $\mu_g$ (passed through FC operations and bilinear interpolation) to the decoder output.
In the end, we perform two last convolution operations followed by the softmax function.
Furthermore, we set our hyperparameters $\lambda_{g, z, s}$ to $1.1$, $0.4$, and $0.4$, respectively.
Finally, we use a static stopping criterion when training our model. We define a static number of epochs (one hundred epochs) to stop the training.

\section{Experiments}
\label{sec:experiments}
In this section, we introduced datasets for SS used in the training and testing of our PHGMM model and the evaluation metrics used to compare the different existing models.
We use the dataset with reduced classes proposed by Valada \etal~\cite{Valada2019}.
It is due to our limited computational resources.
To make the comparison fair (see Table~\ref{tab:result_cityscape} and~\ref{tab:result_synthia}), we train all models from scratch and each model with the same number of classes and same resolution, and use their respective hyperparameters from each model.

\begin{figure*}[tb]
  \centering
  \resizebox{.7\linewidth}{!}{%
    \begin{tikzpicture}%
    \begin{groupplot}[
    group style={
      group size=3 by 1,
      horizontal sep=1.1cm,
      vertical sep=15pt,
      xlabels at=edge bottom,
      x descriptions at=edge bottom,
    },
    minimal plot grid,
    footnotesize, %
    cycle list/Dark2,
    cycle multiindex* list={%
      mark list\nextlist
      Dark2\nextlist
      linestyles\nextlist
    },
    /tikz/mark options={solid},
    /tikz/mark repeat=1,
    /tikz/mark phase=1,
    ymajorgrids,
    major grid style={dashed},
    xtick={2, 4, 8, 12, 16},
    xmax=17,
    xmin=1,
    x tick label style={/pgf/number format/precision=0},
    y tick label style={/pgf/number format/precision=0},
    ylabel near ticks,
    xlabel near ticks,
    legend pos=outer north center,
    legend columns=2,
    legend style={
      cells={anchor=west},
      font=\footnotesize,
      draw=none,
    },
    ]
    \nextgroupplot[%
    ymax=83,
    ymin=77,
    ylabel=IoU,
    ]%
    \addplot table[x=k, y=z, col sep=comma, header=true]{img/ablation_study_IoU_z.csv};%
    \addplot table[x=k, y=zg, col sep=comma, header=true]{img/ablation_study_IoU_z_g.csv};%

    \nextgroupplot[
    ymin=86,
    ymax=91,
    ylabel=Precision,
    xlabel={Number of cluster $k$ in $z$}, %
    ]
    \addplot table[x=k, y=z, col sep=comma, header=true]{img/ablation_study_Prec_z.csv};%
    \addlegendentry{$z$}%
    \addplot table[x=k, y=zg, col sep=comma, header=true]{img/ablation_study_Prec_z_g.csv};%
    \addlegendentry{$z+g$}%

    \nextgroupplot[%
    ymax=90,
    ymin=84,
    ylabel=Recall,
    ]%
    \addplot table[x=k, y=z, col sep=comma, header=true]{img/ablation_study_Rec_z.csv};%
    \addplot table[x=k, y=zg, col sep=comma, header=true]{img/ablation_study_Rec_z_g.csv};%
    \end{groupplot}%
    \end{tikzpicture}%
  }%
  \caption[Ablative study on $z$ and $g$]{Ablation study on the latent spaces $z$ (number of clusters $k$ in the GMM) and $g$ (include or exclude from the model) using the Cityscape validation dataset.}
  \label{fig:ablation_study_z_t}
\end{figure*}
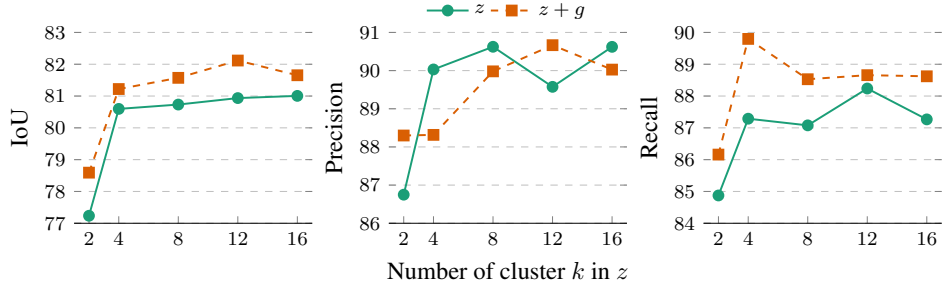

\begin{figure}[tb]
  \centering
  \resizebox{.7\linewidth}{!}{%
    \begin{tikzpicture}%
    \begin{groupplot}[
    group style={
      group size=1 by 2,
      xlabels at=edge bottom,
      ylabels at=edge left,
      x descriptions at=edge bottom,
      y descriptions at=edge left,
      horizontal sep=1.0cm,
      vertical sep=45pt,
    },
    footnotesize,
    height=4cm,
    width=6.8cm,
    minimal plot grid,
    cycle list/Dark2,
    cycle multiindex* list={%
      mark list\nextlist
      Dark2\nextlist
      linestyles\nextlist
    },
    /tikz/mark options={solid},
    /tikz/mark repeat=3,
    /tikz/mark phase=2,
    ymajorgrids,
    major grid style={dashed},
    ylabel={Pixelwise Class.\ Error ($\%$)},
    xlabel={Trimap Width (in pixels)},
    xtick={1, 10, 20, 30},
    xmax=30,
    xmin=0,
    x tick label style={/pgf/number format/precision=0},
    y tick label style={/pgf/number format/precision=0},
    ylabel near ticks,
    xlabel near ticks,
    legend columns=3,
    legend style={
      cells={anchor=west},
      font=\scriptsize,
      draw=none,
    },
    ]

    \nextgroupplot[%
    ytick={25, 36, 47, 58},
    yticklabels={25, 36, 47, 58},
    ymax=60,
    ymin=25,
    legend to name=legend2,
    xlabel style={align=center},
    xlabel={Trimap Width (in pixels)\\Cytiscapes},
    ]%
    \coordinate (c1) at (rel axis cs:1, 1);

    \addplot table[x=width, y=trimap, col sep=comma, header=true]{img/cityscape_adapnet_pp.csv};%
    \addlegendentry{Adapnet++}%

    \addplot table[x=width, y=trimap, col sep=comma, header=true]{img/cityscape_ocnet.csv};%
    \addlegendentry{OCNet}%

    \addplot table[x=width, y=trimap, col sep=comma, header=true]{img/cityscape_dunet.csv};%
    \addlegendentry{DUNet}%

    \addplot table[x=width, y=trimap, col sep=comma, header=true]{img/cityscape_hrnet.csv};%
    \addlegendentry{HRNet}%

    \addplot table[x=width, y=trimap, col sep=comma, header=true]{img/cityscape_hrnet+ocr.csv};%
    \addlegendentry{HRNet+OCR}%

    \addplot table[x=width, y=trimap, col sep=comma, header=true]{img/cityscape_phgmm.csv};%
    \addlegendentry{PHGMM}%

    \nextgroupplot[%
    yticklabels={21, 31, 40, 49, 58},
    ytick={21, 31, 40, 49, 58},
    ymax=60,
    ymin=21,
    legend to name=legend,
    xlabel style={align=center},
    xlabel={Trimap Width (in pixels)\\Synthia},
    ]%
    \addplot table[x=width, y=trimap, col sep=comma, header=true]{img/synthia_adapnet_pp.csv};%
    \addlegendentry{Adapnet++}%

    \addplot table[x=width, y=trimap, col sep=comma, header=true]{img/synthia_ocnet.csv};%
    \addlegendentry{OCNet}%

    \addplot table[x=width, y=trimap, col sep=comma, header=true]{img/synthia_dunet.csv};%
    \addlegendentry{DUNet}%

    \addplot table[x=width, y=trimap, col sep=comma, header=true]{img/synthia_hrnet.csv};%
    \addlegendentry{HRNet}%

    \addplot table[x=width, y=trimap, col sep=comma, header=true]{img/synthia_hrnet+ocr.csv};%
    \addlegendentry{HRNet+OCR}%

    \addplot table[x=width, y=trimap, col sep=comma, header=true]{img/synthia_phgmm.csv};%
    \addlegendentry{PHGMM}%

    \end{groupplot}%
    \node[above] at ($(c1.north)+(-80pt, 20pt)$) {\pgfplotslegendfromname{legend}};
    \end{tikzpicture}%
  }%
  \caption[Pixelwise classification error \vs trimap width]{%
    Pixelwise classification error (lower is better) \vs trimap width (to measure the pixel boundary labeling accuracy) for AdapNet++~\cite{Valada2019}, OCNet~\cite{Yuan2018}, DUNet~\cite{Jin2019}, HRNet~\cite{Wang2020}, HRNet+OCR~\cite{Yuan2021}, and our PHGMM on the validation datasets.
  }
  \label{fig:cityscape_trimap}
\end{figure}
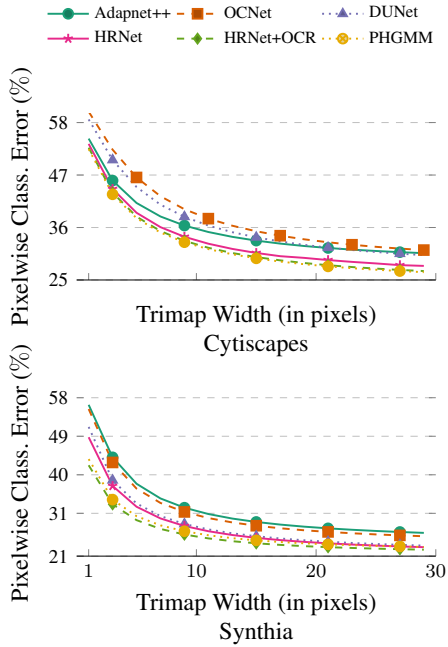

\subsection{Datasets}
\label{sec:datasets}
We evaluated several types of urban/forest scenarios datasets. They are Cityscapes~\cite{Cordts2016} and Synthia~\cite{Ros2016}.

\textbf{Cityscapes:} The dataset has \num{5000} samples.
However, we used crop for augmenting the training set (\ie, transformations of contrast, brightness, and horizontal flip) to generate \num{17500} samples.
For comparison with other models,  we employ the original validation set with a resolution of $768\times384$ (resize).
To facilitate the models comparison, AdapNet++~\cite{Valada2016} proposed the class reduction produced by combining some types of classes.
Thus, we use \num{11} classes: \textit{sky, building, road, sidewalk, fence, vegetation, pole, car/truck/bus, traffic sign, person, rider/bicycle/motorbike, and background}.

\textbf{Synthia:} This dataset contains realistic photo images from a virtual city. We use the original \num{9000} samples (\num{7000} for training and \num{2000} testing) resized to $768\times384$ resolution.
The classes of object labels are the same as the Cityscapes mentioned above label set.

\subsection{Evaluation Metrics}
\label{sec:evaluation_metrics}
To evaluate our results on segmentation, we chose accuracy and intersection-over-union metrics as validation measures~\cite{Csurka2013}.
The intersection-over-union (IoU) is defined by
\begin{equation}
\label{eq:metric_iou}
\text{IoU} = \sum_{i}^N \frac{P_i \cap Y_i}{P_i \cup Y_i} = \sum_{i}^N \frac{\mathit{TP}_i}{\mathit{TP}_i + \mathit{FP}_i + \mathit{FN}_i},
\end{equation}
the precision (Prec) is
\begin{equation}
\label{eq:metric_prec}
\text{Prec} = \sum_{i}^N \frac{\mathit{TP}_i}{\mathit{TP}_i + \mathit{FP}_i},
\end{equation}
and the recall (Rec) is
\begin{equation}
\label{eq:metric_rec}
\text{Rec} = \sum_{i}^N \frac{\mathit{TP}_i}{\mathit{TP}_i + \mathit{FN}_i}.
\end{equation}
We assume that $P_i$ is the set of pixels predicted as the $i$th class, $Y_i$ is pixels set belonging to the $i$th class, and $N$ is the number of classes.
Besides, $\mathit{TP}_i$, $\mathit{FP}_i$, $\mathit{TN}_i$, and $\mathit{FN}_i$ represent True/False Positives and True/False Negatives, respectively, for a given class $i$.
Note, these metrics are widely used in SS\@

We use clustering metrics to measure the latent space behavior.
Consequently, we utilize the Silhouette Coefficient, SSI~\cite{Rousseeuw1987} defined by
\begin{equation}
\label{eq:metric_ssi}
\text{SSI} = \sum_{i}^N \frac{b_i - a_i}{\max\{a_i, b_i\}},
\end{equation}
where $a_i$, $b_i$ are the mean intra-cluster and nearest-cluster distances from $i$, respectively.
The Calinski-Harabasz Index, CHI~\cite{Calinski1974} is given by
\begin{equation}
\label{eq:metric_chi}
\text{CHI} = \frac{\text{SS}_M}{\text{SS}_W} \frac{N-k}{k-1},
\end{equation}
where $k$ is the number of clusters, and $N$ is the samples, $\text{SS}_W$ and $\text{SS}_M$ are the overall within-cluster and between-cluster variances, respectively.
Finally, the Davies-Bouldin Index, DBI~\cite{Davies1979} denoted by
\begin{equation}
\label{eq:metric_dbi}
\text{DBI} = \frac{1}{k}\sum_i^k \max_{j \ne i}\left(\frac{s_i + s_j}{d_{ij}}\right),
\end{equation}
where $s_i$ is the average distance between each point of cluster $i$ and its centroid, and $d_{ij}$ is the distance between cluster centroids $i$ and $j$.
Note, high values are better for SSI and CHI, while low ones are better for DBI\@.

\subsection{Comparisons}
\label{sec:Comparisons}

\begin{figure*}[ht!] %
\centering
\resizebox{\linewidth}{!}{%
\newlength{\hsz}
\setlength{\hsz}{.15\linewidth}
\setlength{\colfig}{2pt}
{\renewcommand{\arraystretch}{0}
\begin{tabular}{%
  @{}%
  c@{\hspace{\colfig}}
  c@{\hspace{\colfig}}
  c@{\hspace{\colfig}}
  c@{\hspace{\colfig}}
  c@{\hspace{\colfig}}
  c@{\hspace{\colfig}}
  c@{\hspace{\colfig}}
  c@{\hspace{\colfig}}
  c@{}
}
  \footnotesize\multirow{6}{*}{\raisebox{-5.7\normalbaselineskip}[0pt][0pt]{\rotatebox[origin=c]{90}{\scriptsize Cityscapes dataset}}} &
  {\raisebox{1.15\normalbaselineskip}[0pt][0pt]{\rotatebox[origin=c]{90}{\tiny Output}}} &
  \includegraphics[width=\hsz]{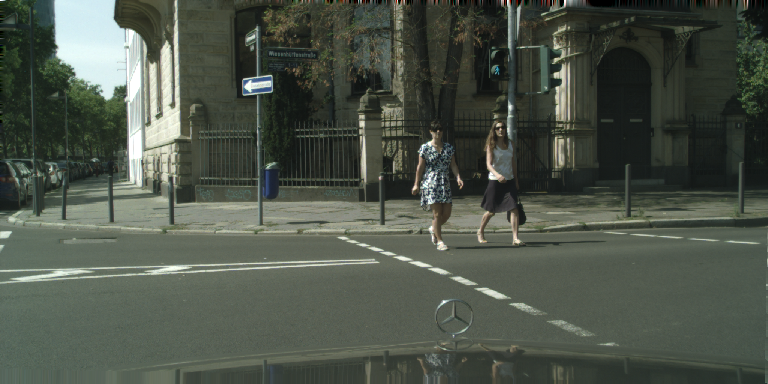} &
  \includegraphics[width=\hsz]{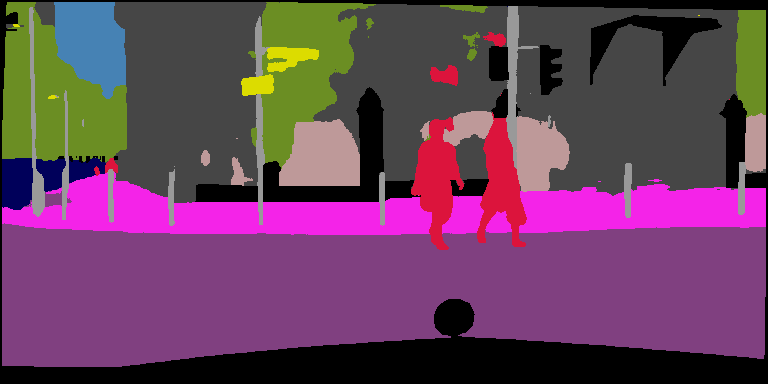} &
  \includegraphics[width=\hsz]{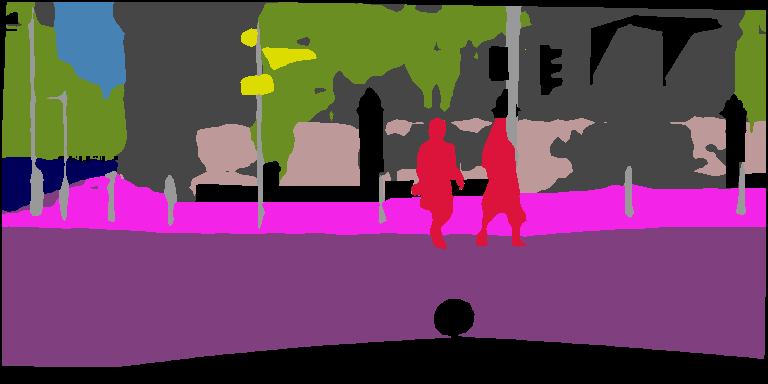} &
  \includegraphics[width=\hsz]{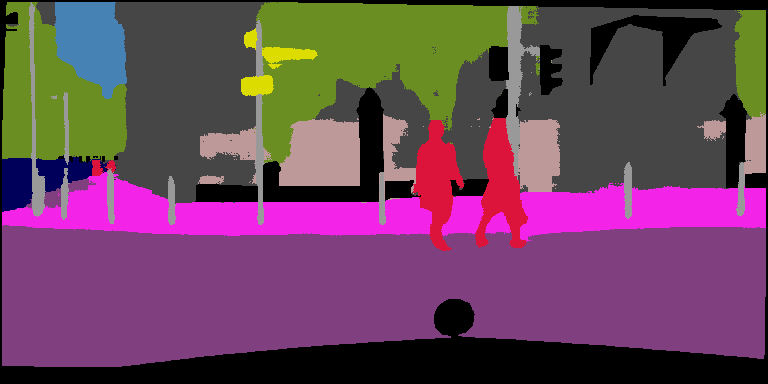} &
  \includegraphics[width=\hsz]{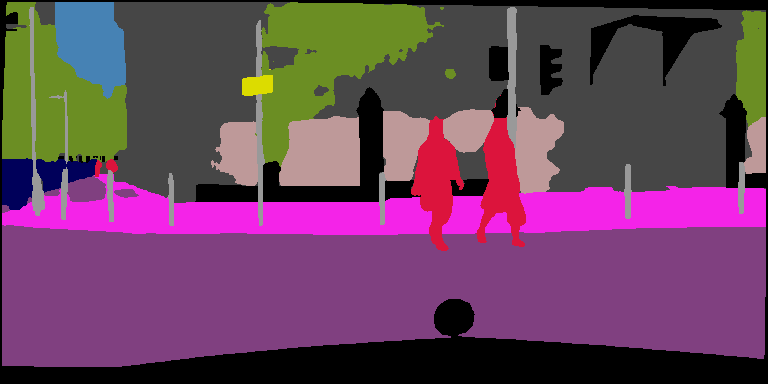} &
  \includegraphics[width=\hsz]{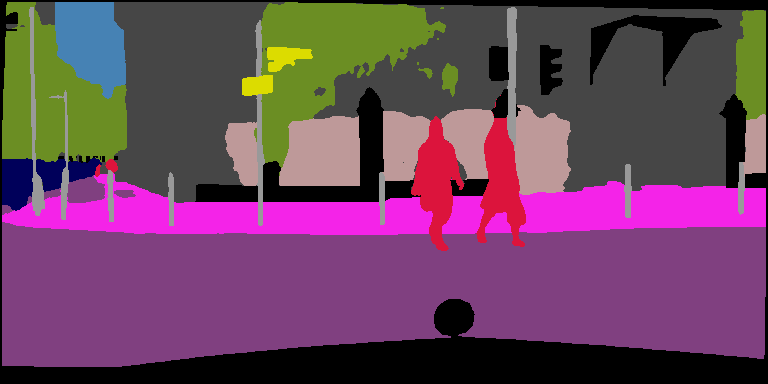} &
  \includegraphics[width=\hsz]{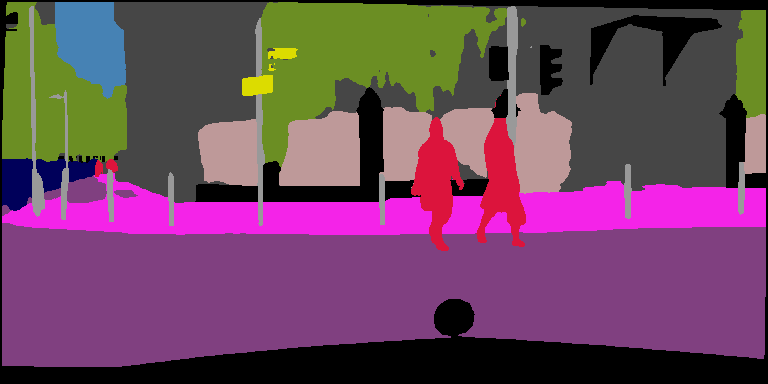}\\[.2mm]
  &
  {\raisebox{1.2\normalbaselineskip}[0pt][0pt]{\rotatebox[origin=c]{90}{\tiny Compare}}} &
  \includegraphics[width=\hsz]{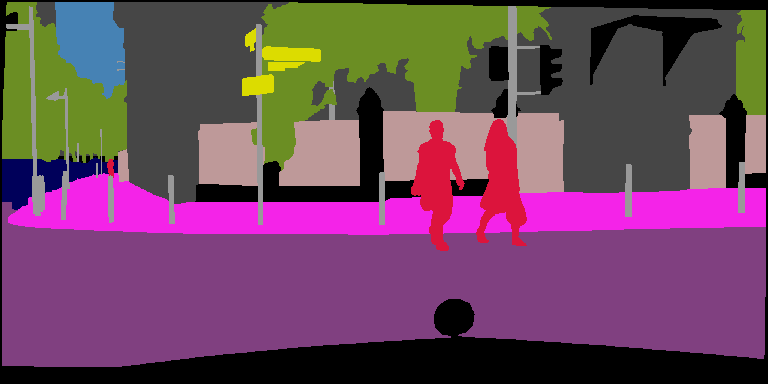} &
  \includegraphics[width=\hsz]{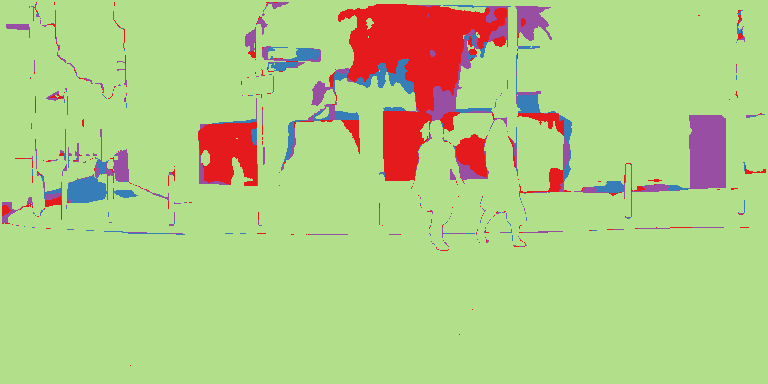} &
  \includegraphics[width=\hsz]{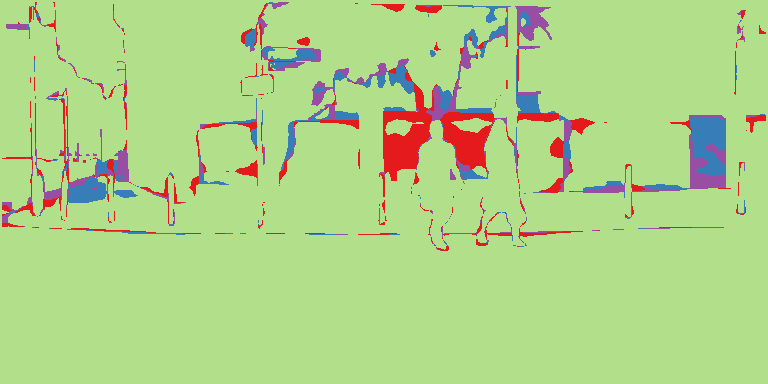} &
  \includegraphics[width=\hsz]{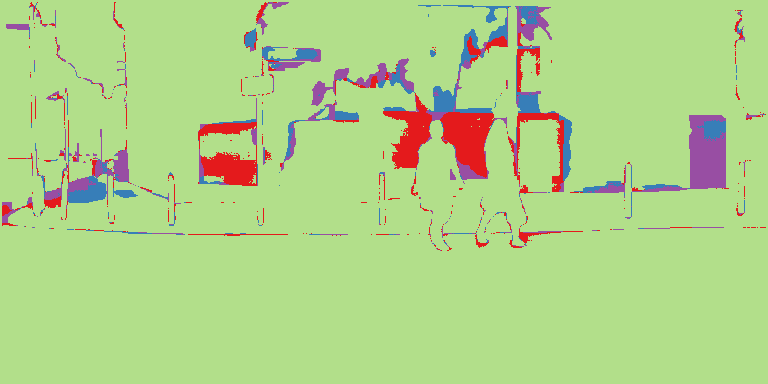} &
  \includegraphics[width=\hsz]{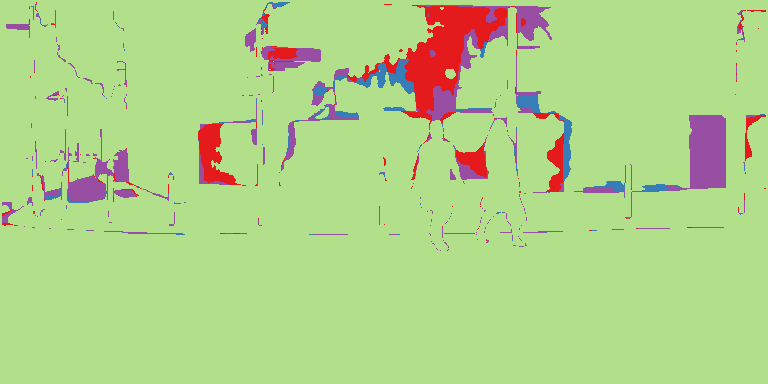} &
  \includegraphics[width=\hsz]{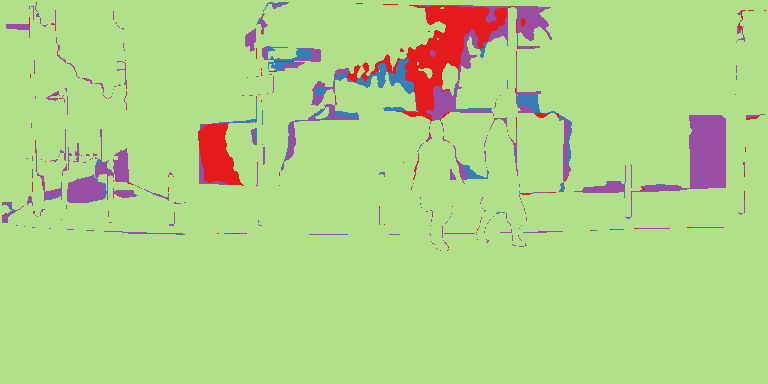} &
  \includegraphics[width=\hsz]{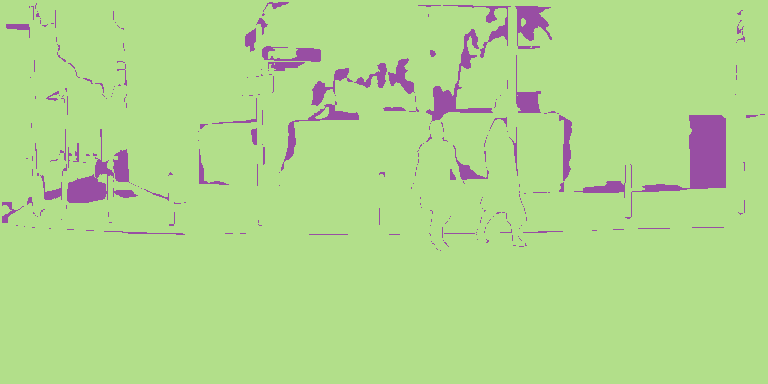}\\[.2mm]
  &
  {\raisebox{1.15\normalbaselineskip}[0pt][0pt]{\rotatebox[origin=c]{90}{\tiny Output}}} &
  \includegraphics[width=\hsz]{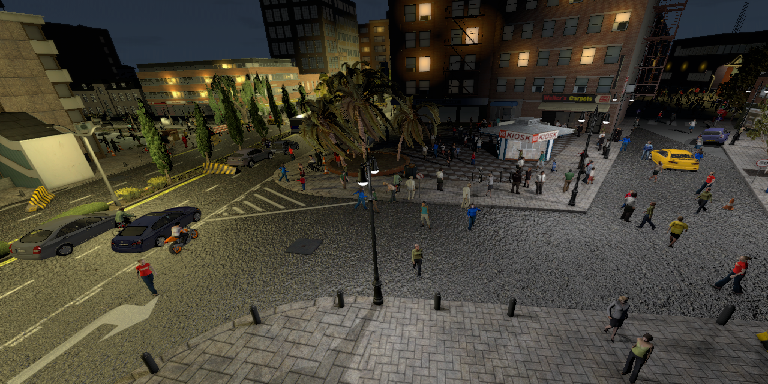} &
  \includegraphics[width=\hsz]{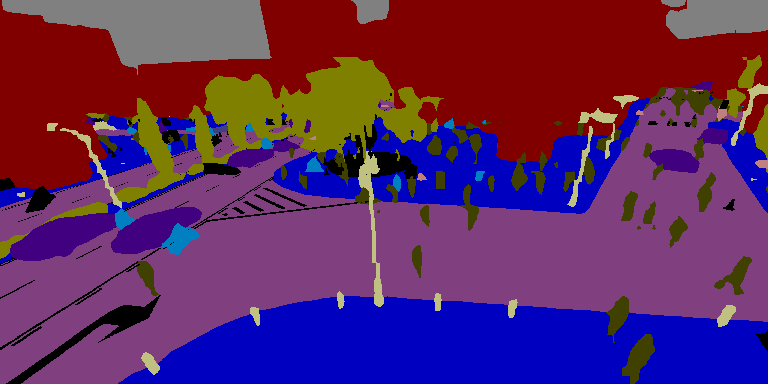} &
  \includegraphics[width=\hsz]{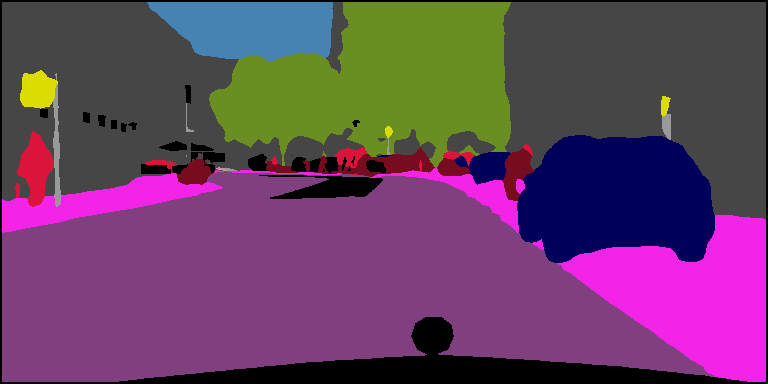} &
  \includegraphics[width=\hsz]{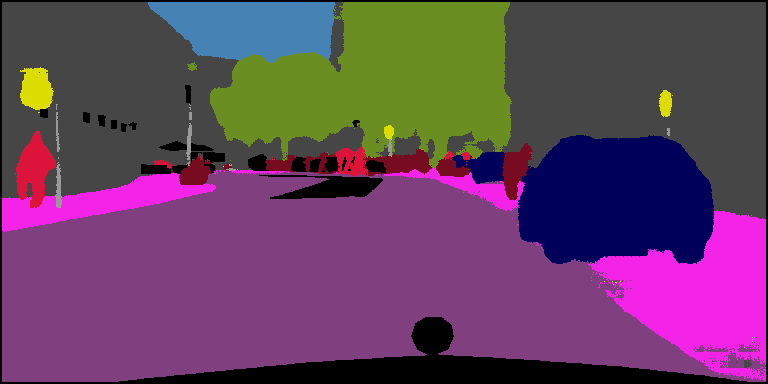} &
  \includegraphics[width=\hsz]{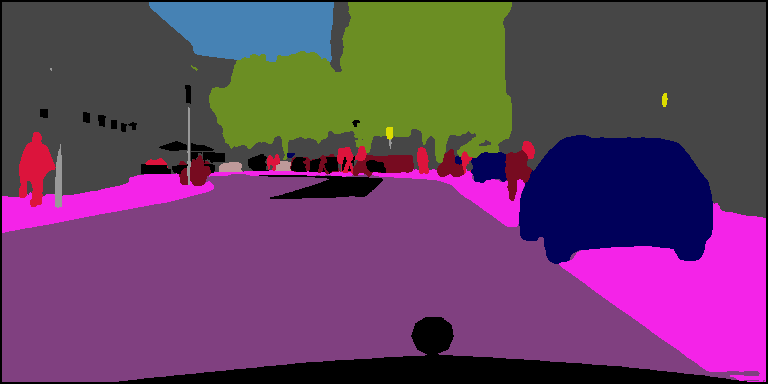} &
  \includegraphics[width=\hsz]{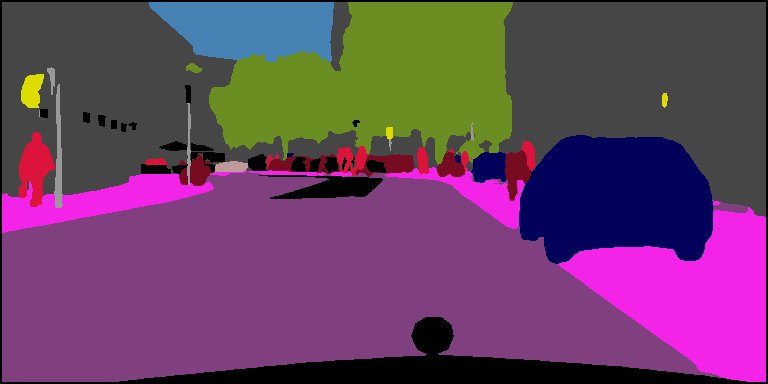} &
  \includegraphics[width=\hsz]{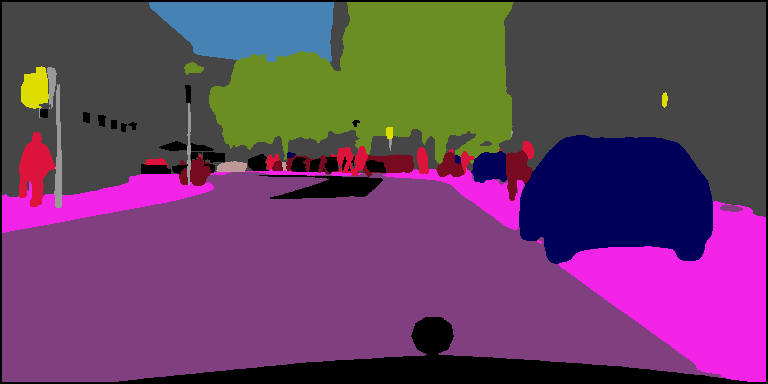}\\[.2mm]
  &
  {\raisebox{1.2\normalbaselineskip}[0pt][0pt]{\rotatebox[origin=c]{90}{\tiny Compare}}} &
  \includegraphics[width=\hsz]{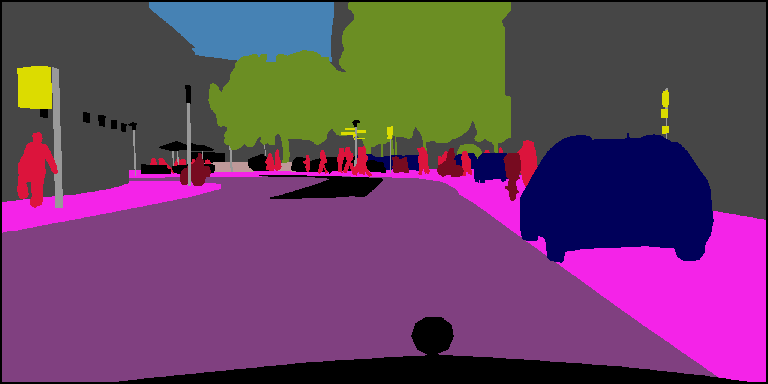} &
  \includegraphics[width=\hsz]{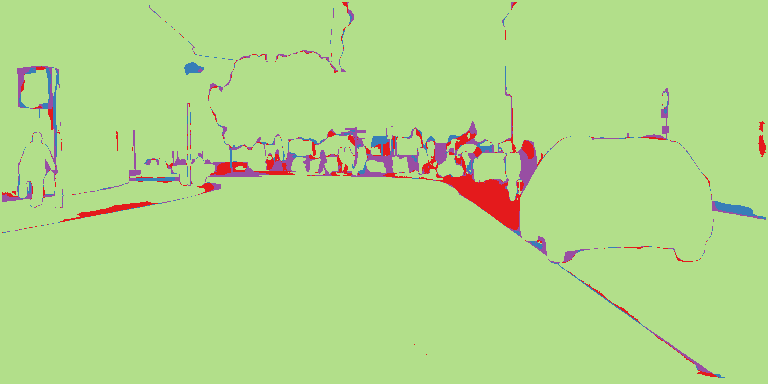} &
  \includegraphics[width=\hsz]{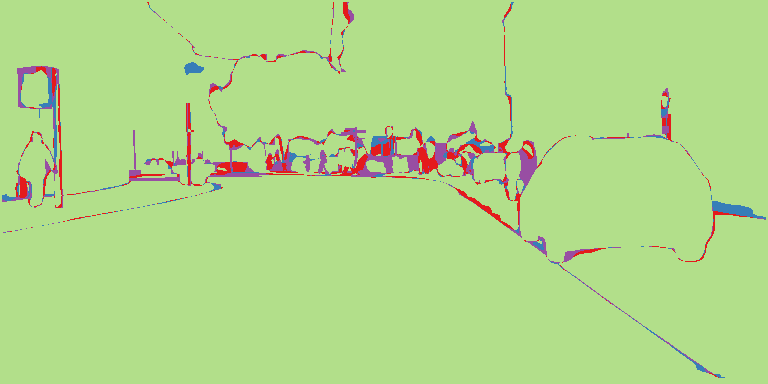} &
  \includegraphics[width=\hsz]{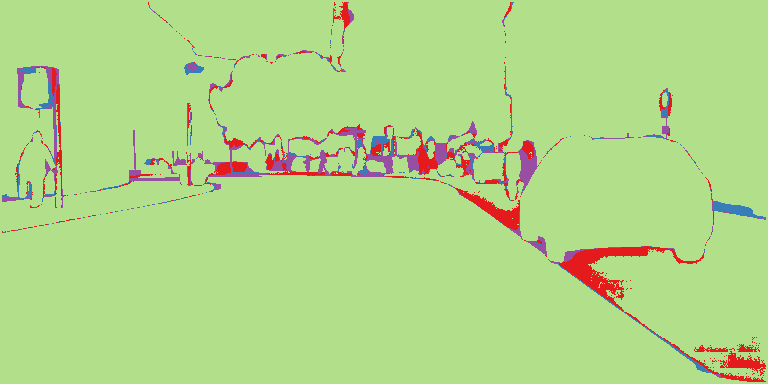} &
  \includegraphics[width=\hsz]{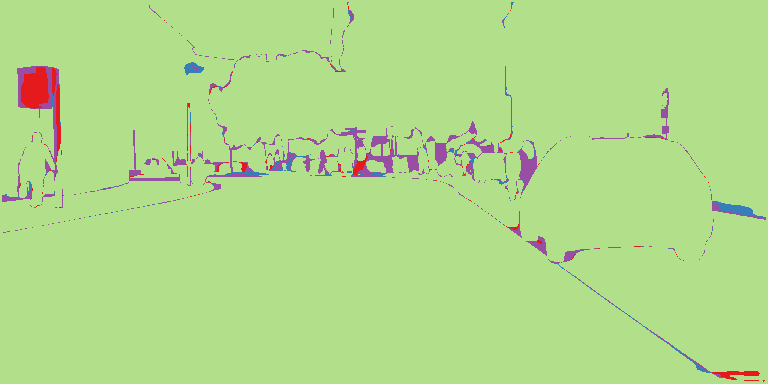} &
  \includegraphics[width=\hsz]{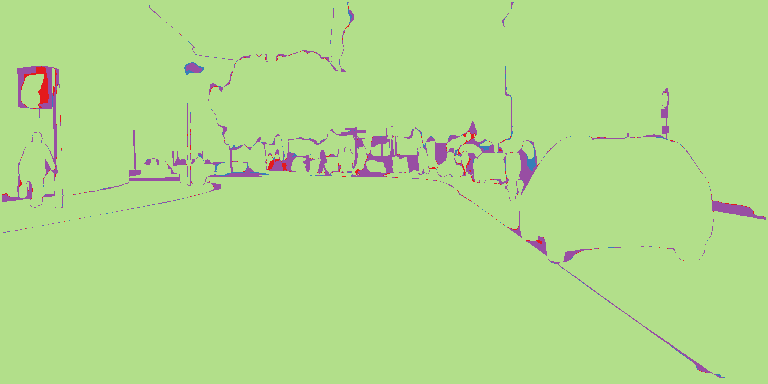} &
  \includegraphics[width=\hsz]{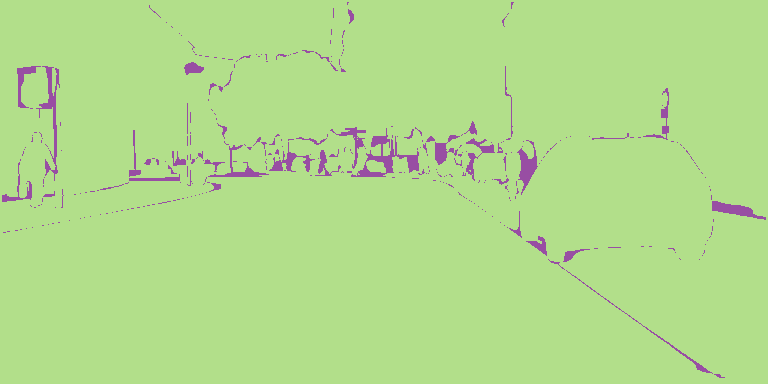}\\
  &
  {\raisebox{1.15\normalbaselineskip}[0pt][0pt]{\rotatebox[origin=c]{90}{\tiny Output}}} &
  \includegraphics[width=\hsz]{img_frankfurt_000000_012009_leftImg8bit} &
  \includegraphics[width=\hsz]{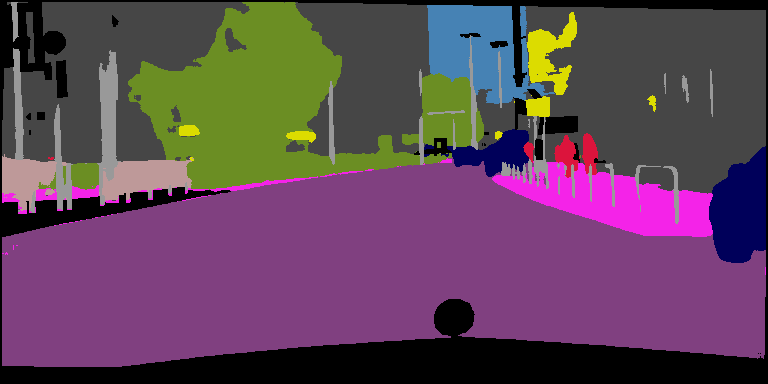} &
  \includegraphics[width=\hsz]{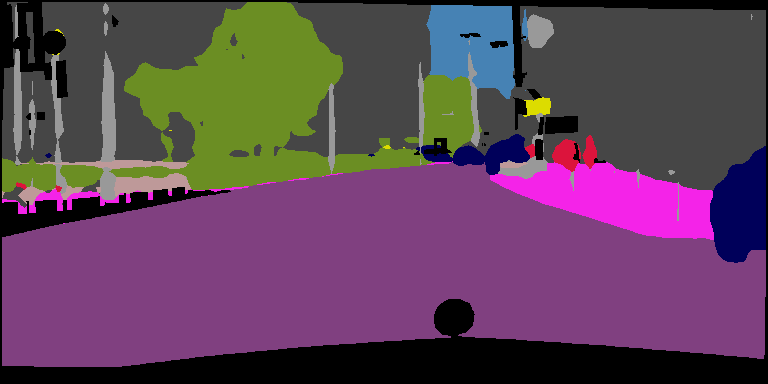} &
  \includegraphics[width=\hsz]{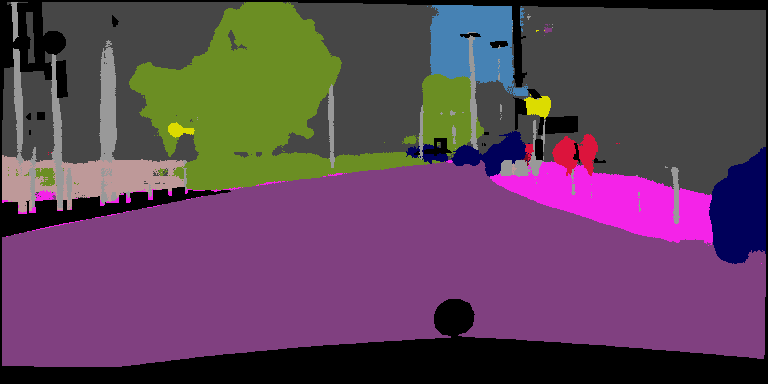} &
  \includegraphics[width=\hsz]{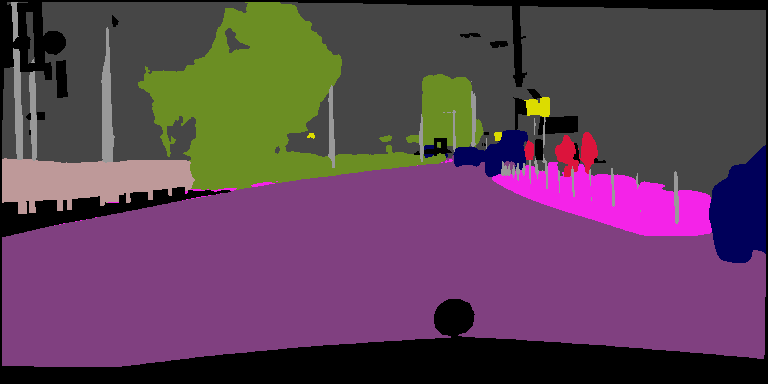} &
  \includegraphics[width=\hsz]{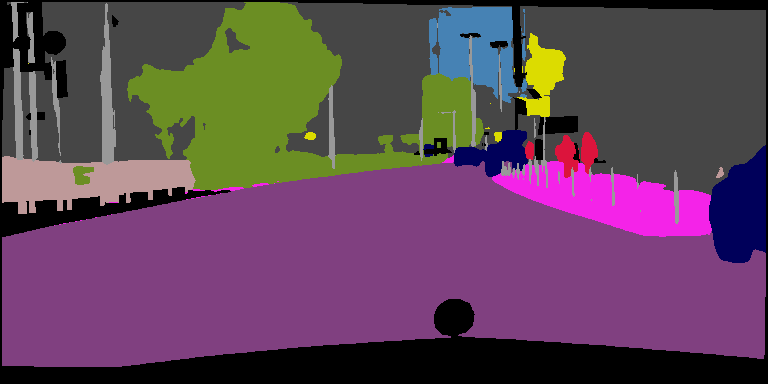} &
  \includegraphics[width=\hsz]{phgmm_frankfurt_000000_012009_leftImg8bit}\\[.2mm]
  &
  {\raisebox{1.2\normalbaselineskip}[0pt][0pt]{\rotatebox[origin=c]{90}{\tiny Compare}}} &
  \includegraphics[width=\hsz]{gt_frankfurt_000000_012009_leftImg8bit} &
  \includegraphics[width=\hsz]{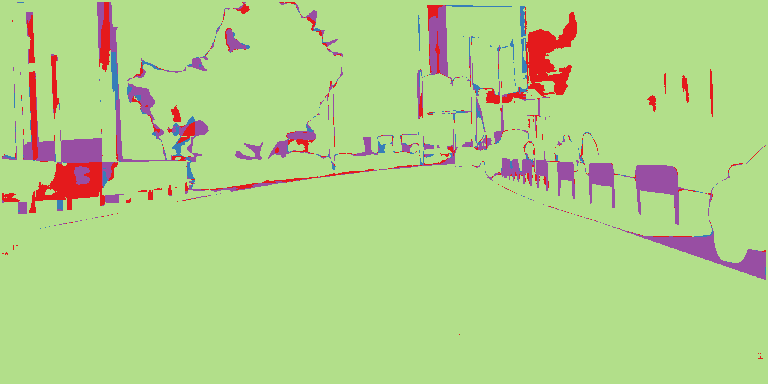} &
  \includegraphics[width=\hsz]{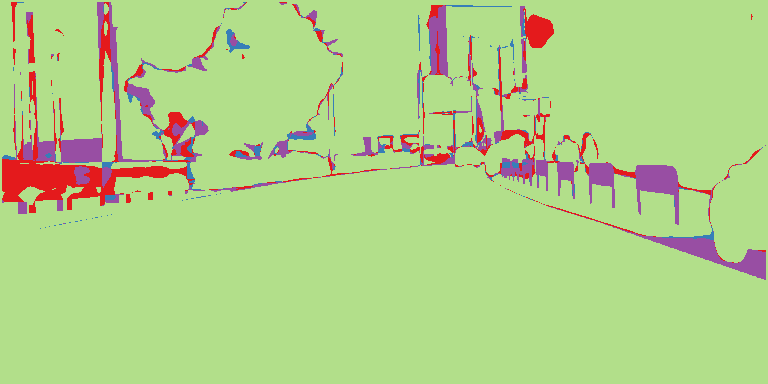} &
  \includegraphics[width=\hsz]{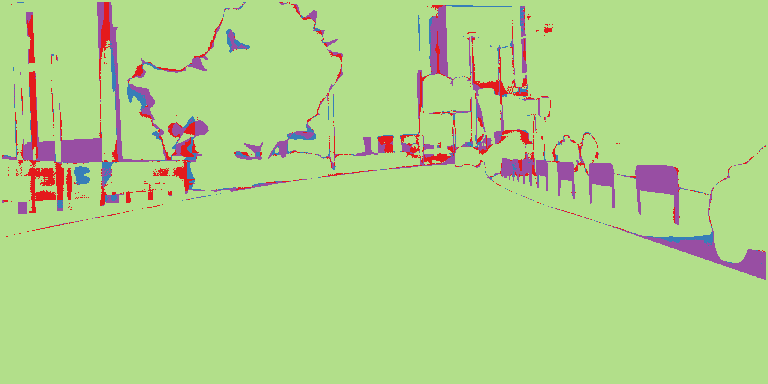} &
  \includegraphics[width=\hsz]{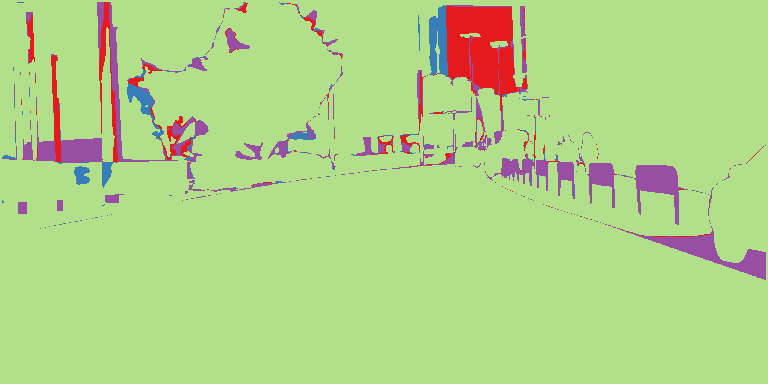} &
  \includegraphics[width=\hsz]{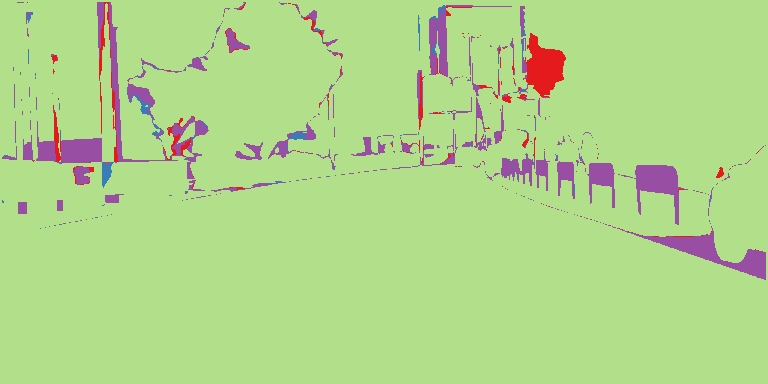} &
  \includegraphics[width=\hsz]{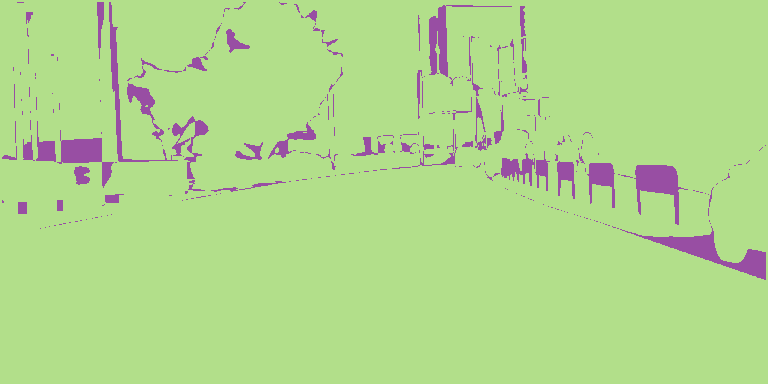}\\
  &
  {\raisebox{1.15\normalbaselineskip}[0pt][0pt]{\rotatebox[origin=c]{90}{\tiny Output}}} &
  \includegraphics[width=\hsz]{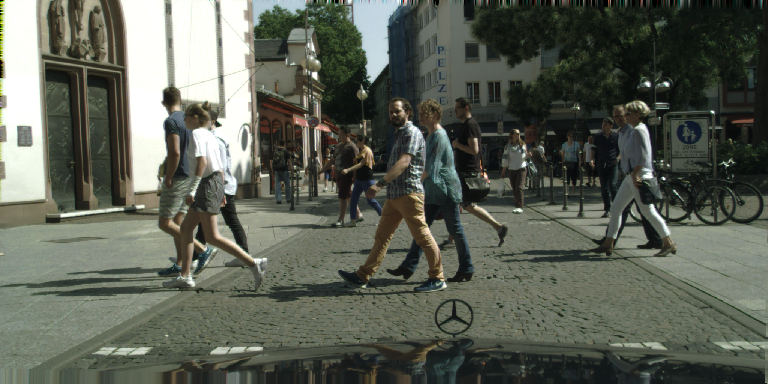} &
  \includegraphics[width=\hsz]{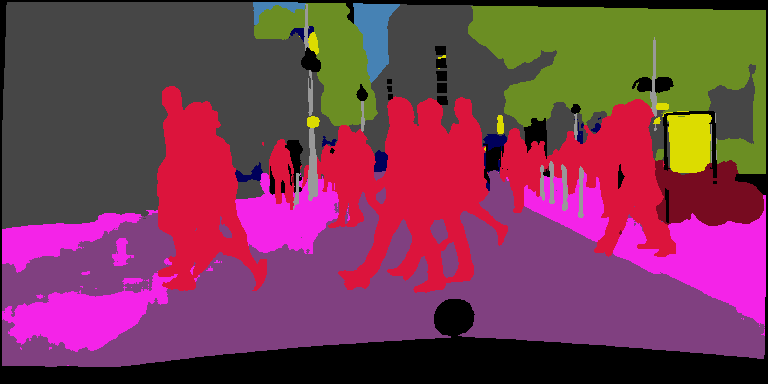} &
  \includegraphics[width=\hsz]{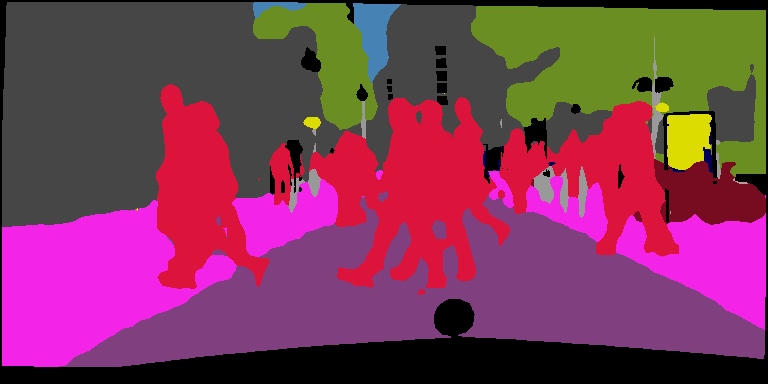} &
  \includegraphics[width=\hsz]{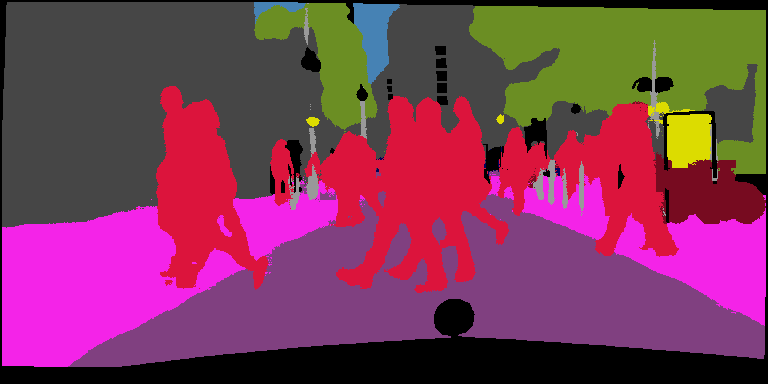} &
  \includegraphics[width=\hsz]{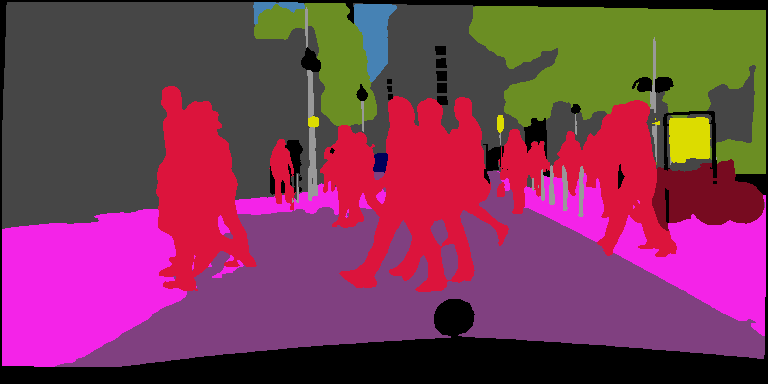} &
  \includegraphics[width=\hsz]{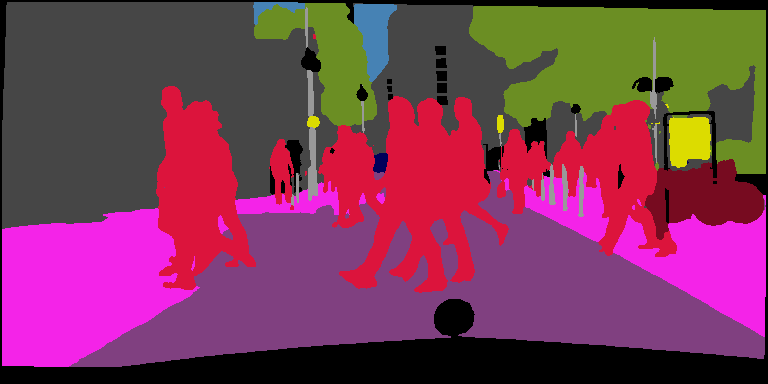} &
  \includegraphics[width=\hsz]{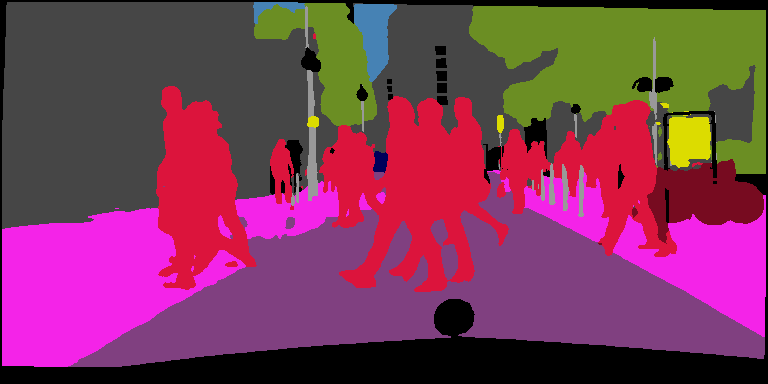}\\[.2mm]
  &
  {\raisebox{1.2\normalbaselineskip}[0pt][0pt]{\rotatebox[origin=c]{90}{\tiny Compare}}} &
  \includegraphics[width=\hsz]{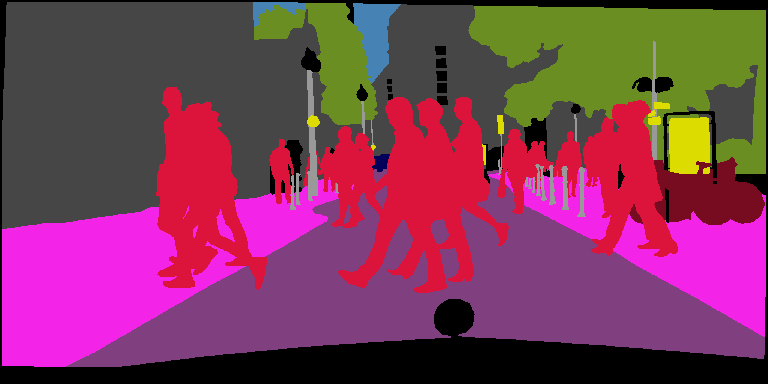} &
  \includegraphics[width=\hsz]{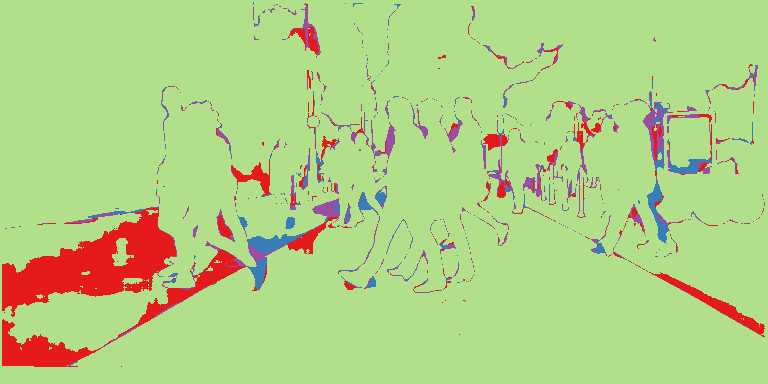} &
  \includegraphics[width=\hsz]{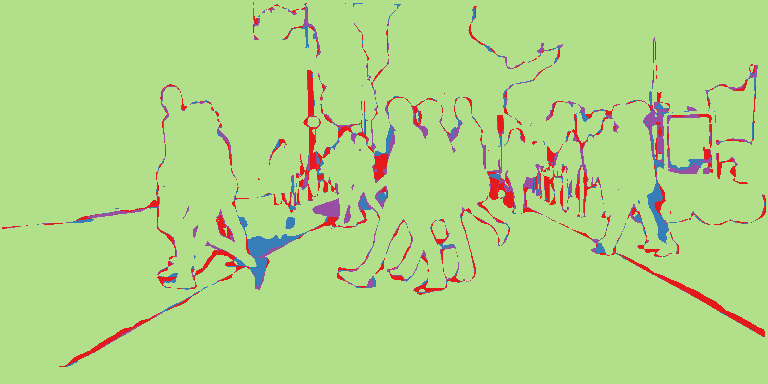} &
  \includegraphics[width=\hsz]{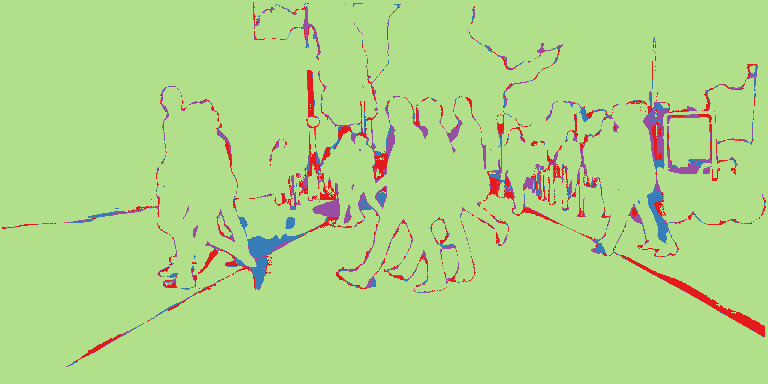} &
  \includegraphics[width=\hsz]{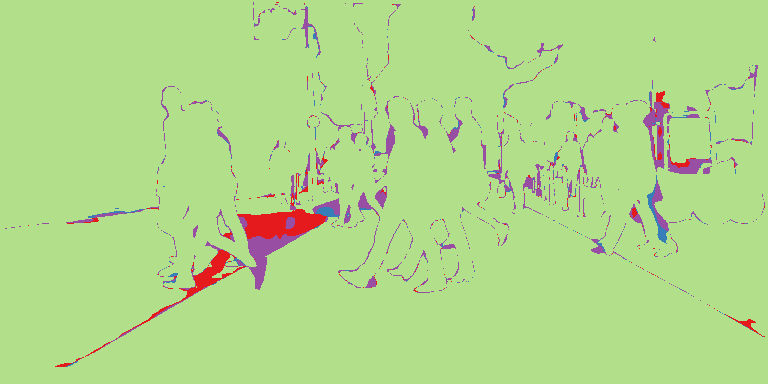} &
  \includegraphics[width=\hsz]{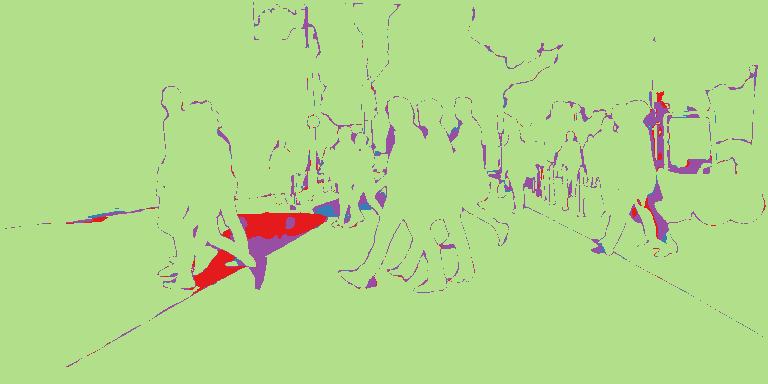} &
  \includegraphics[width=\hsz]{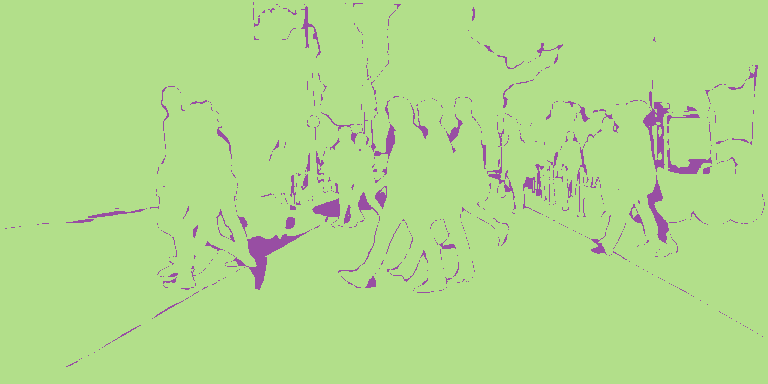}\\
  \\[1.0mm]

  \footnotesize\multirow{6}{*}{\raisebox{-5.7\normalbaselineskip}[0pt][0pt]{\rotatebox[origin=c]{90}{\scriptsize Synthia dataset}}} &
  {\raisebox{1.15\normalbaselineskip}[0pt][0pt]{\rotatebox[origin=c]{90}{\tiny Output}}} &
  \includegraphics[width=\hsz]{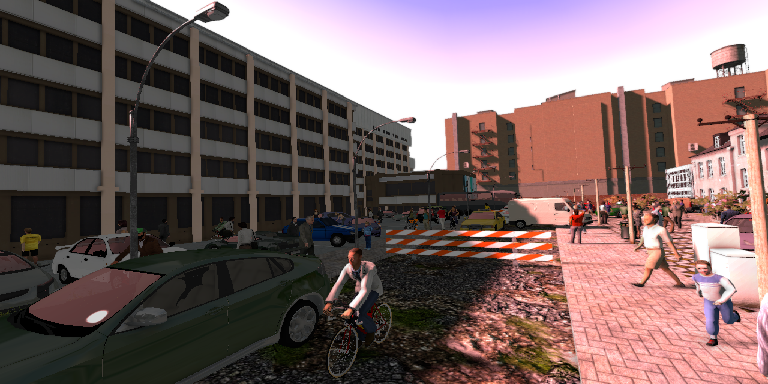} &
  \includegraphics[width=\hsz]{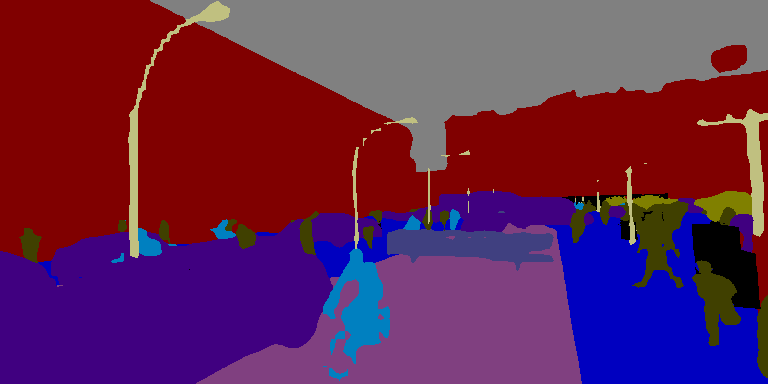} &
  \includegraphics[width=\hsz]{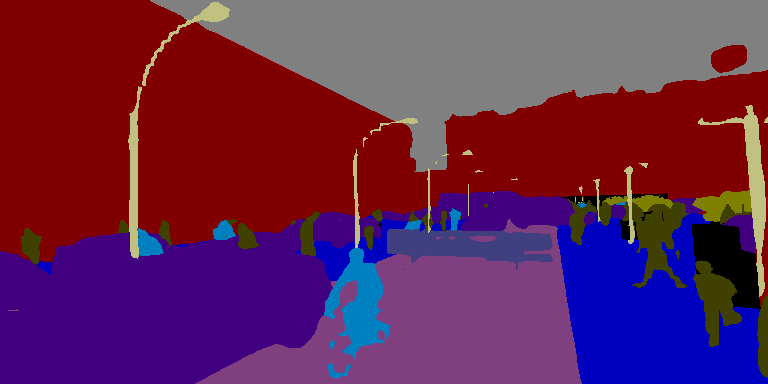} &
  \includegraphics[width=\hsz]{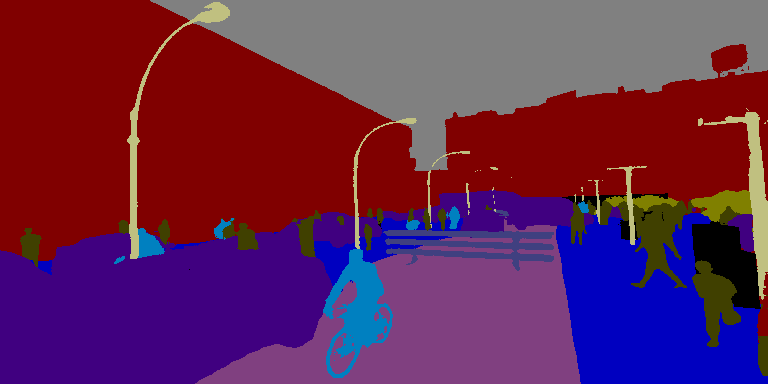} &
  \includegraphics[width=\hsz]{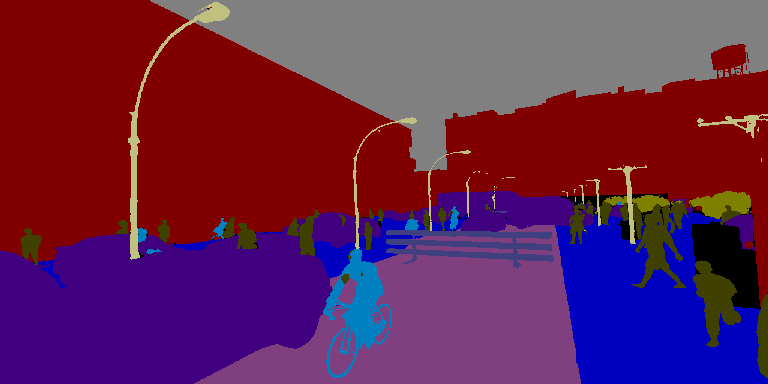} &
  \includegraphics[width=\hsz]{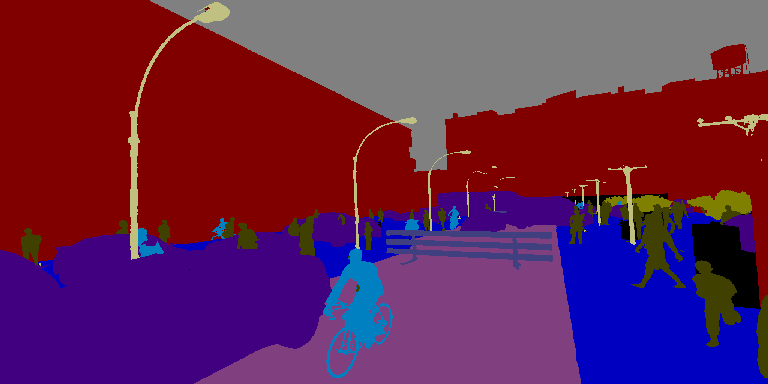} &
  \includegraphics[width=\hsz]{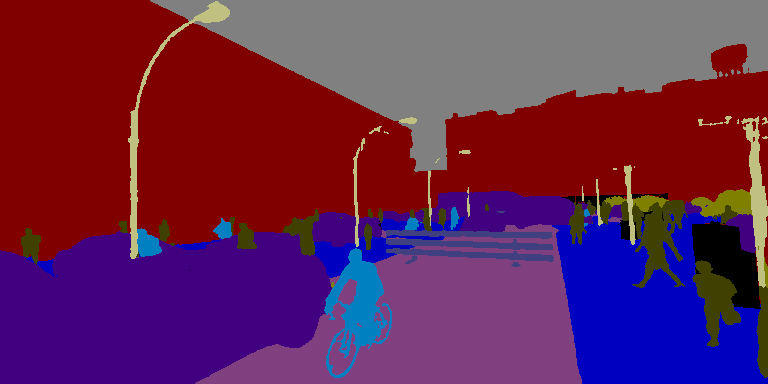}\\[.2mm]
  &
  {\raisebox{1.2\normalbaselineskip}[0pt][0pt]{\rotatebox[origin=c]{90}{\tiny Compare}}} &
  \includegraphics[width=\hsz]{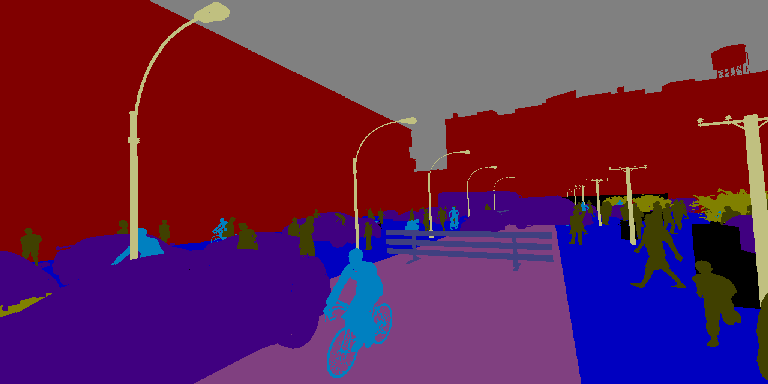} &
  \includegraphics[width=\hsz]{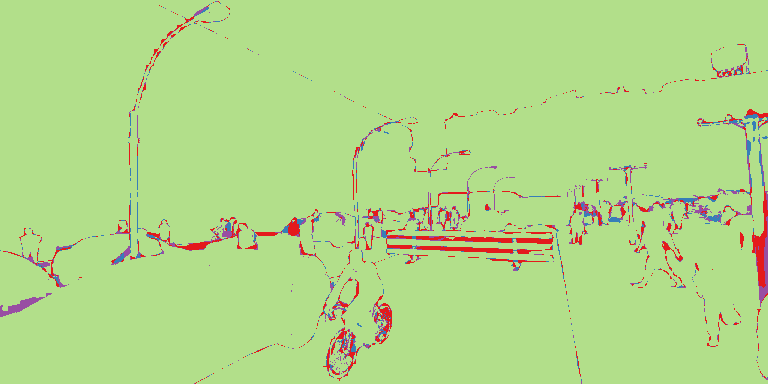} &
  \includegraphics[width=\hsz]{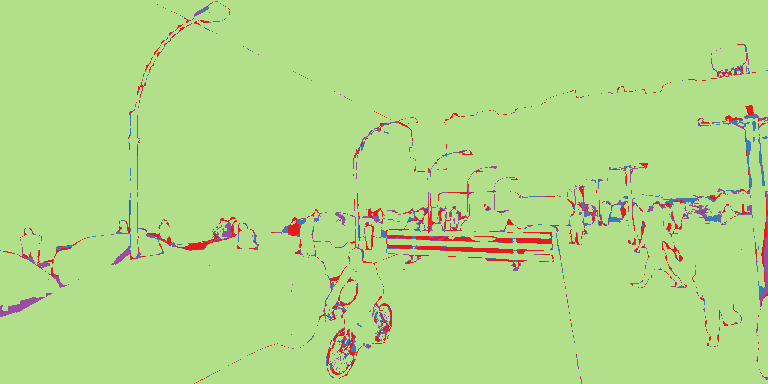} &
  \includegraphics[width=\hsz]{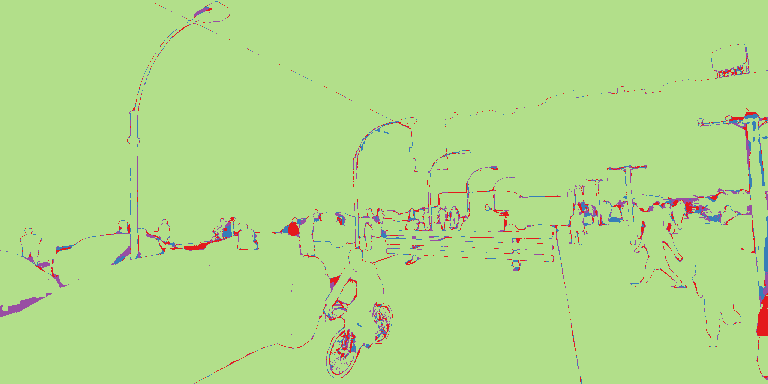} &
  \includegraphics[width=\hsz]{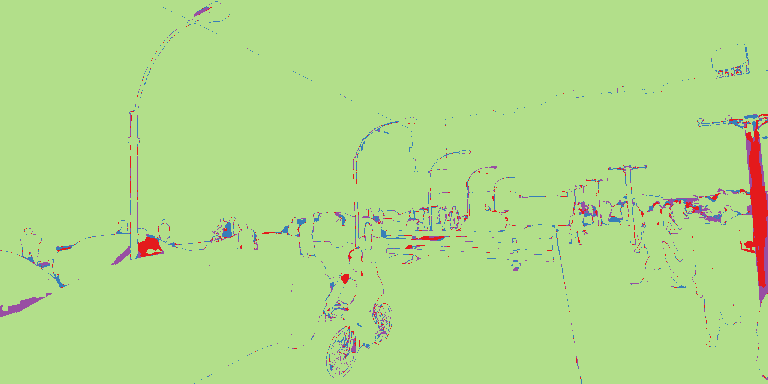} &
  \includegraphics[width=\hsz]{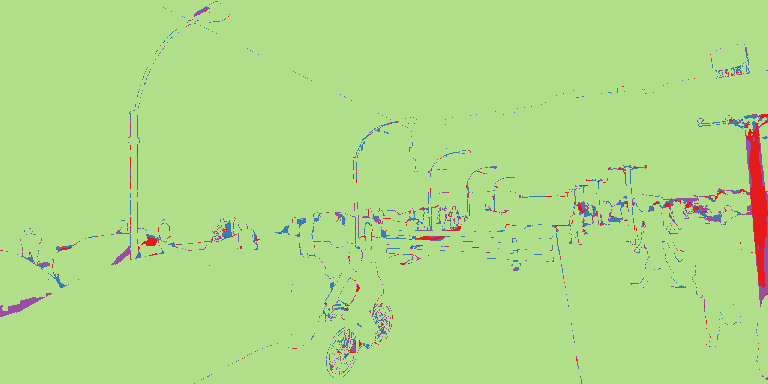} &
  \includegraphics[width=\hsz]{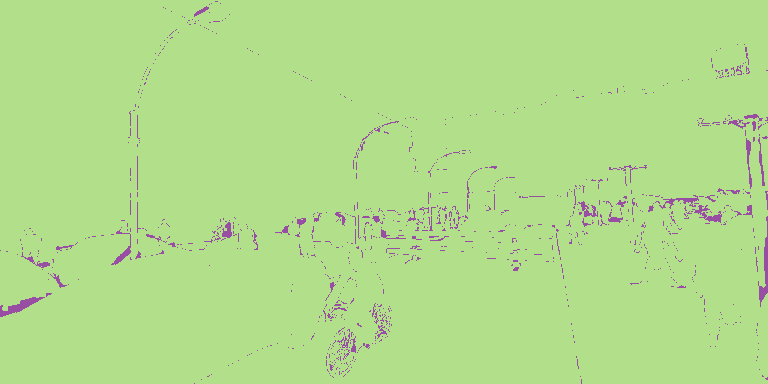}\\[.2mm]
  &
  {\raisebox{1.15\normalbaselineskip}[0pt][0pt]{\rotatebox[origin=c]{90}{\tiny Output}}} &
  \includegraphics[width=\hsz]{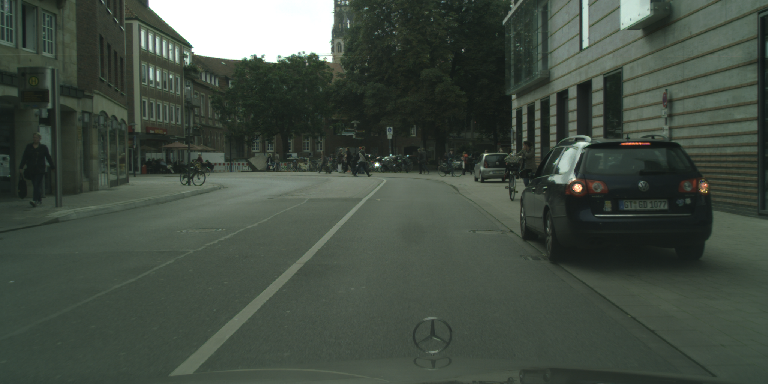} &
  \includegraphics[width=\hsz]{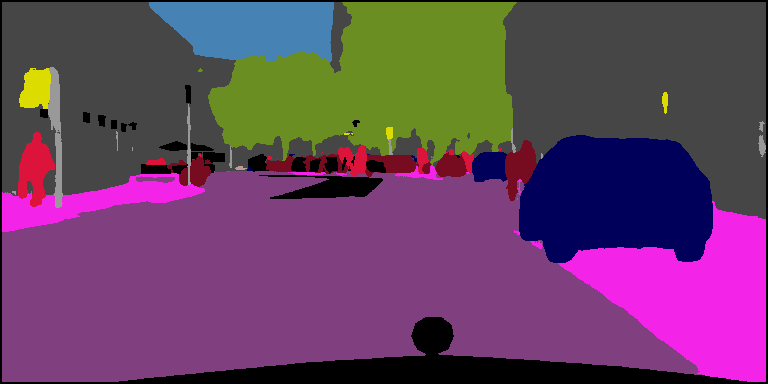} &
  \includegraphics[width=\hsz]{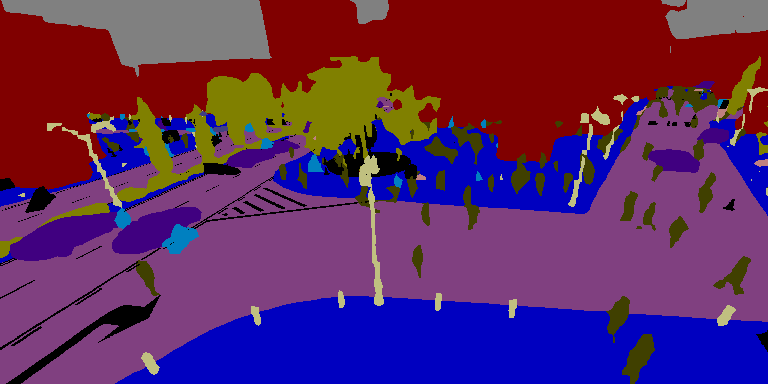} &
  \includegraphics[width=\hsz]{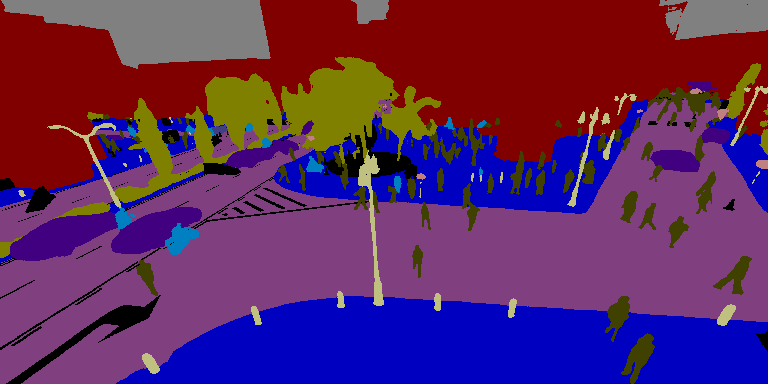} &
  \includegraphics[width=\hsz]{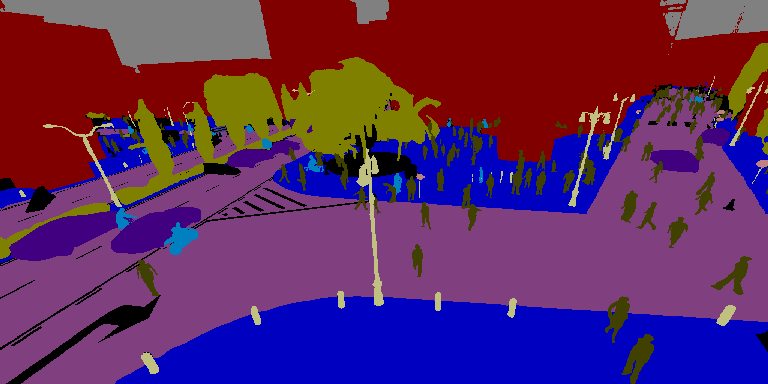} &
  \includegraphics[width=\hsz]{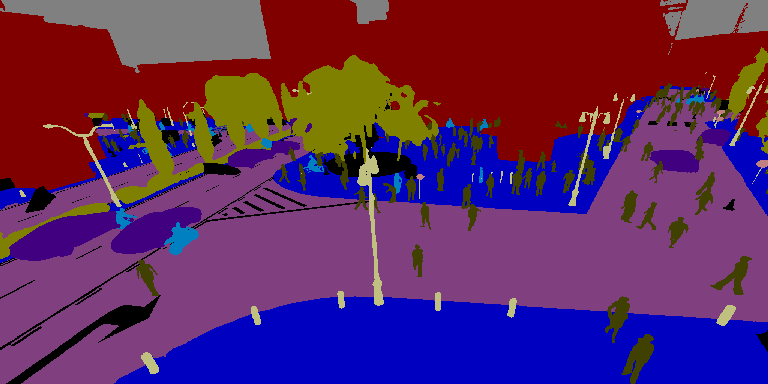} &
  \includegraphics[width=\hsz]{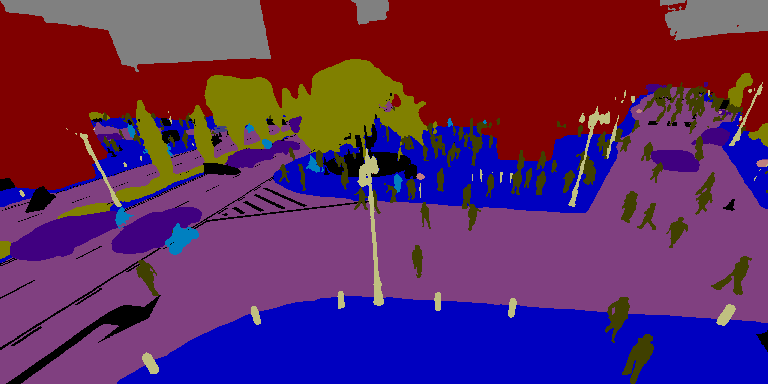}\\[.2mm]
  &
  {\raisebox{1.2\normalbaselineskip}[0pt][0pt]{\rotatebox[origin=c]{90}{\tiny Compare}}} &
  \includegraphics[width=\hsz]{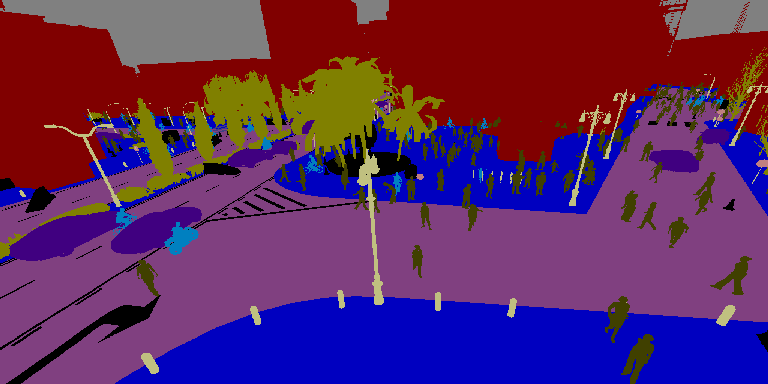} &
  \includegraphics[width=\hsz]{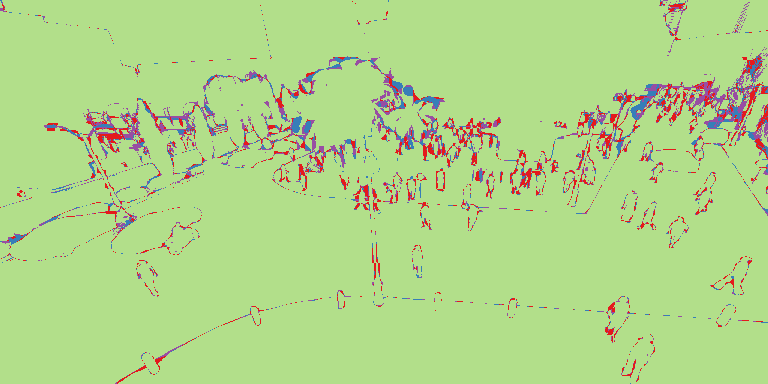} &
  \includegraphics[width=\hsz]{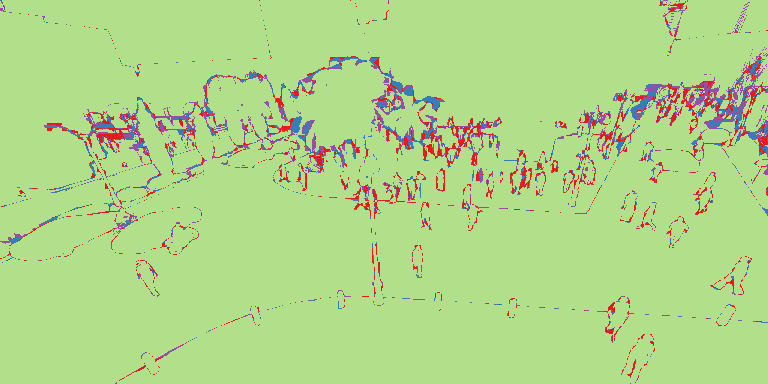} &
  \includegraphics[width=\hsz]{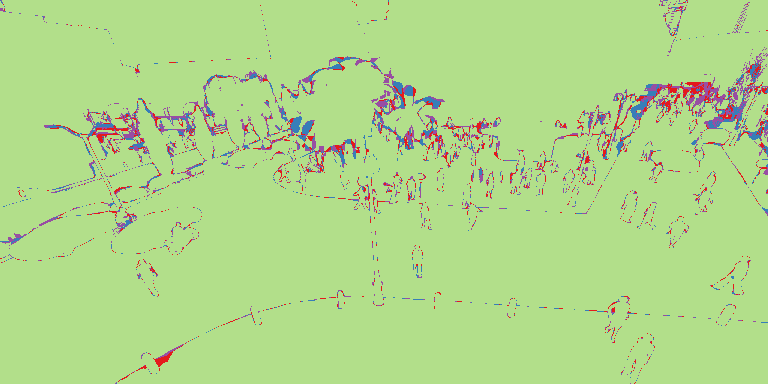} &
  \includegraphics[width=\hsz]{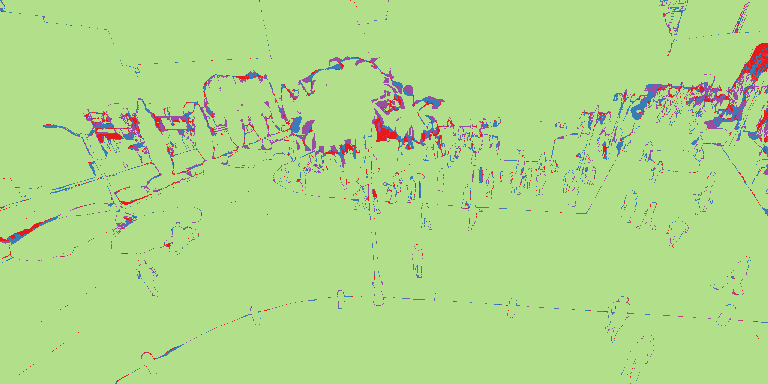} &
  \includegraphics[width=\hsz]{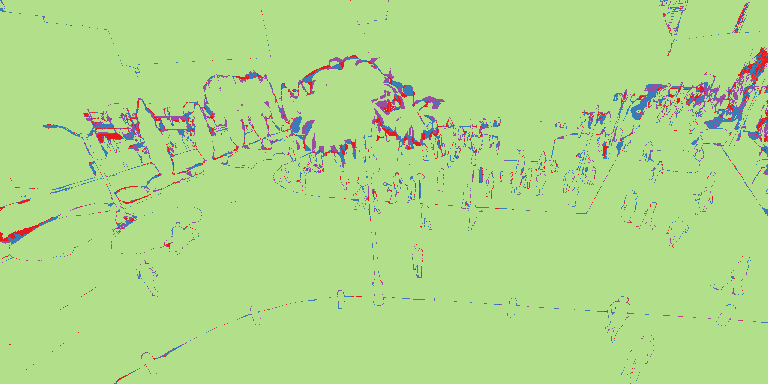} &
  \includegraphics[width=\hsz]{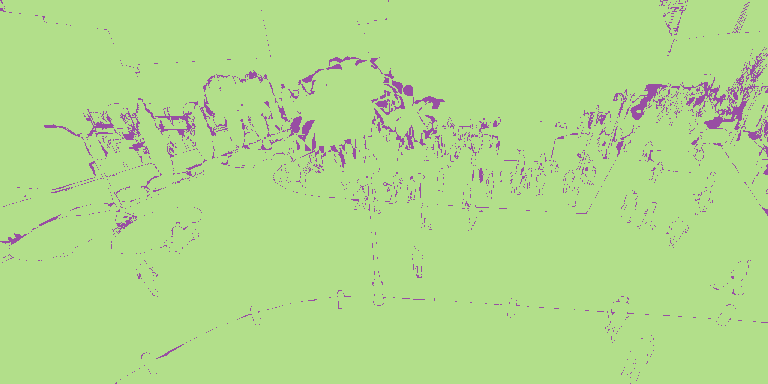}\\
  &
  {\raisebox{1.15\normalbaselineskip}[0pt][0pt]{\rotatebox[origin=c]{90}{\tiny Output}}} &
  \includegraphics[width=\hsz]{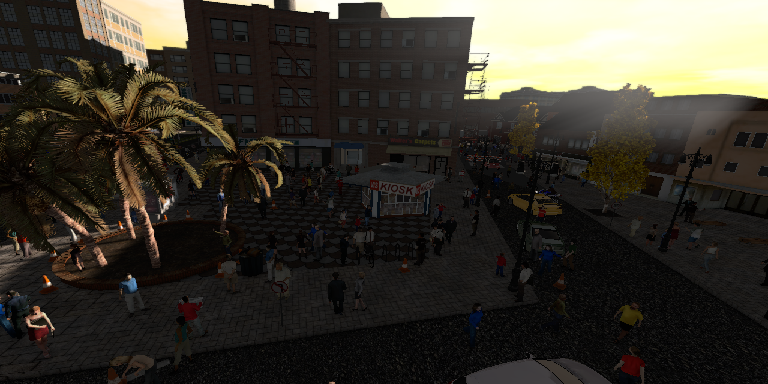} &
  \includegraphics[width=\hsz]{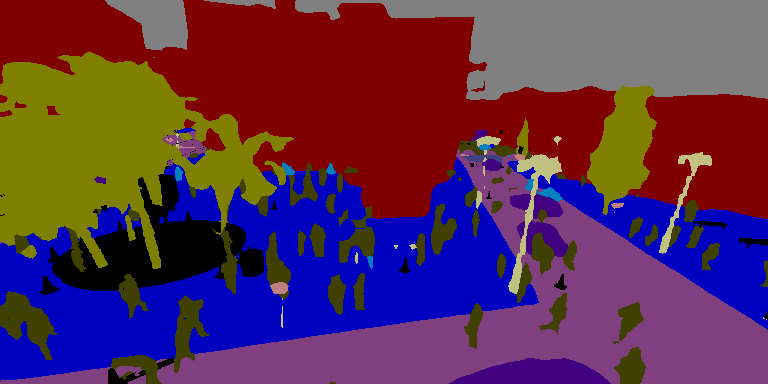} &
  \includegraphics[width=\hsz]{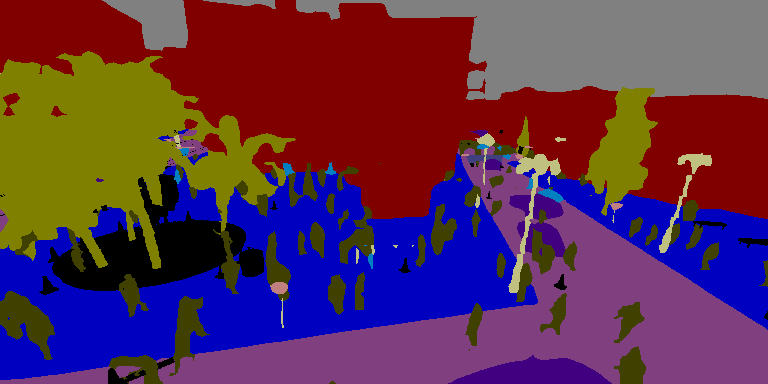} &
  \includegraphics[width=\hsz]{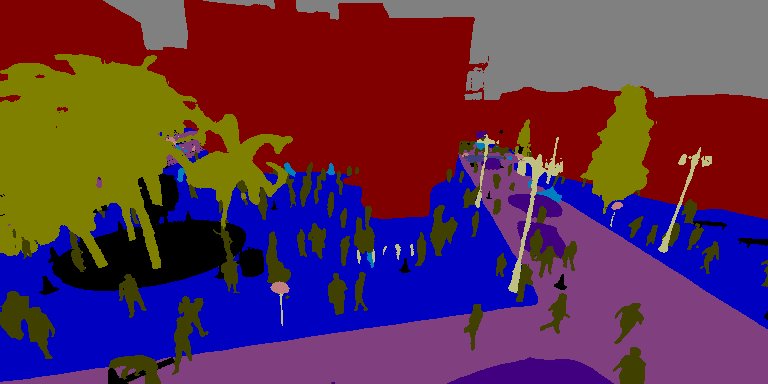} &
  \includegraphics[width=\hsz]{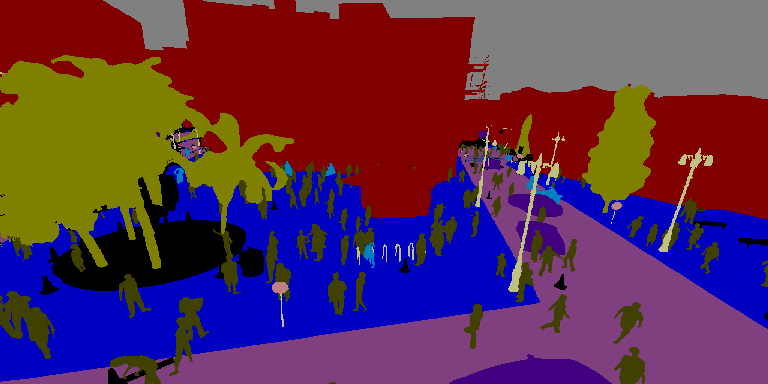} &
  \includegraphics[width=\hsz]{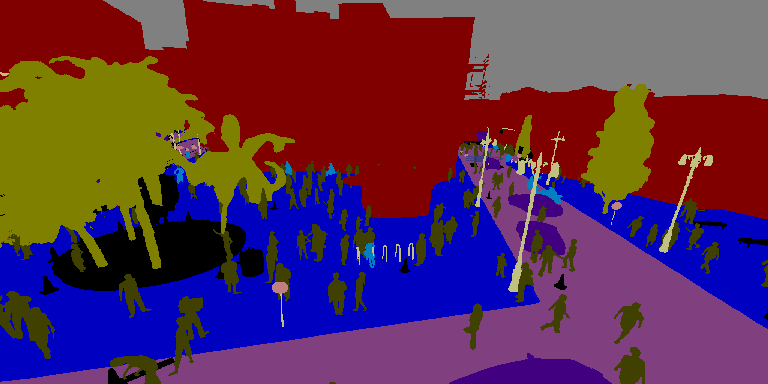} &
  \includegraphics[width=\hsz]{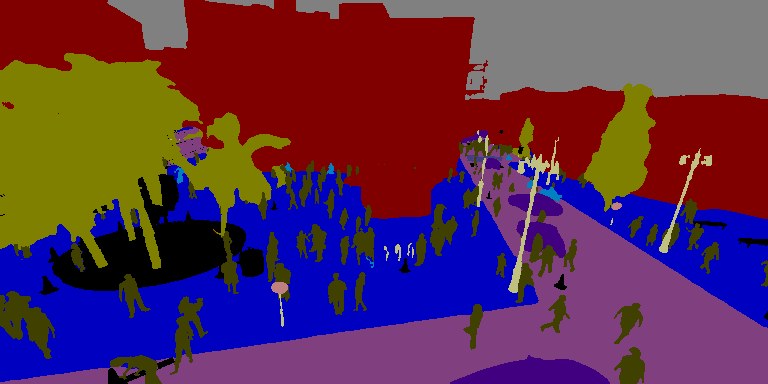}\\[.2mm]
  &
  {\raisebox{1.2\normalbaselineskip}[0pt][0pt]{\rotatebox[origin=c]{90}{\tiny Compare}}} &
  \includegraphics[width=\hsz]{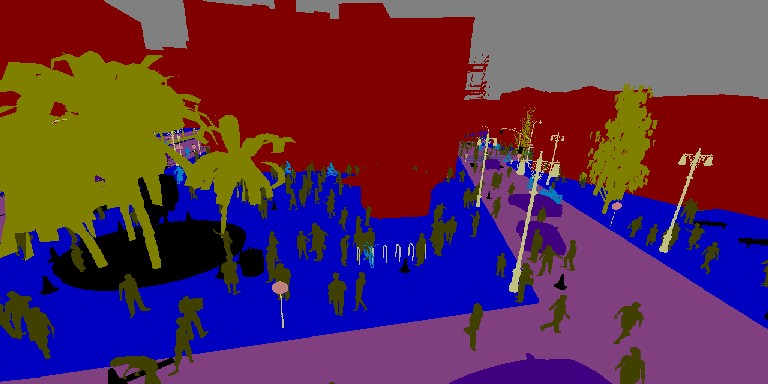} &
  \includegraphics[width=\hsz]{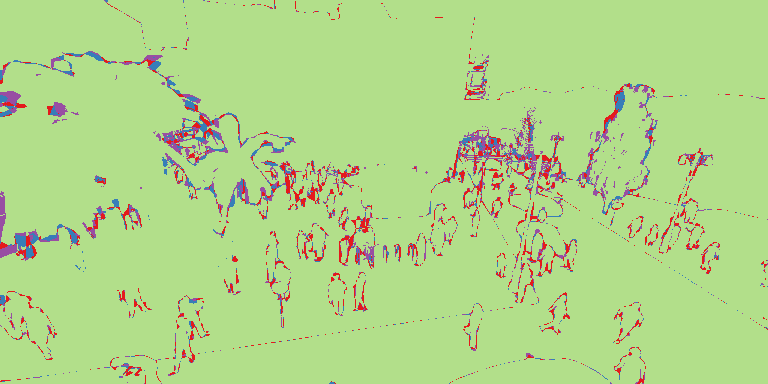} &
  \includegraphics[width=\hsz]{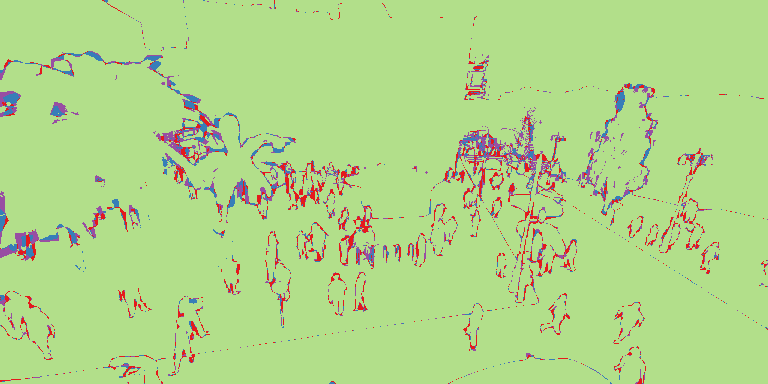} &
  \includegraphics[width=\hsz]{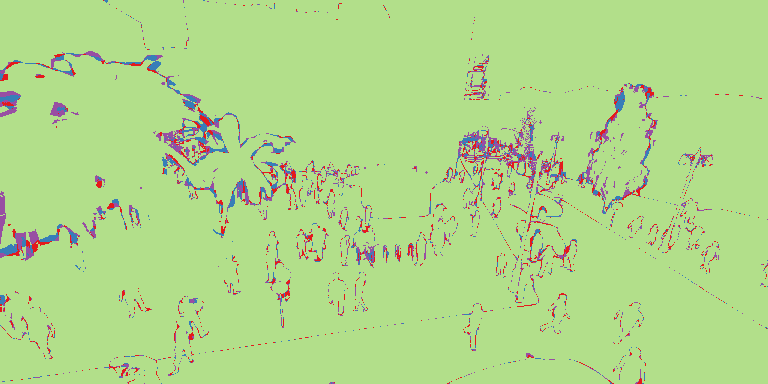} &
  \includegraphics[width=\hsz]{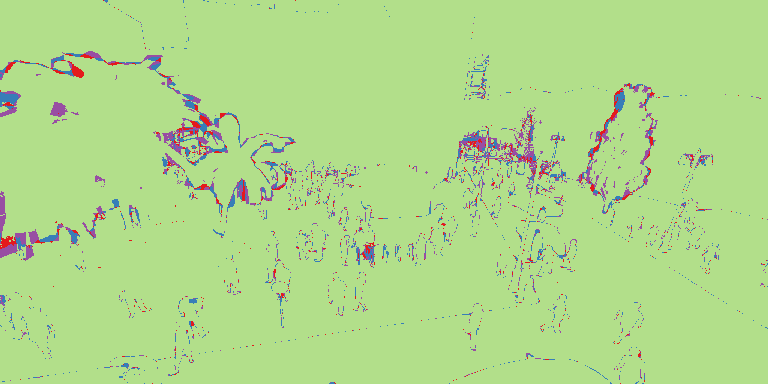} &
  \includegraphics[width=\hsz]{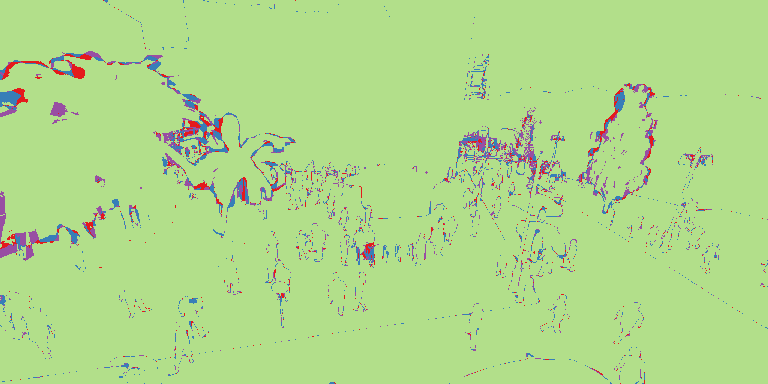} &
  \includegraphics[width=\hsz]{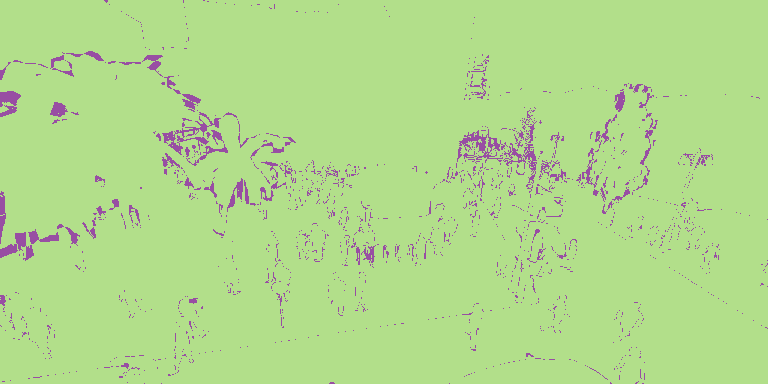}\\
  &
  {\raisebox{1.15\normalbaselineskip}[0pt][0pt]{\rotatebox[origin=c]{90}{\tiny Output}}} &
  \includegraphics[width=\hsz]{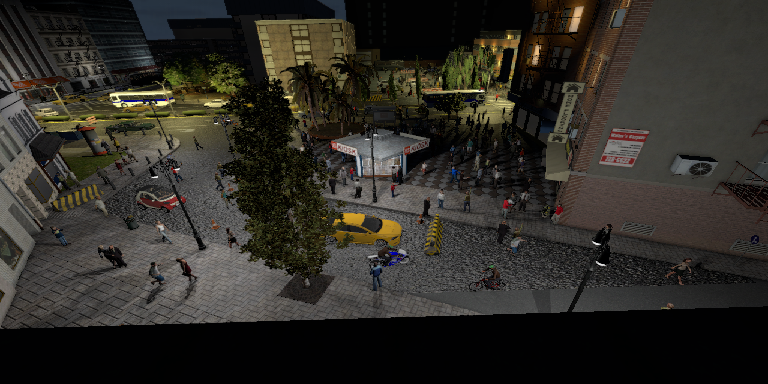} &
  \includegraphics[width=\hsz]{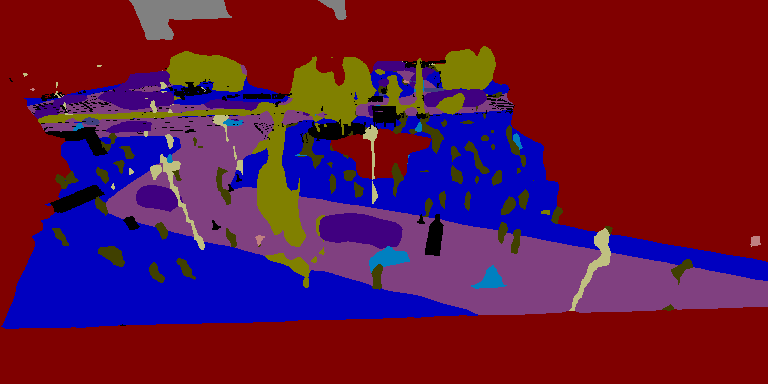} &
  \includegraphics[width=\hsz]{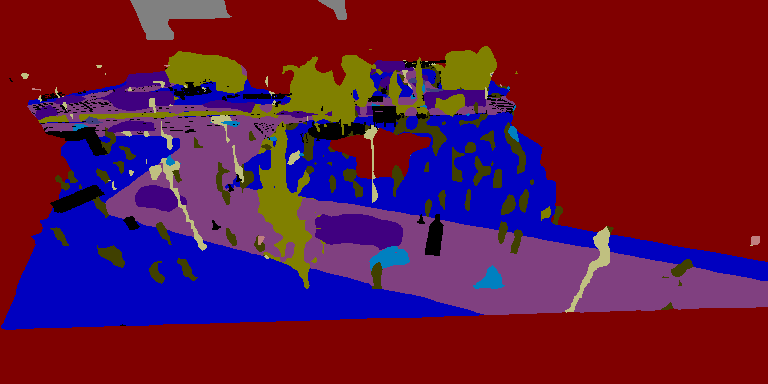} &
  \includegraphics[width=\hsz]{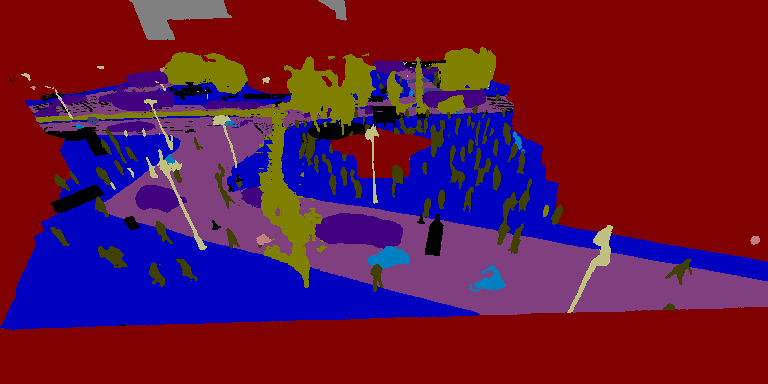} &
  \includegraphics[width=\hsz]{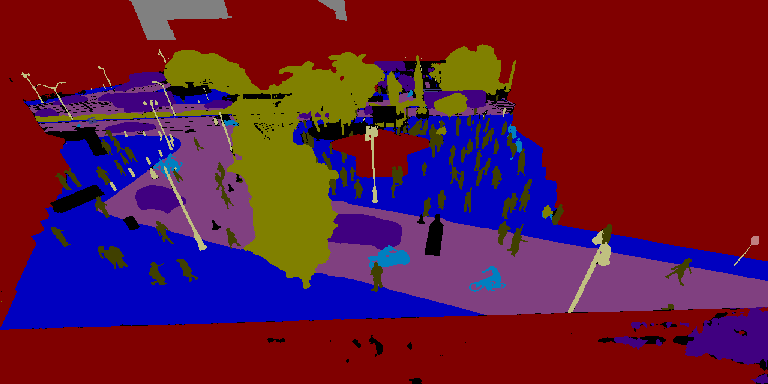} &
  \includegraphics[width=\hsz]{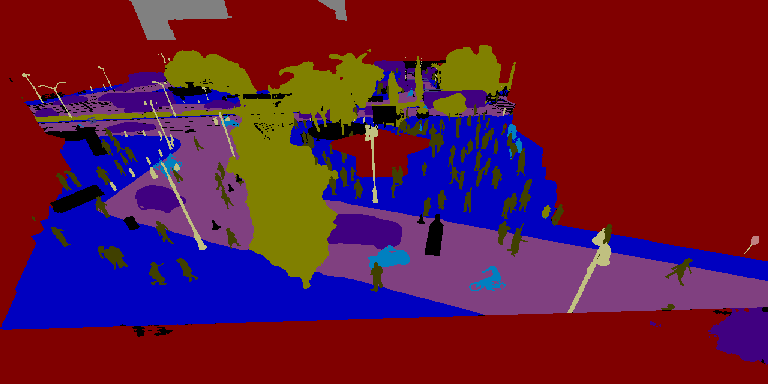} &
  \includegraphics[width=\hsz]{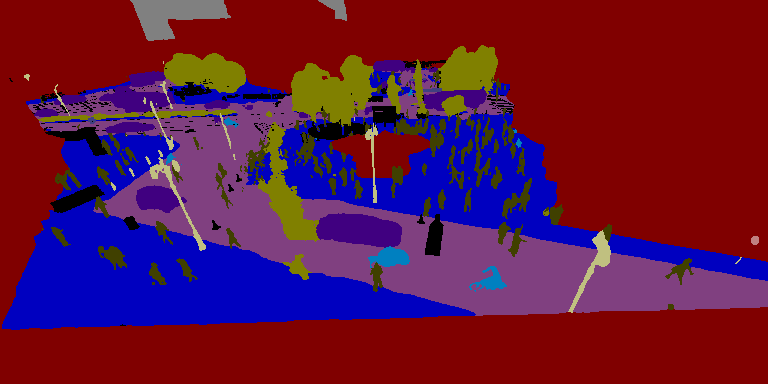}\\[.2mm]
  &
  {\raisebox{1.2\normalbaselineskip}[0pt][0pt]{\rotatebox[origin=c]{90}{\tiny Compare}}} &
  \includegraphics[width=\hsz]{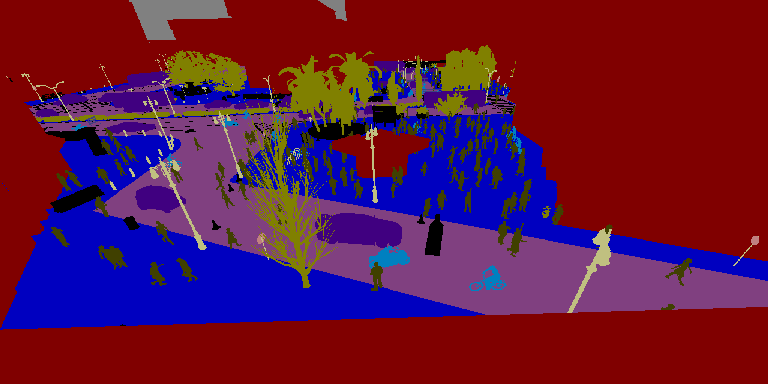} &
  \includegraphics[width=\hsz]{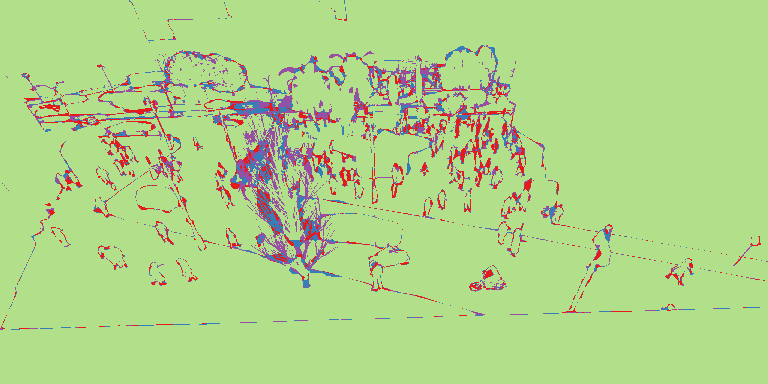} &
  \includegraphics[width=\hsz]{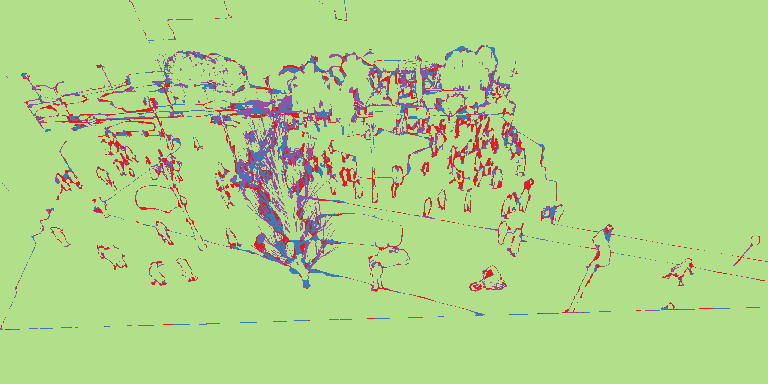} &
  \includegraphics[width=\hsz]{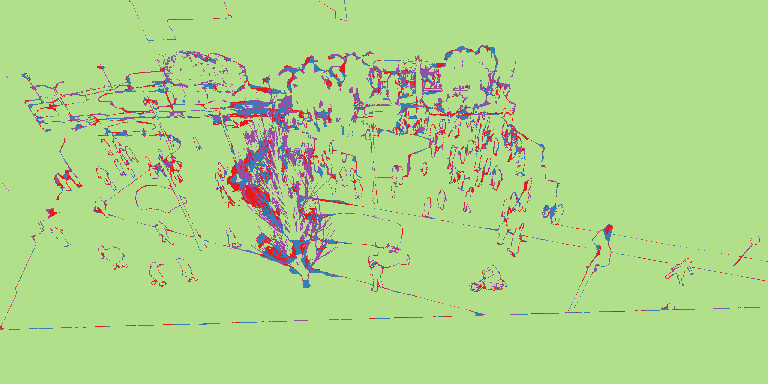} &
  \includegraphics[width=\hsz]{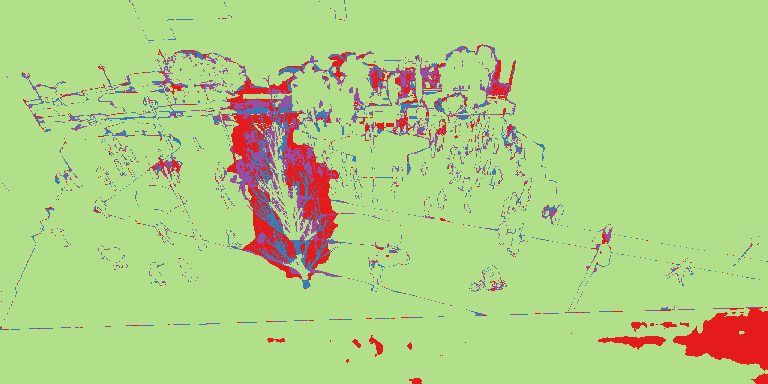} &
  \includegraphics[width=\hsz]{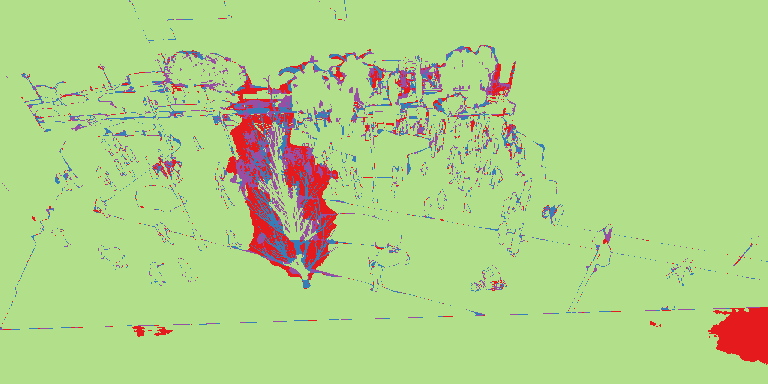} &
  \includegraphics[width=\hsz]{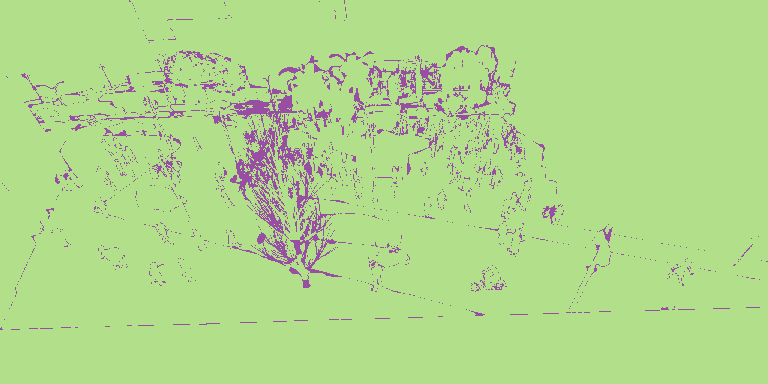}\\
  \\[1.2mm]
  & &
  \scriptsize Image/GT &
  \scriptsize AdapNet++ &
  \scriptsize OCNet &
  \scriptsize DUNet &
  \scriptsize HRNet &
  \scriptsize HRNet+OCR &
  \scriptsize PHGMM \\
\end{tabular}
}
}
\caption{Qualitative comparison of the predictions of the AdapNet++~\cite{Valada2019}, OCNet~\cite{Yuan2018}, DUNet~\cite{Jin2019}, HRNet~\cite{Wang2020}, HRNet+OCR~\cite{Yuan2021}, and PHGMM models against the ground-truth.
The samples are chosen randomly from the validation set.
The prediction rows (output) show the outputs of the different models with their respective labels (colors) of the datasets.
The comparison row shows an overlap of the predictions (PHGMM vs. the model indicated in the column) against the ground truth.
The first column is PHGMM vs. AdapNet++; the second one is PHGMM vs. OCNet++, and so on.
In the comparison images, the green regions are correctly segmented, the red regions are erroneously segmented by original models (AdapNet++, OCNet, DUNet, HRNet, and HRNet+OCR) while the blue ones are errors made by PHGMM\@.
Regions incorrectly segmented by both predictions are purple.}
\label{fig:qualitative_cityscape_synthia}
\end{figure*}

\begin{figure*}[tb]
\centering
\setlength{\hsz}{.25\linewidth}
\setlength{\colfig}{-1pt}

\definecolor{cs-sky}{RGB}{70, 130, 180}
\definecolor{cs-build}{RGB}{70, 70, 70}
\definecolor{cs-road}{RGB}{128, 64, 128}
\definecolor{cs-sidewalk}{RGB}{244, 35, 232}
\definecolor{cs-fence}{RGB}{190, 153, 153}
\definecolor{cs-vege}{RGB}{107, 142, 35}
\definecolor{cs-pole}{RGB}{153, 153, 153}
\definecolor{cs-car}{RGB}{0, 0, 142}
\definecolor{cs-sign}{RGB}{220, 220, 0}
\definecolor{cs-person}{RGB}{220, 20, 60}
\definecolor{cs-cyclist}{RGB}{119, 11, 32}

{\renewcommand{\arraystretch}{0}
  \begin{tabular}{%
      @{}%
      c@{\hspace{\colfig}}
      c@{\hspace{\colfig}}
      c@{\hspace{\colfig}}
      c@{}
    }
    \includegraphics[width=\hsz]{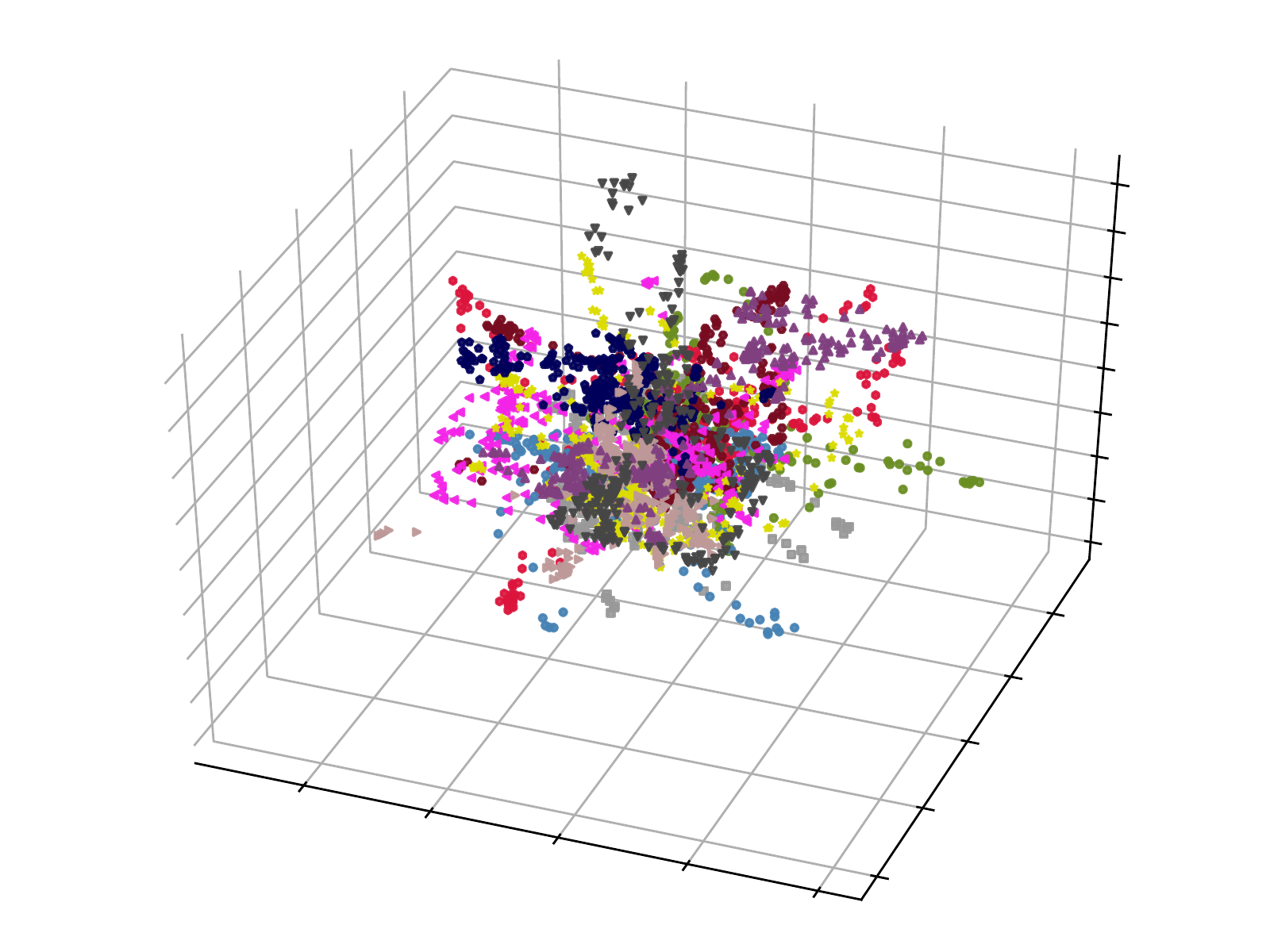} &
    \includegraphics[width=\hsz]{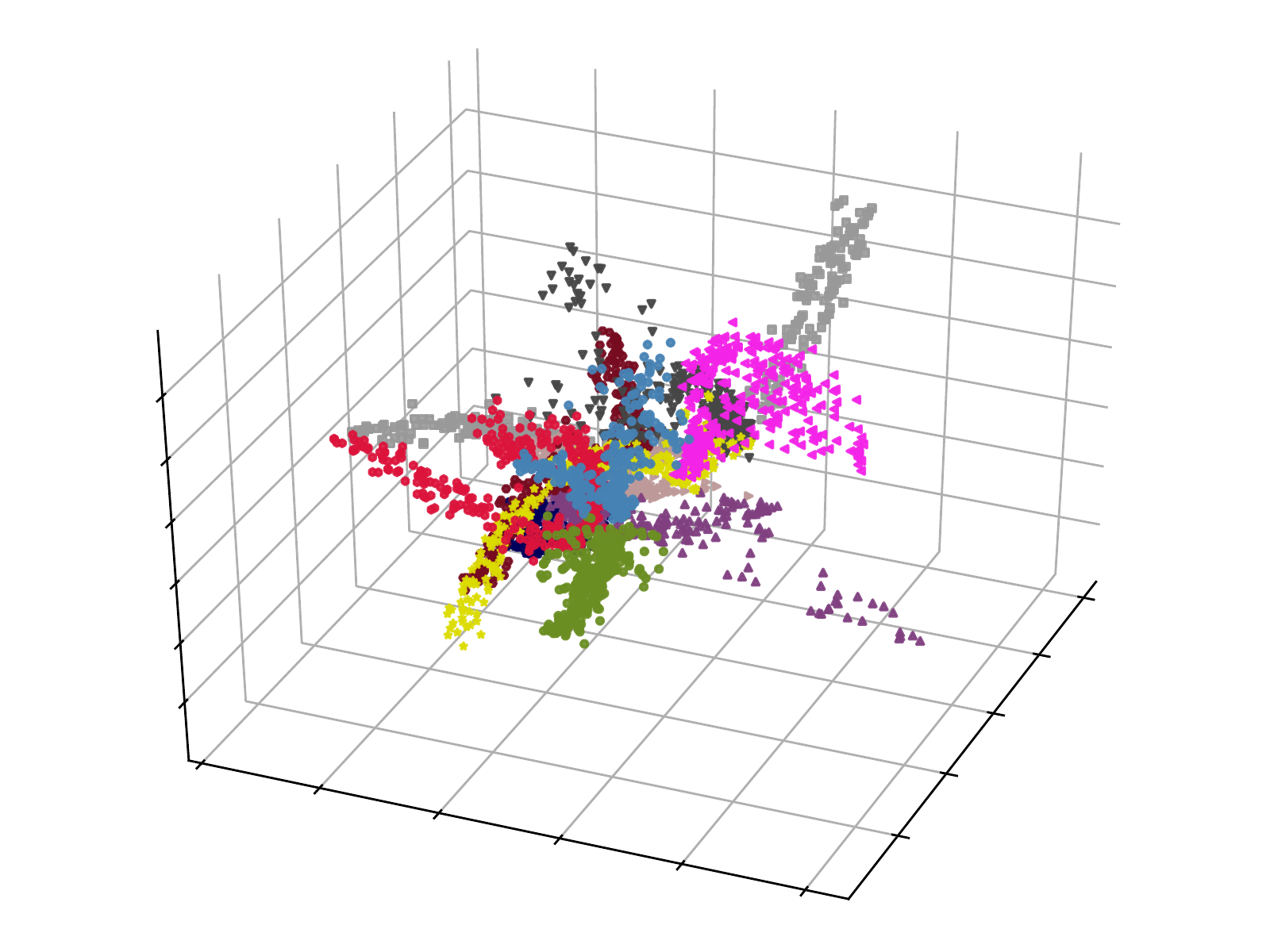} &
    \includegraphics[width=\hsz]{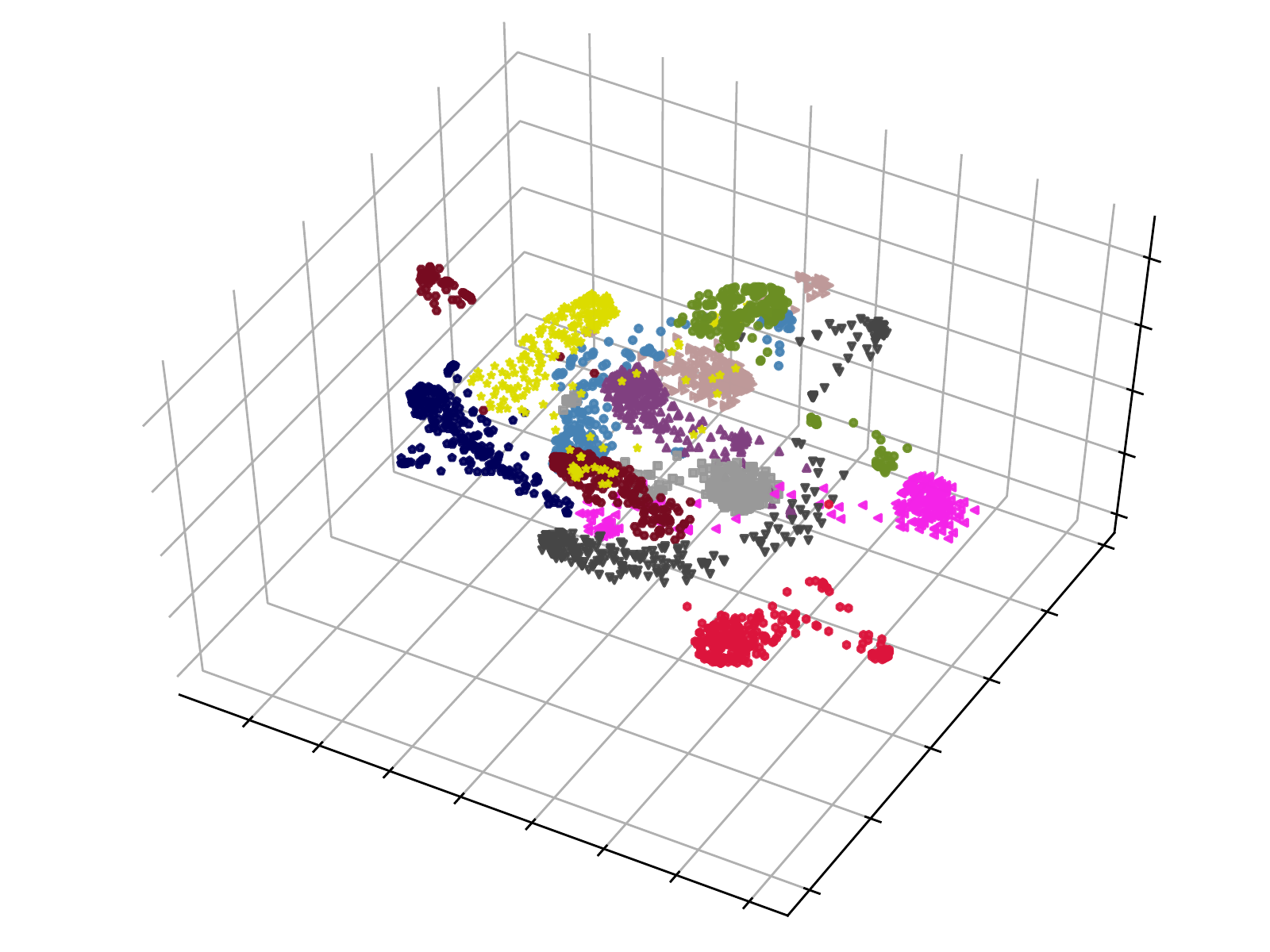} &
    \includegraphics[width=\hsz]{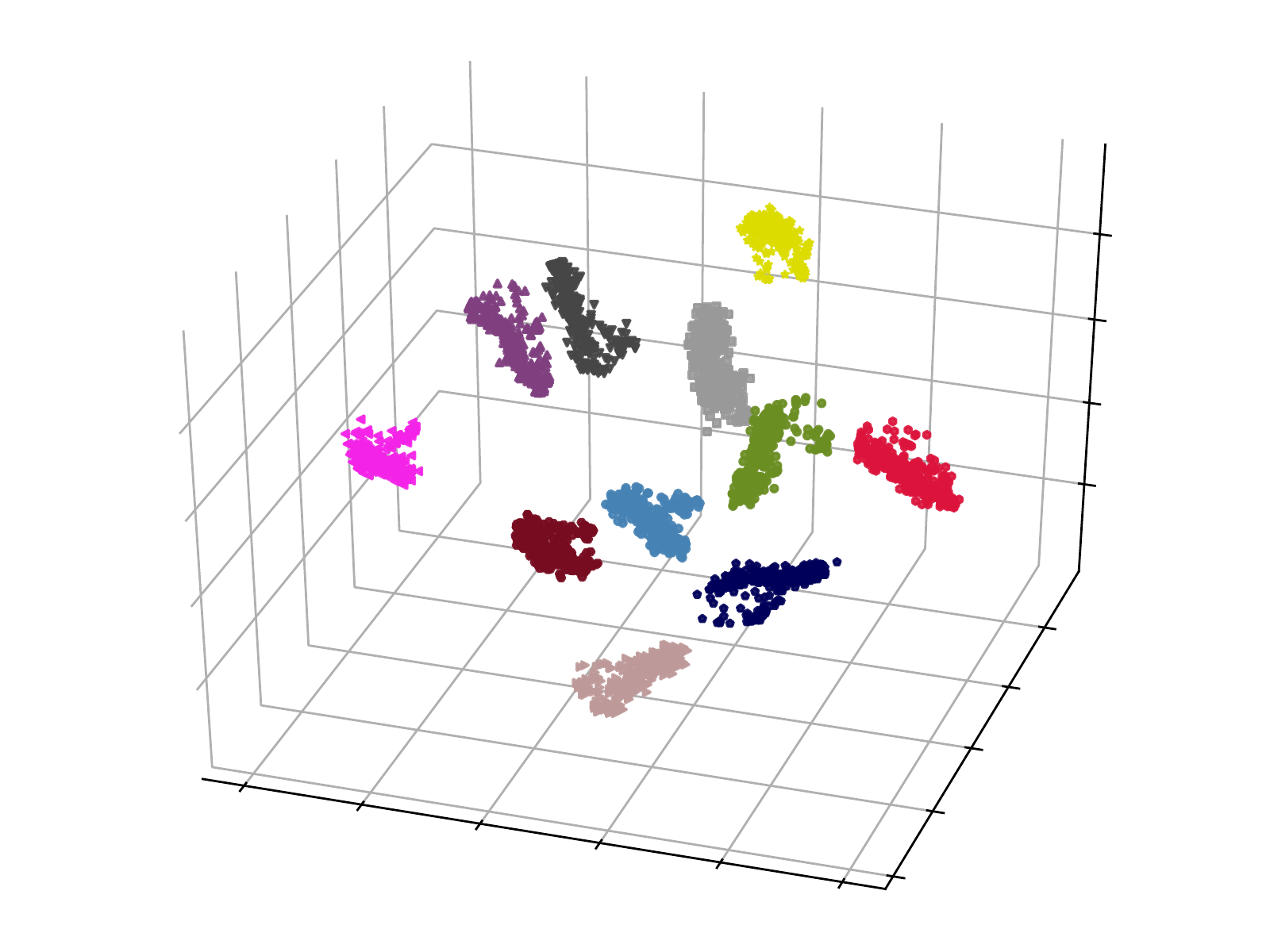} \\[1.2mm]
    \scriptsize Iteration $=10^2$ &
    \scriptsize Iteration $=10^3$ &
    \scriptsize Iteration $=10^4$ &
    \scriptsize Iteration $=10^5$ \\[1.mm]
    \scriptsize SSI $=-0.092$ &
    \scriptsize SSI $=-0.041$ &
    \scriptsize SSI $=0.095$ &
    \scriptsize SSI $=0.703$ \\[1.mm]
    \scriptsize CHI $=127.47$ &
    \scriptsize CHI $=308.57$ &
    \scriptsize CHI $=609.23$ &
    \scriptsize CHI $=12074.30$ \\[1.mm]
    \scriptsize DBI $=6.951$ &
    \scriptsize DBI $=5.144$ &
    \scriptsize DBI $=3.050$ &
    \scriptsize DBI $=0.454$ \\

  \end{tabular}
}
\caption[Plot local latent space]{We show latent space clustering behavior for $z$, using the Cityscape validation dataset, using $11$ classes and with crop size of $384 \times 768$. The classes are: \class{cs-sky}{1} Sky, \class{cs-build}{1} Building, \class{cs-road}{1} Road, \class{cs-sidewalk}{1} Sidewalk, \class{cs-fence}{1} Fence, \class{cs-vege}{1} Vegetation, \class{cs-pole}{1} Pole, \class{cs-car}{1} Car, \class{cs-sign}{1} Sign, \class{cs-person}{1} Person, \class{cs-cyclist}{1} Cyclist.
It through the different iterations measured by Silhouette Coefficient, SSI (higher is better), Calinski-Harabasz Index, CHI (higher is better), and Davies-Bouldin Index, DBI (lower is better). }
\label{fig:vizualization_z}
\end{figure*}

We present a set of experiments (quantitative and qualitative) to compare our PHGMM model with existing models in the literature.
For these, we employed Cityscape and Synthia datasets presented in the previous section.
We present the quantitative results in Tables~\ref{tab:result_cityscape} and~\ref{tab:result_synthia}.
We use the IoU metric (higher is better) for each class on both datasets.
Note that to perform the comparison, we train all the models from scratch, with the same sample size and the same number of classes.
Due to our limited computational resources, we use the sample size (\ie, $768 \times 384$) and the number of classes (\ie, $11$ target classes) defined in previous works by Valada \etal~\cite{Valada2017}.

We present the qualitative results in Fig.~\ref{fig:qualitative_cityscape_synthia}.
The results are displayed on a set of random samples from the validation set.
These figures show the inferences of the models with the best segmentation performance (\ie, IoU).
We present the predictions of the different models named in the columns.
These predictions show each dataset's respective color (label of classes) for Cityscape and Synthia
and are displayed in the odd rows named output.
The even rows named comparison exhibit the comparison (overlap) of our PHGMM model against the different models (AdapNet++, OCNet, DUNet, HRNet, and HRNet+OCR).
In the comparison image (\eg, the first column is PHGMM vs. AdapNet++ second one is PHGMM vs. OCNet++), blue represents regions that the PHGMM model erroneously segments, and red represents those erroneously segmented by the several models (name of the columns).
The regions correctly and incorrectly segmented by both predictions are green and purple, respectively.
Finally, in the last comparison column, we present our model against itself, \ie, PHGMM vs. PHGMM.
Here, we interpret the purple region as showing the incorrectly segmented regions of our PHGMM model.
Note that our results decrease spatial precision loss at the boundaries compared to the other models (quantitative improvement in Fig.~\ref{fig:cityscape_trimap}).

Remember, we conducted this research to address the loss of spatial precision in segmentation.
We use Trimap~\cite{Csurka2013, Krahenbuhl2011}, which focuses on measuring the pixel-wise error on boundary regions of segmentation, with a distance to the boundaries given by trimap width (pixels).
The plot in Fig.~\ref{fig:cityscape_trimap} (error curve comparison) demonstrates that our model has less loss of spatial precision compared to the literature (\ie, HRNet + OCR).

In addition, we present an ablation study (Fig.~\ref{fig:ablation_study_z_t}) on the latent space $z$ (number of clusters $K$ used in the GMM) and latent space $g$ (including or excluding it from the architecture).
While it may be interesting to also evaluate the global features alone, we did not perform this experiment due to the low performance that these features may produce given their location on the decoder (\cf Fig.~\ref{fig:PHGMMPos_train}).
However, from these plots, we can observe that our best results are produced when using the learning representation $z$ (local features) with the number of clusters $k$ equal to the number of classes $c$ in the dataset (\ie, $ c = k = 12$).
Thus, we intuit that, in an unsupervised way, each cluster of $z$ (\ie, GMM) is modeled to obtain the best representation of local features focused on each class.
Furthermore, our ablation study also shows that the latent space $g$ is crucial to improving the final segmentation.
The combination of features ($z+g$) produces the best predictions.

Finally, we visualize the behavior of latent space $z$ (Fig.~\ref{fig:vizualization_z}).
A Mixture of Gaussian fits $z$ through the different iterations using the multidimensional projection method t-SNE~\cite{Maaten2008}.
Additionally, we show the influence of $z$ in the prediction of the segmentation.
The visualization, obtained with t-SNE~\cite{Maaten2008}, shows that the segmentation results are improved by providing clustering behavior to the latent space $z$.

\makeatletter
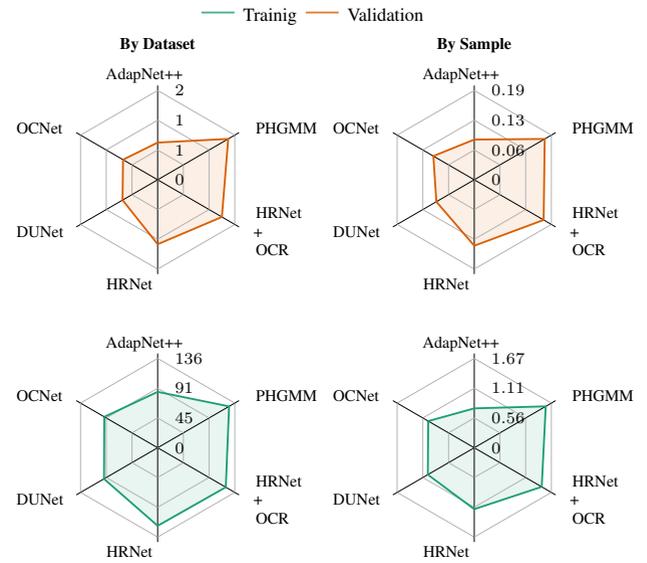
\begin{figure}[tb]\centering
  \resizebox{.95\linewidth}{!}{%
    \begin{tikzpicture}[
    label distance=.5,
    scale=0.15,
    plot/.style={
      thick,
      draw=#1,
      fill=#1,
      opacity=.1
    },
    plot1/.style={plot=Dark2-A},
    plot2/.style={plot=Dark2-B},
    plot3/.style={plot=Dark2-C},
    plot4/.style={plot=Dark2-D},
    /kiviat/.cd,
    font=\scriptsize,
    label style/.append style={text width=.65cm, /kiviat/set anchor, anchor=\orient, align=\alg},
    set anchor/.code={\pgfmathparse{360/\tkz@kiv@radial*\rang > 180 ? 1:0}\ifnum\pgfmathresult>0 \def\orient{west}\def\alg{left} \else \def\orient{east}\def\alg{right}\fi},
    ]
    \begin{scope}[local bounding box=train,shift={(0,0)}]
    \newcommand\KivStep{0.022}
    \pgfmathsetmacro\Unity{1/\KivStep}
    \newcommand\zeroshift{0}

    \tkzKiviatDiagram[
    radial  style/.style ={-},
    rotate=90,
    lattice style/.style ={black!30},
    step=\KivStep,
    gap=3,
    lattice=3,
    ]%
    {AdapNet++, OCNet, DUNet, HRNet, HRNet + OCR, PHGMM}

    \tkzKiviatLine[
    /kiviatline/zero point=\zeroshift,
    plot1
    ](85.322746125,94.31658963,94.91375462,119.13963774,120.45661651,126.31642723)

    \tkzKiviatGrad[unity=\Unity, label precision=0](0)
    \end{scope}

    \begin{scope}[local bounding box=test,shift={(32,0)}]

    \newcommand\KivStep{1.8}
    \pgfmathsetmacro\Unity{1/\KivStep}
    \newcommand\zeroshift{0}

    \tkzKiviatDiagram[
    radial  style/.style ={-},
    rotate=90,
    lattice style/.style ={black!30},
    step=\KivStep,
    gap=3,
    lattice=3,
    font=\scriptsize
    ]%
    {AdapNet++, OCNet, DUNet, HRNet, HRNet + OCR, PHGMM}

    \tkzKiviatLine[
    /kiviatline/zero point=\zeroshift,
    plot1
    ](0.73497945,0.99467231,1.00379154,1.14569878,1.45823654,1.54524451)

    \tkzKiviatGrad[unity=\Unity, label precision=2, zero point=\zeroshift](0) %
    \end{scope}

    \begin{scope}[local bounding box=train,shift={(0,27)}]
    \newcommand\KivStep{1.7}
    \pgfmathsetmacro\Unity{1/\KivStep}
    \newcommand\zeroshift{0}

    \tkzKiviatDiagram[
    radial  style/.style ={-},
    rotate=90,
    lattice style/.style ={black!30},
    step=\KivStep,
    gap=3,
    lattice=3,
    ]%
    {AdapNet++, OCNet, DUNet, HRNet, HRNet + OCR, PHGMM}

    \tkzKiviatLine[
    /kiviatline/zero point=\zeroshift,
    plot2
    ](0.7349794585,0.79316816,0.80319467,1.27317931,1.46984196,1.61524451)

    \tkzKiviatGrad[unity=\Unity, label precision=0](0)
    \end{scope}

    \begin{scope}[local bounding box=test,shift={(32,27)}]

    \newcommand\KivStep{16.0}
    \pgfmathsetmacro\Unity{1/\KivStep}
    \newcommand\zeroshift{0}

    \tkzKiviatDiagram[
    radial  style/.style ={-},
    rotate=90,
    lattice style/.style ={black!30},
    step=\KivStep,
    gap=3,
    lattice=3,
    font=\scriptsize
    ]%
    {AdapNet++, OCNet, DUNet, HRNet, HRNet + OCR, PHGMM}

    \tkzKiviatLine[
    /kiviatline/zero point=\zeroshift,
    plot2
    ](0.08440677,0.09943298,0.09216745,0.13845123,0.16871544,0.17129784)

    \tkzKiviatGrad[unity=\Unity, label precision=2, zero point=\zeroshift](0) %
    \end{scope}

    \node [above] at (test.north) {\scriptsize \textbf{By Sample}};
    \node [above] at (train.north) {\scriptsize \textbf{By Dataset}};

    \matrix [above=8.5pt of $(train.north)!.5!(test.north)$, font=\footnotesize, ampersand replacement=\&] {
      \draw [plot1,scale=0.5,opacity=.8] plot coordinates {(0,0.25)(1,0.25)} -- cycle; \& \node[anchor=base]{Trainig}; \&
      \draw [plot2,scale=0.5,opacity=.8] plot coordinates {(0,0.25)(1,0.25)} -- cycle; \& \node[anchor=base]{Validation}; \\
    };
    \end{tikzpicture}}%
  \caption[Time execution]{%
    Comparison of the execution time of semantic segmentation models for the entire dataset (left) and per sample (right), both in minutes. We use the Cityscapes dataset with \num{17500} and \num{500} samples for training and validation, respectively.
    Note that the PostNet network is only used in the training phase.
    Our PHGMM model in the testing phase presents an execution time comparable to the state-of-the-art.
  }
  \label{fig:time_exec}
\end{figure}
\makeatother

Also, we present an efficiency comparison of the SS models (see Fig.~\ref{fig:time_exec}) for the entire dataset and per sample.
We used the Cityscapes dataset with \num{17500} training samples and \num{500} validation samples (in our case, they are the validation samples) for these experiments.
The results show the execution time of one epoch (a forward-pass over the entire dataset) in a single GPU\@.
We executed the process five times and reported the averages of the entire training and validation dataset (left) and per sample (right), both in minutes.
Our plots on training (Fig.~\ref{fig:time_exec}) show an increase in the time required to train the PHGMM model.
This increment is directly related to the use of the PostNet network and the optimization algorithm used to fit the weights in the training phase.
Note that the PostNet proved robust enough to fit a Gaussian mixture model structure in the latent space.
Remember, we do not use PostNet for the validation dataset, significantly reducing execution time.
Finally, the previous experiments have been conducted on four Nvidia GTX Titan~X GPU \SI{12}{\giga\byte} graphics cards,
\SI{1}{\tera\byte} of RAM, and \num{56} Intel(R) Xeon(R) CPU E5-2680 v4 @ \SI{2.40}{\giga\hertz} (multi-GPU).
Also, we use Python3, Numpy, Pillow, Tensorflow~$1.8$, Docker, and Ubuntu operating system.

Our previous experiments show that the SS task can be improved by endowing/providing a structure to the latent space $z$ (\ie, learning representation) in the form of a Gaussian mixture.
In order to take advantage of this rich feature extraction with structure and behavior clustering, we intelligently merge these features with encoder skip connections at the decoding stage.
We are achieving in this way to recover the geometric information of the segmented objects (\ie, background and boundary of the objects).
In addition, it is necessary to add complementary information (latent space $t$) focused on the contour of the objects, to improve the boundary of the segmentation (see Fig.~\ref{fig:ablation_study_z_t}).
Finally, we address the loss of spatial precision due to the combination of three factors:
the PHGMM architecture (merge features in the decoder stage),
the extraction of specific features given by the latent space $t$, and
the internal structure (Gaussian mixture model) imposed on the latent space $z$.

\section{Future Works}
\label{sec:future_works}
Current models extract the context (generally global) of the objects in the images and delimit the object regions.
However, small regions on the edges of these objects demonstrate spatial difficulties and are lost (see Fig.~\ref{fig:SS_problems}).
This loss in the edges of objects can be caused by noise, blurring, smoothing, and image resolution low, among others.
This work addresses the loss of spatial precision problem in semantic segmentation.
As described in previous sections, the loss of information is commonly reflected in the edge of segmented objects.

Our PHGMM model works with local and global contexts presenting improved results at the edges of segmented objects.
This improvement in the results has a direct correlation with the provided (pushed) clustering behavior in the latent space, see Figs.~\ref{fig:cityscape_trimap} and~\ref{fig:vizualization_z}.
In addition, this correlation is influenced by the number of clusters used in the clustering algorithm in the
training stage, shown in Fig.~\ref{fig:ablation_study_z_t}.
The following research step would be to increase the number of clusters through hierarchical clustering.
In this way, we would group objects semantically at a higher level, and break them down into other sub-classes down in the hierarchy.
Also, to accomplish this model idea, we would need a larger number of classes.

\section{Conclusion}
\label{sec:conclusion}
In this work, we show that by endowing the latent spaces ($z$ and $g$) with clustering behavior and providing them with a structural representation (\ie, GMM for $z$ and Normal distribution for $g$), we improved the results of the SS task.
This improvement is mainly due to the combination (produced in the decoding stage) of low-level features (from global-context information in $g$) and high-level ones (from local-context information in $z$).
The combination of features is performed in the encoder stage with the latent spaces $z$, $g$, and the multi-scale features from the encoder in order to recover the geometric information.
Furthermore, our comparative results show that more details can be extracted from PHGMM by setting the number of clusters equal to the number of classes.
The improvements are produced at the segmented objects' boundaries, thus addressing the loss of spatial precision problem.

\bibliographystyle{IEEEtran}
\footnotesize
\bibliography{abrv,phgmm_references}

\newpage
\begin{IEEEbiography}[{\includegraphics[width=1in,height=1.1in,clip,keepaspectratio]{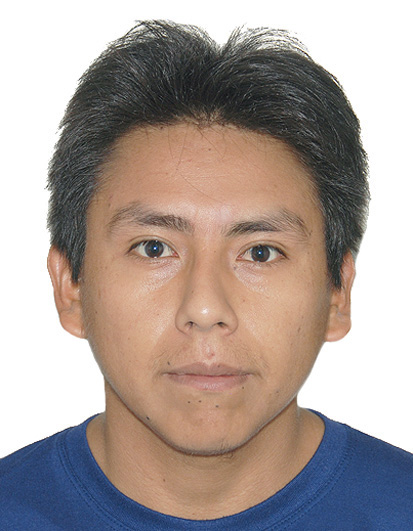}}]{Darwin Saire} received his B.Eng.\ degree in Computer Engineering from Universidad de San Agustin (UNSA), Arequipa in 2013.  He completed his M.Sc.\ degree in Computer Engineering from University of Campinas, Brazil in 2017.  He is currently a Ph.D.\ candidate at the Institute of Computing, University of Campinas, Brazil.  His research interests are computer vision, pattern recognition, image processing, machine learning and deep learning.
\end{IEEEbiography}
\vspace*{-30pt}
\begin{IEEEbiography}[{\includegraphics[width=1in,height=1.1in,clip,keepaspectratio]{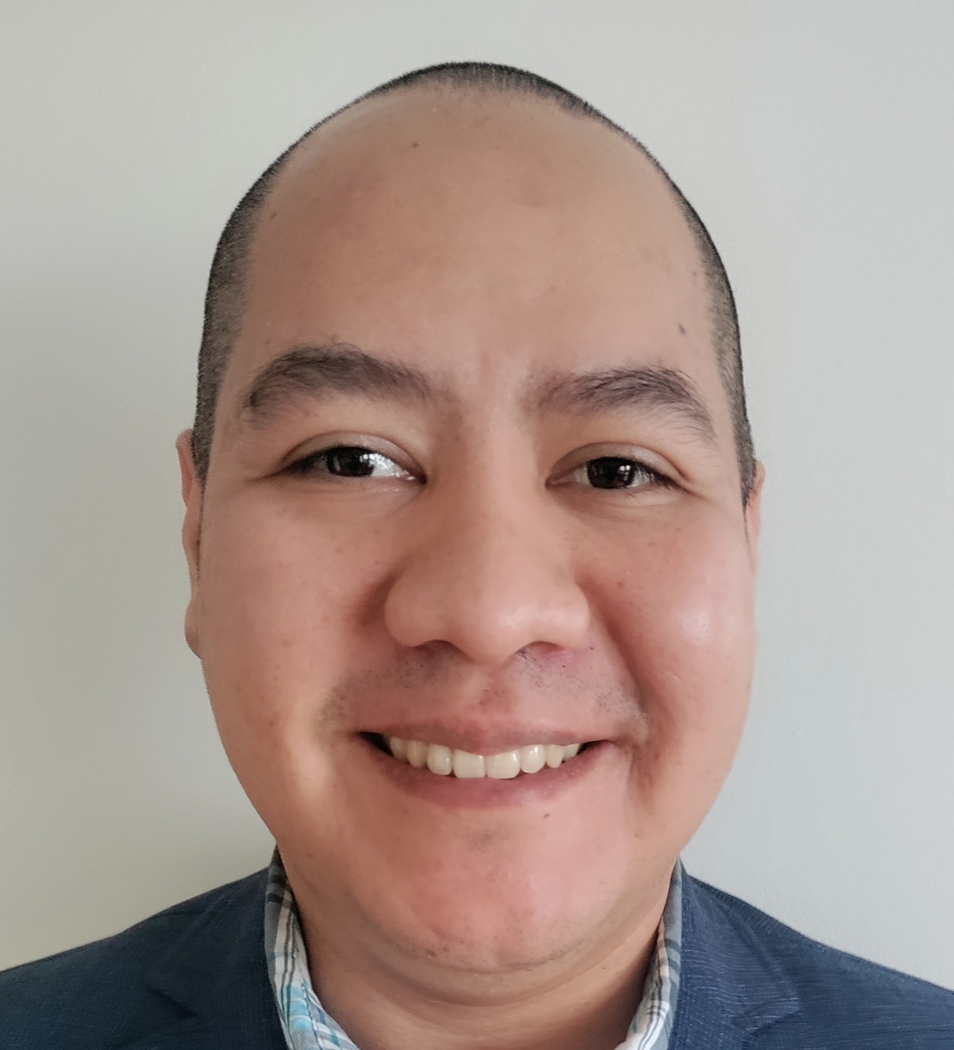}}]{Ad\'{i}n Ram\'{i}rez Rivera} (S'12, M'14, SM'21) received his B.Eng.\ degree in Computer Engineering from Universidad de San Carlos de Guatemala (USAC), Guatemala in 2009.  He completed his M.Sc.\ and Ph.D.\ degrees in Computer Engineering from Kyung Hee University, South Korea in 2013.  He is currently an Associate Professor at the Department of Informatics, University of Oslo, Norway.  His research interests are video understanding (including video classification, semantic segmentation, spatiotemporal feature modeling, and generation), and understanding and creating complex feature spaces.
\end{IEEEbiography}
\vfill

\EOD
\end{document}